%% file: main.tex
\documentclass[12pt]{article}
\usepackage{a4wide}
\usepackage{amssymb,amsmath,bm}
\usepackage{setspace}
\usepackage{natbib}
\usepackage{graphicx}
\usepackage{subcaption}
\usepackage{color}
\usepackage{mdframed}
\usepackage{xr}
\usepackage{hyperref}

\usepackage{multirow}
\usepackage{listings}

\newcommand{\chref}[2]{\href{#1}{\textcolor{blue}{#2}}}

\title{Generalized Kernel Regularized Least Squares\footnote{We thank Michael Auslen, Saad Gulzar, Chad Hazlett, Adeline Lo, Marc Ratkovic, Brandon Stewart, and participants at MPSA 2022 and APSA 2022 for helpful feedback on this project. This research was supported in part by the University of Pittsburgh Center for Research Computing, RRID:SCR\_022735, through the resources provided. Specifically, this work used the H2P cluster, which is supported by NSF award number OAC-2117681. All mistakes and errors are our own. Replication data and code are available at \chref{https://doi.org/10.7910/DVN/WNW0AD}{https://doi.org/10.7910/DVN/WNW0AD} \texttt{R} package to implement the methods in this paper can be found at \chref{https://github.com/mgoplerud/gKRLS}{github.com/mgoplerud/gKRLS}. Appendix~\ref{app:software} provides a demonstration of the accompanying software.}}

\author{Qing Chang\thanks{PhD Candidate, Department of Political Science, University of Pittsburgh. Email: qic47@pitt.edu. Website: \chref{https://qingcchang.com}{qingcchang.com}} ~~and Max Goplerud\thanks{Assistant Professor, Department of Political Science, University of Pittsburgh. Email: mgoplerud@pitt.edu. Website: \chref{https://mgoplerud.com}{mgoplerud.com}}}

\newcommand{\argmax}{\operatornamewithlimits{argmax}}

\begin{document}

\maketitle

\begin{abstract}
    Kernel Regularized Least Squares (KRLS) is a popular method for flexibly estimating models that may have complex relationships between variables. However, its usefulness to many researchers is limited for two reasons. First, existing approaches are inflexible and do not allow KRLS to be combined with theoretically-motivated extensions such as random effects, unregularized fixed effects, or non-Gaussian outcomes. Second, estimation is extremely computationally intensive for even modestly sized datasets. Our paper addresses both concerns by introducing generalized KRLS (\texttt{gKRLS}). We note that KRLS can be re-formulated as a hierarchical model thereby allowing easy inference and modular model construction where KRLS can be used alongside random effects, splines, and unregularized fixed effects. Computationally, we also implement random sketching to dramatically accelerate estimation while incurring a limited penalty in estimation quality. We demonstrate that \texttt{gKRLS} can be fit on datasets with tens of thousands of observations in under one minute. Further, state-of-the-art techniques that require fitting the model over a dozen times (e.g. meta-learners) can be estimated quickly.
    
    \vspace{0.5em}
    
\end{abstract}
\thispagestyle{empty}

\clearpage
\setcounter{page}{1}
\doublespacing

\section{Introduction}

Designing models that can correctly estimate complex interactions between covariates or non-linear effects of continuous predictors is an important but challenging problem. These models are increasingly popular not only as a robustness test to check the impact of functional form assumptions, but also as key constituent components to a variety of increasingly popular machine learning algorithms designed to estimate causal effects.

One popular method in political science to estimate a highly flexible model while maintaining good out-of-sample predictive performance is ``Kernel Regularized Least Squares'' (KRLS; \citealt{hainmueller2014kernel}), also known as ``kernel ridge regression'' (e.g. \citealt{yang2017randomized}). This method provides a flexible approach to estimate a possibly complex underlying function and can easily capture interactions between covariates or non-linear effects of certain predictors. It is simple to use as it only requires the researcher to provide a matrix of relevant predictors.  \cite{hainmueller2014kernel} describe other attractive features. However, it has two noticeable drawbacks that have likely limited its more widespread adoption. First, 
traditional approaches to estimating KRLS are rather inflexible as they require that all variables are included in a single kernel and regularized.\footnote{\texttt{KSPM} \citep{schramm2020kspm} is an exception, although it has some limitations discussed in Appendix~\ref{app:software}.} This prevents common extensions such as (unregularized) fixed effects, random effects, or multiple kernels for different sets of predictors from being included; further, it is challenging to estimate models with non-Gaussian outcomes (e.g., binary, ordered, or categorical outcomes) and difficult to implement alternative standard errors (e.g., cluster-robust standard errors). In many applied settings, researchers desire a ``modular'' approach like that found when using hierarchical models where different variables can be included in the model in different ways based on the researcher's theoretical beliefs.

Second, and equally importantly, traditional versions of KRLS are highly computationally expensive as the cost of estimation is dominated by the cube of the number of observations (\citealt{yang2017randomized,mohanty2019messy}). Without additional modification, it is difficult to fit these models with more than 10,000 observations---and even this may take many hours.

We introduce ``generalized KRLS'' (hereafter \texttt{gKRLS}) to tackle these issues. Our solution has two parts; first, some existing literature shows that (regular) KRLS can be re-formulated as a carefully chosen hierarchical model (e.g., \citealt{liu2007semiparametric,zhang2011bayesian}). Theoretically, this reformulation facilitates a modular model building strategy that can contain multiple kernels in addition to random effects, other smooth terms, and unpenalized fixed effects. However, using rich modular models can considerably complicate estimation using existing approaches given the need to tune multiple different regularization parameters. Fortunately, this hierarchical perspective also facilitates estimation techniques for fast tuning of the regularization parameters without expensive grid searches or cross-validation. These techniques also immediately extend to non-Gaussian outcomes and provide well-calibrated standard errors on key quantities of interest (\citealt{wood2017mgcv}). This reformulation alone, however, is insufficient to make \texttt{gKRLS} practical on large datasets given the cubic cost noted previously. We address this by using the popular ``sub-sampling sketching'' to reduce the cost of estimation by building the kernel based on a random sample of the original dataset (\citealt{drineas2005nystrom,yang2017randomized}).

Our paper proceeds as follows. Sections~\ref{sec:define_gKRLS} and~\ref{sec:scalable_gKRLS} describe \texttt{gKRLS}. Section~\ref{sec:validate_gKRLS} provides two simulations to illustrate its advantages; first, we examine the scalability of \texttt{gKRLS}.\footnote{\cite{dataverse_gKRLS} contains the code to replicate these analyses.} While maintaining accurate estimates, \texttt{gKRLS} takes around six seconds for a dataset with 10,000 observations and two covariates and around two minutes with 100,000 observations without any parallelization and only 8GB of RAM. This compares with hours needed for existing approaches. Our second simulation shows the importance of having a flexible modular approach.  We consider a data generating process that includes fixed effects for a group membership \emph{outside} of the kernel. Traditional KRLS includes the fixed effects in the kernel which assumes the effect of all covariates can vary by group. We find this model is too flexible for modestly-sized datasets and performance can be improved by including the fixed effects as unregularized terms ``outside'' the kernel.

Finally, we conduct two empirical analyses. Section~\ref{sec:newman} reanalyzes \cite{newman2016breaking}'s study of gender and beliefs in meritocracy. Building on theory from the original paper, we use the modular nature of \texttt{gKRLS} to estimate a logistic regression includes three hierarchical terms (random effects, splines, and KRLS) as well as unpenalized covariates (fixed effects). Estimation takes around ten minutes with 8GB of RAM. Section~\ref{sec:main_gulzar} explores \cite{gulzar2020does}'s study of the implications of political affirmative action for development in India. This is a larger dataset (around 30,000 observations), and our preferred model includes many unpenalized covariates and a single kernel. To address regularization bias, we also use \texttt{gKRLS} in algorithms that require  fitting \texttt{gKRLS} between 10 and 15 times (e.g., double/debiased machine learning; \citealt{chernozhukov2018dml}). Estimation takes a few minutes. 

\section{Generalizing KRLS}
\label{sec:define_gKRLS}

There are many different approaches to presenting KRLS (\citealt{hainmueller2014kernel}). We focus on the penalized regression presentation to build connections with hierarchical models. In this view, KRLS creates covariates that measure the similarity of observations (e.g., the transformed distance between covariate vectors) while penalizing the estimated coefficients to encourage estimation of conditional expectation functions that are relatively \emph{smooth} and penalize excessively ``wiggly" functions where the outcome would vary dramatically given small changes in the predictors \citep{hainmueller2014kernel}. This is a common goal for smoothing methods, and different underlying models lead to different design matrices and penalty terms (\citealt[Ch. 5]{wood2017mgcv}). KRLS is especially useful when there are multiple variables that could interact in complex and possibly non-linear ways as it does not require the explicit formulation of which interactions or non-linearities may be relevant. This differs from sparsity-based frameworks such as the LASSO that require creating a set of possibly relevant interactions and bases \emph{before} deciding which ones are relevant.

Formally, assume the dataset has $N$ observations with covariate vectors $\bm{w}_i$. We assume that $\{\bm{w}_i\}_{i=1}^N$ has been standardized---as our software does automatically---to ensure different covariates are comparable in scale. This prevents arbitrary changes (e.g., changing units from meters to feet) from affecting the analysis. \cite{hainmueller2014kernel} center each covariate to have mean zero and variance one. We use Mahalanobis distance to also address potentially correlated input covariates; we thus assume that a mean-centering and whitening transformation has been applied to $\{\bm{w}_i\}_{i=1}^N$ such that the covariance of the stacked $\bm{w}_i$ equals the identity matrix.

Given this standardized data, we create an $N \times N$ kernel matrix $\bm{K}$ that contains the similarity between two observations. We use the popular Gaussian kernel, but our method can be used with other kernels. Equation~\ref{eq:kern} defines $\bm{K}$ that depends on a transformation of the squared Euclidean distance between the observations scaled by the kernel bandwidth which we fix to $P$---the number of covariates in $\bm{w}_i$---following \cite{hainmueller2014kernel}.\footnote{If the design matrix of stacked $\bm{w}_i$ is not full rank, we use its rank instead of $P$ and use a generalized inverse in the whitening transformation.} 

\begin{equation}
\label{eq:kern}
\bm{K}_{ij} = \exp\left(-\frac{||\bm{w}_i - \bm{w}_j||^2}{P}\right)
\end{equation}

In traditional KRLS, $\bm{K}$ becomes the design matrix in a least-squares problem with parameters $\bm{\alpha}$ to predict the outcome $y_i$ with error variance $\sigma^2$. To prevent overfitting, KRLS includes a term that penalizes the wiggliness of the estimated function where a parameter $\lambda$ determines the strength of the penalty.  As $\lambda$ grows very large, all observations are predicted the same value (i.e., there is no effect of any covariate on the outcome). As $\lambda$ approaches zero, the function becomes increasingly wiggly, and predicted values might change dramatically for small changes in the covariates.

Equation~\ref{eq:hh_krls} presents the KRLS objective where $\bm{k}_i$ denotes row $i$ of kernel $\bm{K}$. It is equivalent to traditional KRLS as maximizing Equation~\ref{eq:hh_krls}, for a fixed $\lambda$, gives coefficient estimates $\hat{\bm{\alpha}}_\lambda$ (denoting the dependence on $\lambda$) that are identical to \citet{hainmueller2014kernel}.

\begin{equation}
\label{eq:hh_krls}
\hat{\bm{\alpha}}_{\lambda} = \argmax_{\bm{\alpha}}\left\{-\frac{1}{2\sigma^2} \left[\sum_{i=1}^N (y_i - \bm{k}_i^T\bm{\alpha})^2 + \lambda \bm{\alpha}^T \bm{K} \bm{\alpha}\right]\right\}; \quad
\hat{\bm{\alpha}}_{\lambda} = \left(\bm{K} + \lambda \bm{I}\right)^{-1} \bm{y}
\end{equation}

We start by viewing the problem from a more Bayesian perspective and choose a Gaussian prior for $\bm{\alpha}$ that implies a posterior mode on $\bm{\alpha}$, conditional on $\sigma^2$ and $\lambda$, that is identical to the penalized objective (see also Appendix 2 of \citealt{hainmueller2014kernel}). This prior, sometimes known as the ``Silverman g-prior'', can also be derived from an independent and identically distributed Gaussian prior on each of the coefficients from the underlying feature space associated with the kernel $\bm{K}$ (\citealt{zhang2011bayesian}). Thus, KRLS can be viewed as a traditional random effects model (or ridge regression) on the feature space associated with $\bm{K}$. Equation~\ref{eq:bayesian} displays this generative view of KRLS where $\bm{K}^{-}$ denotes the pseudo-inverse of $\bm{K}$ in the case of a non-invertible kernel.

\begin{align}
\label{eq:bayesian}
y_i \sim N(\bm{k}_i^T\bm{\alpha}, \sigma^2); \quad \bm{\alpha} \sim N\left(\bm{0}, \frac{\sigma^2}{\lambda} \bm{K}^-\right)
\end{align}

A key advantage of this Bayesian view is that KRLS becomes simply a hierarchical model with particular choice of design and prior.  This leads to the idea of ``modular'' model construction where different priors are used for different components of the model. For example, it is common to have unpenalized terms (e.g., ``fixed effects'') alongside the regularized terms. Alternatively, theory may call for the inclusion of more traditional random effects for a geographic unit such as county. We define generalized KRLS, therefore, as a hierarchical model with at one least KRLS term on some covariates. Equation~\ref{eq:multi_gKRLS} presents the general model. Fixed effects ($\bm{\beta}$) have design $\bm{x}_i$ for each observation. There are $J$ penalized terms, indexed by $j \in \{1, \cdots, J\}$, with parameters $\bm{\alpha}_j$ and designs $\bm{z}_{ij}$. As is standard for hierarchical models, each $\bm{\alpha}_j$ has a multivariate normal prior with \emph{precision} $\bm{S}_j$. Each hierarchical term $j$ has its own parameter $\lambda_j$ that governs the amount of regularization.

\begin{subequations}
\label{eq:multi_gKRLS}
\begin{align}
&y_i \sim N\left(\bm{x}_i^T \bm{\beta} + \sum_{j=1}^J \bm{z}_{ij}^T \bm{\alpha}_j,~\sigma^2\right); \quad \bm{\alpha}_j \sim N\left(\bm{0}, \frac{\sigma^2}{\lambda_j} \bm{S}^-_j\right) \quad \mathrm{for} \quad j \in \{1, \cdots, J\} \\
\label{eq:logposterior_multi}
&\ln p\left(\bm{\beta}, \{\bm{\alpha}_j\}_{j=1}^J |~ \{y_i\}_{i=1}^N, \sigma^2, \{\lambda_j\}_{j=1}^J\right) \propto-\frac{1}{2\sigma^2} \left[\begin{array}{l}\sum_{i=1}^N \left(y_i - \bm{x}_i^T\bm{\beta} - \sum_{j=1}^J \bm{z}_{ij}^T \bm{\alpha}_j\right)^2 + \\ \sum_{j=1}^J \lambda_j \bm{\alpha}_j^T \bm{S}_j \bm{\alpha}_j\end{array}\right]
\end{align}
\end{subequations}

Specific choices of design and prior give well-known models. If $\bm{z}_{ij}$ is a vector of group membership indicators and $\bm{S}_j$ is an identity matrix, this is a traditional random intercept. If $\bm{z}_{ij} = \bm{k}_i$ and $\bm{S}_j = \bm{K}$, we recover KRLS from Equation~\ref{eq:bayesian}. 

If one fixes $\sigma^2$ and $\{\lambda_j\}_{j=1}^J$, point estimates can be obtained by maximizing the log-posterior (Equation~\ref{eq:logposterior_multi}). Equation~\ref{eq:pen_estimates} shows the estimates, noting their dependence on the vector of smoothing parameters denoted as $\bm{\lambda} = \{\lambda_j\}_{j=1}^J$. Despite our different presentation, this gives identical point estimates to classical presentations of multilevel models (e.g., \citealt{hazlett2022mlm}; see our Appendix~\ref{app:robust_SE}).  We use $\bm{X}$ for the design of the fixed effects; $\bm{Z}$ denotes the matrix corresponding to all of the design matrices for the hierarchical effects stacked together and $\bm{\alpha}$ denotes the concatenated parameters $\{\bm{\alpha}_j\}_{j=1}^J$. $\bm{S}_{\bm{\lambda}}$ represents the block-diagonal concatenation of each penalty term $\lambda_j \bm{S}_j$.

\begin{equation}
\label{eq:pen_estimates}
\left[\begin{array}{l} \hat{\bm{\beta}}_{\bm{\lambda}} \\ \hat{\bm{\alpha}}_{\bm{\lambda}} \end{array}\right] = \left[\begin{array}{ll} \bm{X}^T\bm{X} & \bm{X}^T\bm{Z} \\ \bm{Z}^T\bm{X} & \bm{Z}^T\bm{Z} + \bm{S}_{\bm{\lambda}}\end{array}\right]^{-1} \left[\begin{array}{l} \bm{X}^T \\ \bm{Z}^T\end{array} \right]\bm{y}
\end{equation}

A key difficulty in using (generalized) KRLS is choosing the appropriate amount of regularization, i.e. calibrating $\{\lambda_1, \lambda_2, \cdots, \lambda_J\}$. In the case of a single KRLS term (e.g., $J=1$) and a Gaussian likelihood, \cite{hainmueller2014kernel} use an efficient method where the leave-one-out cross-validated error can be computed as a function of $\lambda$ and requires only a single decomposition of the kernel $\bm{K}$. One could employ $K$-fold cross-validation to tune $\lambda$ if a non-Gaussian likelihood were used \citep{sonnet2018krls}. However, existing strategies encounter considerable challenges when there are multiple hierarchical terms ($J > 1$). Since \cite{hainmueller2014kernel}'s method may not be available, a popular alternative---grid searches across different possible values for each $\lambda_j$ to minimize some criterion (e.g., cross-validated error)---is very costly even for modest $J$.

Our hierarchical and Bayesian perspective provides a different strategy for tuning $\bm{\lambda}$ for any choice of $J$: Restricted Maximum Likelihood (REML).\footnote{\cite{wood2017mgcv} discusses other criterion, e.g., generalized cross-validation, that could be employed.}  This approach observes that $\bm{\beta}$ has a flat (improper) prior and considers the marginal likelihood after integrating out $\bm{\beta}$ and all $\bm{\alpha}_j$. A REML strategy estimates $\bm{\lambda}$ and $\sigma^2$ by maximizing the log of this marginal likelihood; this is also referred to as an empirical Bayes approach (\citealt[p. 263]{wood2017mgcv}). ~Equation~\ref{eq:REML} shows this objective, noting that it is a function of $\bm{\lambda}$ and $\sigma^2$. $\ell(\hat{\bm{\beta}}_{\bm{\lambda}}, \hat{\bm{\alpha}}_{\bm{\lambda}})$ denotes the log-likelihood (Equation~\ref{eq:logposterior_multi}) evaluated at the penalized estimates given $\bm{\lambda}$ (Equation~\ref{eq:pen_estimates}). $|\bm{S}|_+$ denotes the product of the non-zero eigenvalues of $\bm{S}$; $M_p$ is the dimension of the null space of $\bm{S}_{\bm{\lambda}}$.

\begin{align}
    \label{eq:REML}    
    \hat{\bm{\lambda}}, \widehat{\sigma^2} = \argmax_{\bm{\lambda}, \sigma^2}\left\{\begin{aligned}&\ell(\hat{\bm{\beta}}_{\bm{\lambda}}, \hat{\bm{\alpha}}_{\bm{\lambda}}) - \frac{\hat{\bm{\alpha}}_{\bm{\lambda}}^T \bm{S}_{\bm{\lambda}} \hat{\bm{\alpha}}_{\bm{\lambda}}}{2\sigma^2} + \frac{\ln|\bm{S}_{\bm{\lambda}}/\sigma^2|_+}{2}  + \\ &\quad - \frac{1}{2} \ln\left\vert \frac{1}{\sigma^2}\left[\begin{array}{ll} \bm{X}^T\bm{X} & \bm{X}^T\bm{Z} \\ \bm{Z}^T\bm{X} & \bm{Z}^T\bm{Z} + \bm{S}_{\bm{\lambda}} \end{array}\right]\right\vert + \frac{M_p}{2} \ln(2\pi) \end{aligned}\right\}
\end{align}

After finding $\hat{\bm{\lambda}}$ and $\widehat{\sigma^2}$, point estimates for $\bm{\beta}$ and $\bm{\alpha}$ are obtained by plugging the estimated $\hat{\bm{\lambda}}$ into Equation~\ref{eq:pen_estimates}. \cite{liu2007semiparametric} use the REML approach for a single KRLS hierarchical term (e.g., $J=1$), and we push that intuition further by noting that that KRLS can be part of a general $J$ approach to hierarchical and generalized additive models.

In practical terms, \cite{wood2017mgcv} summarizes the extensive research into numerically stable and efficient approaches to optimizing Equation~\ref{eq:REML} and describes well-established and high-quality software (\texttt{mgcv} in \texttt{R}). For very large problems (in terms of the number of observations or parameters), further acceleration may be needed. Appendix~\ref{app:gam_bam} discusses a set of less stable but faster estimation techniques implemented in the same software.

The final piece of inference is quantifying uncertainty. The Bayesian perspective on hierarchical models suggests using the inverse of the Hessian of the log-posterior on $\{\bm{\beta}, \bm{\alpha}\}$ for the estimated variance matrix (\citealt{wood2017mgcv}).\footnote{\cite{wood2016smoothing} discuss how to incorporate uncertainty from estimating $\hat{\bm{\lambda}}$.} In the linear case, this is the first term in Equation~\ref{eq:pen_estimates}, scaled by $\widehat{\sigma^2}$. Appendix~\ref{app:robust_SE} summarizes existing literature that suggests this should have good frequentist coverage.

\subsection{Extensions to Generalized KRLS}

The above presentation focused on a Gaussian outcome with arbitrary $J$ and homoskedastic errors. We discuss four important extensions that our hierarchical perspective facilitates. First, the preceding exposition is easily generalized to non-Gaussian likelihoods:  One changes the likelihood in Equation~\ref{eq:multi_gKRLS}, e.g. $y_i \sim \mathrm{Poisson}(\exp(\psi_i))$ where $\psi_i = \bm{x}_i^T\bm{\beta} + \sum_{j=1}^J \bm{z}_{ij}^T\bm{\alpha}_j$, and adjusts the objective in Equation~\ref{eq:REML}. This is justified using a Laplace approximation for evaluating the integral of the log-posterior; $\hat{\bm{\beta}}_{\bm{\lambda}}$ and $\hat{\bm{\alpha}}_{\bm{\lambda}}$ are obtained using penalized iteratively re-weighted least squares (\citealt[Ch. 3]{wood2017mgcv}).

Second, the hierarchical perspective also justifies robust and/or clustered standard errors. Appendix~\ref{app:robust_SE} provides a detailed justification of the typical ``sandwich'' formula with slight modifications. We also show existing standard errors for KRLS (\citealt{hainmueller2014kernel}) differ from those derived using the Bayesian perspective discussed above. A simple example suggests that using the Bayesian perspective results in considerably better coverage.

Third, a key use for \texttt{gKRLS} is in machine learning techniques such as stacking or double/debiased machine learning. We provide a software integration of \texttt{gKRLS} (and \texttt{mgcv}) into popular packages for both methods; Appendix~\ref{app:software} provides details.

Finally, we provide new software for easily calculating marginal effects and predicted outcomes for a variety of likelihoods (e.g., Gaussian, binomial, multinomial, etc.). Among other quantities, this allows users to calculate the ``average marginal effect'' (i.e., the partial derivative of the prediction with respect to a specific covariate averaged across all observations in the data; \citealt{hainmueller2014kernel}). Appendix~\ref{app:mfx_diff} provides details. We are able to properly incorporate uncertainty for \emph{both} fixed and random effects for these quantities.

\section{Improving Scalability of Generalized KRLS}
\label{sec:scalable_gKRLS}

The optimism of the above discussion, however, elides a critical limitation of \texttt{gKRLS} as currently proposed. We focus on the traditional KRLS case ($J=1$, no fixed effects) to illustrate the problem. Recall that the model has $N$ observations but requires the estimation of $N$ coefficients. Estimation is extremely time- and memory-intensive as the computational cost is roughly cubic in the number of observations and requires storing a possibly huge $N \times N$ 
matrix (\citealt{hainmueller2014kernel,yang2017randomized}). While some work in political science has focused on reducing this cost, the fundamental problem remains and, in practice, limits its applicability to around 10,000 observations with 8GB of memory (\citealt{mohanty2019messy}) and possibly taking hours to estimate---as Section~\ref{sec:validate_gKRLS} shows. Thus, using \texttt{gKRLS} without modifications is simply impractical for most applied settings. Further, if one needs to fit the model repeatedly (e.g., for cross-validation), it is prohibitively expensive.

Fortunately, there is a large literature on how to approximately estimate kernel methods on large datasets. We employ ``random sketching'', focusing on ``sub-sampling sketching'' or ``uniform sampling'' (e.g., \citealt{drineas2005nystrom,yang2017randomized,lee2020econometric}) to dramatically accelerate the estimation; other methods could be explored in future research (e.g., random features; \citealt{rahimi2007random}).\footnote{Appendix~\ref{app:alt_sketch} discusses an alternative form of sketching  (``Gaussian sketching'')  and shows it incurs a significantly higher computational cost at little systematic improvement in performance.}  The sub-sampling sketching method takes a random sample of $M$ data points and uses them to build the kernel, reducing the size of the design to $N \times M$. If $M$ is much smaller than $N$, this can reduce the cost of estimation considerably. Formally, define the $M$ sampled observations as $\bm{w}^*_{m},~m \in \{1, \cdots, M\}$. If $k(\bm{w}_i, \bm{w}_m)$ is the function to evaluate the kernel (e.g., Equation~\ref{eq:kern}), define the sketched kernel $\bm{K}^*$ as an $N \times M$ matrix with the $(i,m)$-th element as follows:

\begin{equation}
    \bm{K}^*_{im} = k(\bm{w}_i, \bm{w}^*_m)
\end{equation}

Equivalently, one can define $\bm{K}^*$ by multiplying $\bm{K}$ by a sketching matrix $\bm{S}$ with dimensionality $M \times N$, i.e. $\bm{K}^* = \bm{K}\bm{S}^T$. For sub-sampling sketching, $\bm{S}$ is proportional to a sparse matrix of zeros where each row $m$ contains a ``1'' for the column index corresponding to the sampled observation $m$.  Returning to simplest version of KRLS (Equation~\ref{eq:hh_krls}), Equation~\ref{eq:sketched} shows the sketched version. $\bm{\alpha}_S$ denotes a $M \times 1$ vector of coefficients for the \emph{sketched} kernel, where $\bm{k}^*_i$ is the $i$-th row of $\bm{K}^*$. The analogue for more complex models is straightforward.

\begin{equation}
\label{eq:sketched}
\hat{\bm{\alpha}}_S = \argmax_{\bm{\alpha}_S}\left\{-\frac{1}{2\sigma^2} \left[\sum_{i=1}^N (y_i - \left[\bm{k}^*_i\right]^T\bm{\alpha}_S)^2 + \lambda \bm{\alpha}_S^T \bm{P} \bm{\alpha}_S\right]\right\}; \quad \bm{P} = \bm{S} \bm{K} \bm{S}^T
\end{equation}

\subsection{Calibrating the Sketched Kernel}

We note two key points to consider when using sub-sampling sketching. First, the sketching dimension $M$ clearly affects performance. As $M$ increases, the model will likely perform better (see Appendix~\ref{app:sketch_sims}). Inspired by some literature on the Laplace approximation for standard hierarchical models (e.g., \citealt{shun1995laplace}), the default setting in our software sets $M = \delta N^{1/3}$, i.e. growing at a rate of $N^{1/3}$ times a (constant) sketching multiplier $\delta$; this can be manually increased by the researcher as appropriate.

We show that $\delta = 5$ often provides good performance, but one could use a larger multiplier such as $\delta = 15$ if feasible. The sub-sampling sketching method can be used on very large datasets with this slowly growing $M$; for example, if $N = 100,000$, then $M = 232$ with a multiplier of five and $M = 696$ with a multiplier of fifteen. Section~\ref{sec:validate_gKRLS} shows both can be fit quite rapidly.

Even if $M$ is relatively large, the sub-sampling sketching method may sometimes fail to provide a good representation of the original data (\citealt{yang2017randomized}). We also find some evidence of this when the kernel is complex (see Appendix~\ref{app:sketch_sims}). \cite{lee2020econometric} review the literature on how to improve these methods; future research could explore these techniques.

Second, sub-sampling sketching will not generate identical estimates if the model is re-estimated due to different sketching matrices. While this randomness is common to some statistical methods (e.g. random forests), researchers should carefully examine the sensitivity of their results to the specific sketching matrix chosen. Exactly characterizing the impact of this variability is outside of the scope of this paper, although it may often be relatively small especially when $\delta = 15$. Appendix~\ref{app:sketch_sims} and Appendix~\ref{app:gulzar} examine this for our simulations and applied examples. Corroborating the above discussion about potential limitations of the sub-sampling sketching method, we find that when the kernel is relatively simple, there is a high degree of stability. When the kernel is complex, a larger multiplier may be needed to ensure stable estimates. Assuming it is computationally feasible, a researcher might fit the model multiple times with different sketching matrices to show robustness. If the quantity of interest seems to vary considerably, we suggest increasing the size of the sketching dimension.

\section{Evaluating the Performance of Generalized KRLS}
\label{sec:validate_gKRLS}

We evaluate the scalability of \texttt{gKRLS} when performing the tasks used in standard applications: estimating the model, calculating average marginal effects, and generating predictions on a new dataset of the same size as the training data. We compare \texttt{gKRLS} against popular existing implementations: KRLS (\citealt{hainmueller2014kernel}) and bigKRLS (\citealt{mohanty2019messy})---where we examine truncating the eigenvalues to speed estimation (``bigKRLS (T)'' using a truncation threshold of 0.001) and not doing so (``bigKRLS (NT)'').\footnote{KRLS also can truncate eigenvalues and returns nearly identical results to bigKRLS.} Finally, to examine the role of the sketching multiplier, we fit \texttt{gKRLS} with $\delta \in \{5, 15\}$ [``gKRLS (5)'' and ``gKRLS (15)'', respectively]. All numerical results in this paper are run on a single core with 8GB of RAM. We explore a range of sample sizes spaced from 100 to 1,000,000---spaced evenly on the log-10 scale. For this initial examination, we rely on a generative model from \cite{hainmueller2014kernel} (``Three Hills, Three Valleys'') shown below:

\begin{equation}
y_i \sim N(\mu_i, 0.25); \quad \mu_i = \sin(x_{i,1}) \cdot \cos(x_{i,2})
\end{equation}

\begin{figure}[!htbp]
\begin{center}
\caption{Comparison of Running Time for Different Models}
\label{fig:run_time}
\includegraphics[width=\textwidth]{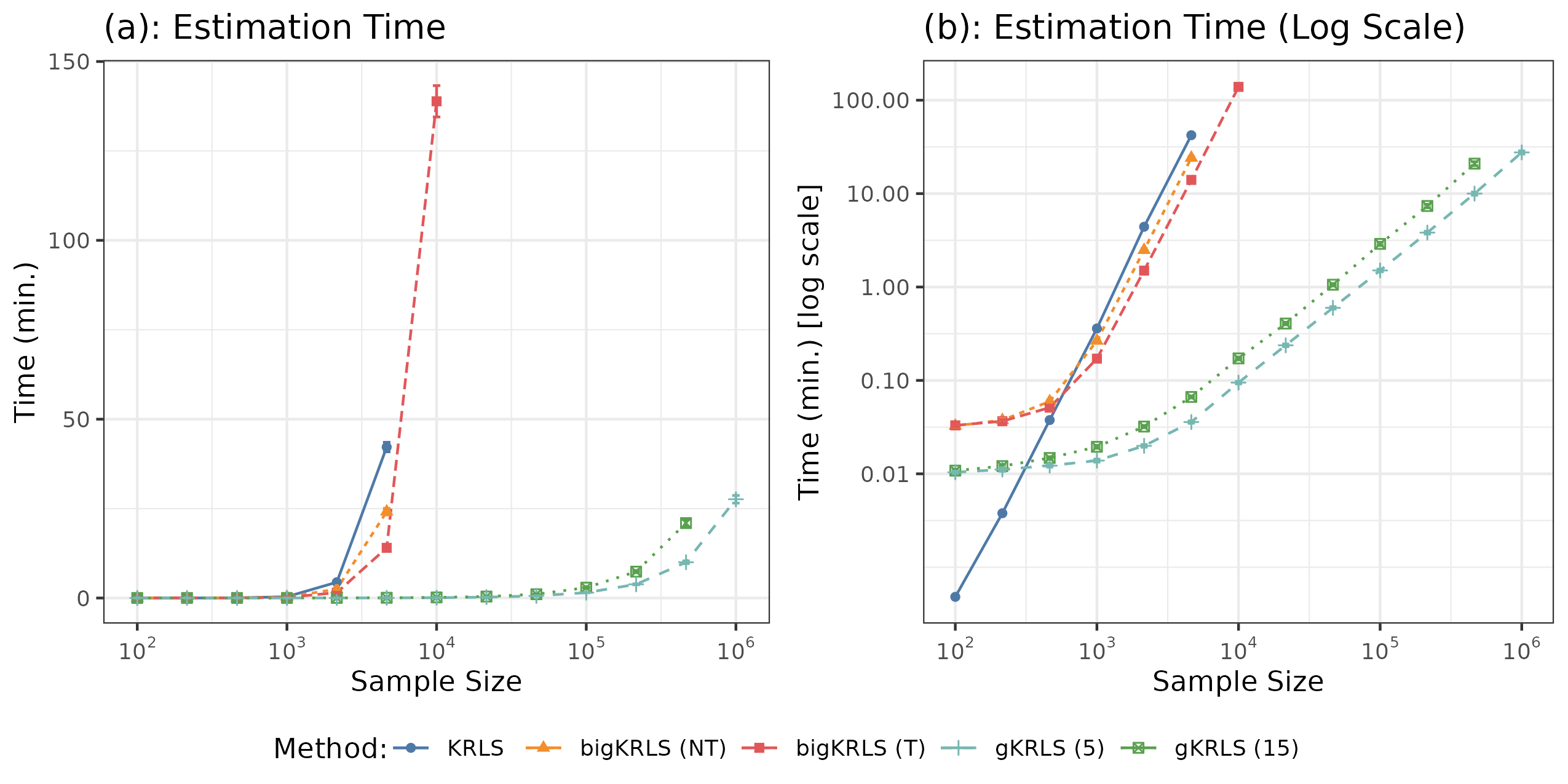}
\end{center}
\caption*{\footnotesize \emph{Note:} 
This figure shows the average computational time in minutes averaged across simulations with 95\% confidence intervals. Figure~\ref{fig:run_time}a presents average time in minutes. Figure~\ref{fig:run_time}b uses a logarithmic scale.}
\end{figure}

We generate fifty datasets and calculate the average estimation time and accuracy across the simulations. We stop estimating methods once costs increase dramatically to limit computational burden. Figure~\ref{fig:run_time} reports the estimation time: KRLS and bigKRLS (with truncation) can be estimated quickly when the number of observations is relatively small, but this increases rapidly as the sample size grows (around the rate of $N^3$). When there are more than 10,000 observations, even bigKRLS would take hours to estimate. By contrast, \texttt{gKRLS} is at least an order of magnitude faster.

The right panel of Figure~\ref{fig:run_time} illustrates this more starkly by reporting the logarithm of time on the vertical axis. Even with the large multiplier (``gKRLS (15)''), \texttt{gKRLS} takes a few minutes for 100,000 observations. Appendix~\ref{app:alt_sketch} calculates an empirical estimate of the computational complexity of \texttt{gKRLS} and shows it is substantially lower than traditional methods. Even with one million observations, \texttt{gKRLS} ($\delta=5$) takes under one hour. Appendix~\ref{app:gam_bam} discusses an alternative estimation technique (\texttt{bam}) that decreases this time to around three minutes with no decline in performance.

\begin{figure}[!htbp]
\begin{center}
\caption{Performance on Out of Sample Predictions}
\label{fig:perform}
\includegraphics[width=\textwidth]{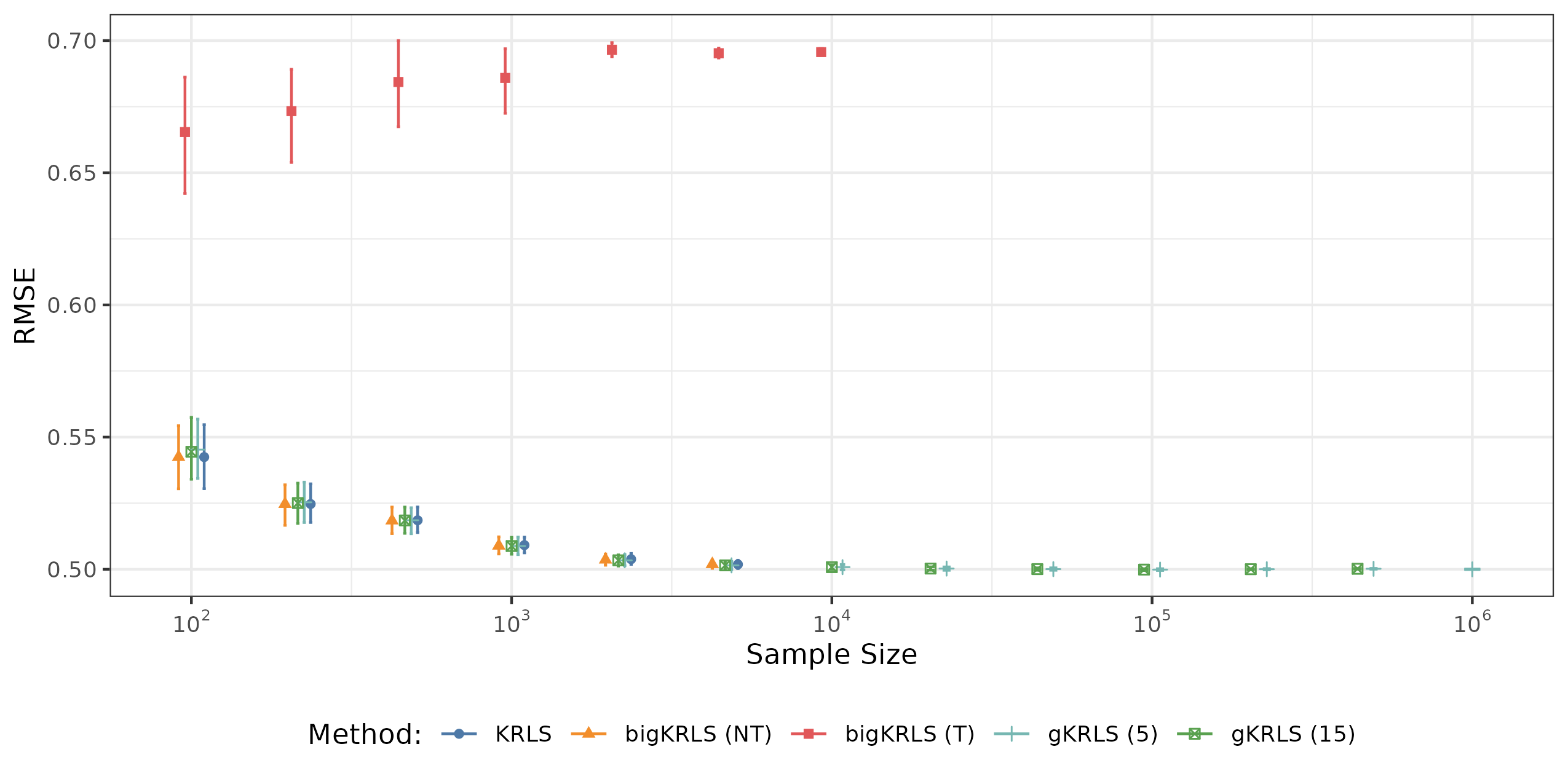}    
\end{center}
\caption*{\footnotesize \emph{Note:} This figure shows the RMSE of predicting the outcome, averaged across fifty simulations. 95\% confidence intervals using a percentile bootstrap (1,000 bootstrap samples) are shown.}
\end{figure}

Figure~\ref{fig:perform} demonstrates that sketching does not come at a material expense of performance in this simple case. We assess the out-of-sample predictive accuracy by generating a 
test dataset of equivalent size to the training data and report the root mean squared error (RMSE) of the predicted values. With the exception of bigKRLS with truncation (``bigKRLS(T)'') that performs considerably worse, Figure~\ref{fig:perform} shows all that methods have similar performance. Appendix~\ref{app:define_perf} examines the error on estimating the average marginal effect; it shows similarly equivalent performance.

\subsection{Kernels and Fixed Effects}
\label{sec:sim2main}

Traditional KRLS usually requires that one include all covariates in a single kernel. This has the benefit of allowing the marginal effect of each variable to depend on all others. However, this could be \emph{too} flexible and require enormous amounts of data to reliably learn the underlying relationship. This problem is likely especially severe when considering fixed effects for group membership. Allowing all marginal effects to vary by group (e.g., a non-linear analogue to interacting group indicators with all covariates) is often too flexible given the potentially limited data in each group. 

However, if one has theoretical reason to believe parts of the underlying model are additive (e.g., including fixed effects to address [additive] unobserved confounding), then including indicators for group \emph{outside} the kernel (i.e., in $\bm{\beta}$) will likely improve performance for modestly sized datasets. Since the group indicators are unregularized, this ensures that the usual ``within-group’’ and ``de-meaning’’ interpretation associated with fixed effects holds; this would not occur if they were included in the kernel.

We use a simulation environment that mimics traditional explorations of fixed effects (e.g., \citealt{bell2015explaining}) but where the functional form of two continuous covariates is possibly non-linear. One of these covariates ($x_{i,1}$) is correlated with the fixed effects and thus its estimation should be more challenging as the correlation increases. The data generating process is shown below.

\begin{mdframed}
\begin{itemize}
    \item Assume there are $J$ groups with some number of observations $T$.
    \item Define $f_{\mathrm{linear}}(x_1, x_2) = 0.5 x_1 + 0.2 x_2$. Define $f_{\mathrm{nonlinear}}(x_1,x_2)$ as follows, following Table 3 in \citet{hainmueller2014kernel}.
    \begin{equation*}
    \label{eq:thin-spline}
    \begin{split}
    f_{\mathrm{nonlinear}}(x_1, x_2) = &\exp\left(\frac{-(x_{1}-0.15)^{2}-(x_{2}-0.15)^{2}}{4}\right) +  \\& 2.5 \cdot \exp\left(\frac{-(x_{1}-0.5)^{2}-(x_{2}-0.5)^{2}}{2.5}\right)
    \end{split}    
    \end{equation*}
    \item Assign each observation $i$ to some group $j$ at random. 
    \item Generate the covariates for each observation as follows. First, draw a fixed effect $\mu_j$ and a group level mean $\bar{x}_j$ for each group. $\rho$ controls the amount of correlation. Larger $\rho$ implies ``random effects'' should perform less well. 

    $$\left[\begin{array}{l}\mu_j \\ \bar{x}_j \end{array}\right] \sim N\left(\left[\begin{array}{l} 0 \\ 0 \end{array}\right], \left[\begin{array}{ll} 3 & \rho \\ \rho & 0.3 \end{array}\right]\right); \quad x_{i,1} \sim N(\bar{x}_{j[i]}, 1); \quad x_{i,2} \sim N(0, 1)$$
    \item Generate the outcome as follows for each $m \in \{\mathrm{linear}, \mathrm{nonlinear}\}$: 
    
    $$y_i = f_{m}(x_{i,1}, x_{i,2}) + \mu_{j[i]} + \epsilon_i; \quad \epsilon_i \sim N(0,1.25)$$
\end{itemize}
\end{mdframed}

\begin{figure}[!htbp]
\begin{center}
\caption{Performance for Average Marginal Effect}
\label{fig:sim2}
\includegraphics[width=\textwidth]{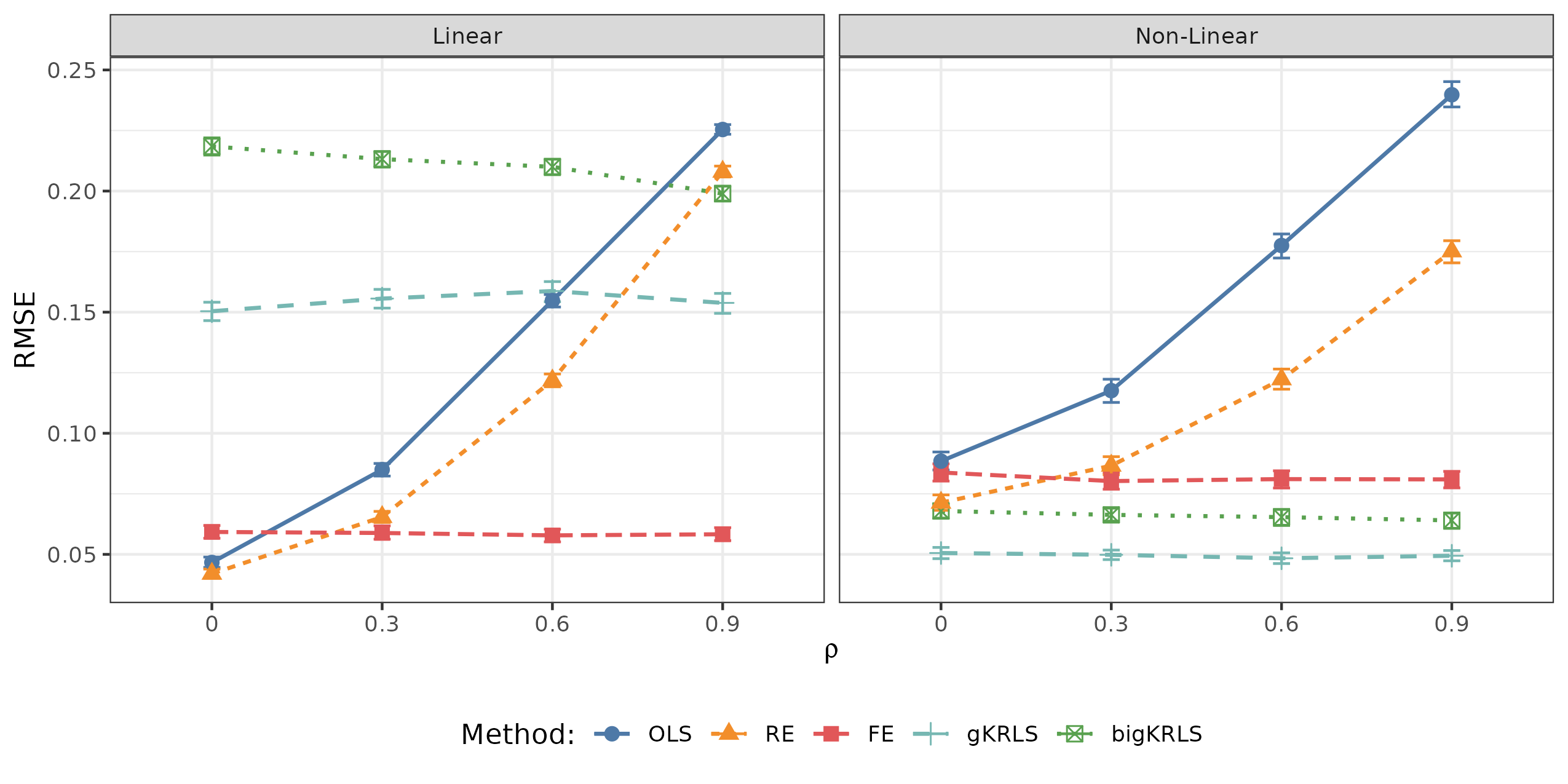}    
\end{center}
\caption*{\footnotesize \emph{Note:} The figure reports the RMSE of the estimated average marginal effect on $x_{i,1}$ as $\rho$ varies. Each panel shows a different data generating process (linear or non-linear). 95\% confidence intervals using a percentile bootstrap (1,000 bootstrap samples) are shown.}
\end{figure}

In our analysis, we set $\rho \in \{0, 0.3, 0.6, 0.9\}$ to vary the degree of correlation between $x_{i,1}$ and $\mu_j$.\footnote{The ``true $R^2$'' (i.e., the $R^2$ of a model that knew the true function) are similar to those in \citet{hainmueller2014kernel}, on average falling between 0.35 and 0.50.} We assume a reasonable number of groups ($J = 50$) and ten observations per group ($T=10$). We compare the following models: (linear) OLS, fixed and random effect models. We also examine two kernel methods: bigKRLS (without truncation; with all variables in the kernel) and \texttt{gKRLS} (with a multiplier of five). For \texttt{gKRLS}, we use a kernel on $x_{i,1}$ and $x_{i,2}$ and include indicators for group membership outside the kernel as unregularized fixed effects ($\bm{\beta}$). We run each simulation 1,000 times.  We expect that all kernel methods should incur some penalty versus linear fixed effects when the true data generating process is linear. Figure~\ref{fig:sim2} reports the RMSE of estimating the average marginal effect (following \citealt{hainmueller2014kernel}) on the correlated covariate $x_{i,1}$.

First considering the linear data generating process (left panel), the traditional estimators (OLS, random effects, and fixed effects) behave as expected: OLS and random effects perform increasingly poorly as $\rho$ increases. In the non-linear data generating process, the same pattern holds although all three linear models perform less well as they are not able to capture the true underlying non-linearity. 

When we compare the kernel methods used in the linear data generating process, both perform worse than fixed effects---i.e. a correctly specified model---and neither method is affected much by $\rho$. However, \texttt{gKRLS} consistently outperforms bigKRLS by a considerable margin. In the non-linear case, we see that both kernel methods perform well versus the linear alternatives, although \texttt{gKRLS} still has a considerable and constant advantage over bigKRLS. Appendix~\ref{app:alt_flavors} shows that including the two covariates as fixed effects ($\bm{\beta}$) in addition to their inclusion in the kernel improves performance considerably on the linear data generating process but incurs some penalty for the non-linear case.

Appendix~\ref{app:extra_sims} provides additional simulations. Appendix~\ref{app:alt_metrics} considers alternative metrics for assessing the performance of the methods, e.g., out of sample predictive accuracy. The results show a similar story: \texttt{gKRLS} is either close to bigKRLS or beats it by a considerable margin. Appendix~\ref{app:alt_control} also explores the performance on estimating the effect of the second covariate ($x_{i,2}$): \texttt{gKRLS} outperforms bigKRLS. Appendix~\ref{app:vary_sample_size} considers an increasing number of observations per group ($T$). As $T$ grows, both kernel methods improve---although \texttt{gKRLS} continues to perform better even when $T=50$. To better understand why \texttt{gKRLS} improves upon bigKRLS, Appendix~\ref{app:alt_flavors} shows that the improvement can be attributed solely to including the fixed effects \emph{outside} the kernel---not additional changes such as how the smoothing parameter is selected, using Mahalanobis distance for creating the kernel, or sub-sampling sketching.

Finally, Appendix~\ref{app:sketch_sims} explores the impact of sketching in this more complex case. It estimates models with different sketching matrices for fixed multiplier $\delta$ to understand the impact on the RMSE versus the unsketched estimates. It finds that sketching incurs some penalty on the accuracy of the estimated average marginal effect, although this declines as the sketching multiplier increases. When fixed effects are included in the kernel, this decline is considerably slower. When fixed effects are not included in the kernel, virtually any sketching multiplier can recover nearly identically accurate estimates to the corresponding unsketched procedure.

\section{Generalized KRLS for Observational Data}
\label{sec:newman}

Our first empirical application examines an observational study by \cite{newman2016breaking}. The paper focuses on the contextual effects of gender-based earnings inequality for women's belief in meritocracy. The key theoretical discussion concerns how gender inequality in earnings in the local area where a woman lives affects their rejection of a belief in meritocracy (e.g., ``hard work and determination are no guarantee of success for most people''). \citet[p. 1009-111]{newman2016breaking} compares a number of theoretical perspectives: Some (e.g., relative deprivation theory) suggest that women in areas with more economic inequality between men and women should show more rejection of meritocracy. However, \cite{newman2016breaking}'s preferred theoretical expectation, drawing on literature on ``glass ceilings'' and rising expectations theory, suggests a non-linear effect: Rejection of meritocracy should be highest when women have come close to---but not quite achieved---economic parity as they have experienced large gains but still have failed to achieve equality. Once parity is achieved, the rejection of meritocracy should fall. Specifically, \citet[p. 1011]{newman2016breaking} expects a ``nonlinear, concave quadratic effect of local gender-based earnings inequality on women's likelihood of rejecting meritocracy.'' \cite{newman2016breaking} tests this using hierarchical logistic regressions where the key variable (earnings inequality, operationalized as the ratio of female median income to male median income at the county of residence) is included quadratically. 

\texttt{gKRLS}'s modularity allows us to more robustly test \cite{newman2016breaking}'s argument. Our first hierarchical term is a kernel including all covariates to capture possible interactions or non-linearities omitted by the original (additive) model and thereby improve the robustness of the reported results. We also include a random intercept for county, following \cite{newman2016breaking}, to address the nested nature of the data. 

However, Section~\ref{sec:validate_gKRLS} illustrated that relying exclusively on \texttt{gKRLS} given limited data may be undesirable as it could be too flexible. An additional risk of relying exclusively on KRLS is that if the estimated $\lambda$ were very large, that would effectively exclude all covariates and mimic an intercept-only model. A more modular approach uses a KRLS term to flexibly estimate interactions or non-linear effects while additionally including ``primary'' covariates of interest.

We include additional terms following \cite{newman2016breaking}. First, we include all controls in the fixed effects ($\bm{\beta}$). Second, we perform a more robust examination of the effect of earnings inequality. Rather than assuming the relationship is quadratic, we additively include a thin plate regression spline \citep[p. 216]{wood2017mgcv} on earnings inequality. This does not impose a specific functional form and allows the data to \emph{reveal} whether the relationship is quadratic or has some other shape. This expanded model ensures that we include a specification that is comparable to \cite{newman2016breaking} while also allowing for extra interactions using KRLS.

Overall, we estimate a logistic regression with four parts ($J=3$): (i) a KRLS term including all controls and earnings inequality, (ii) a random effect for county; (iii) a spline on earnings inequality; and (iv)  twenty-four controls entered in linearly and unpenalized (in $\bm{\beta}$). The three tuning parameters (separate $\lambda_j$ for [i], [ii], and [iii]) are estimated using REML.

\begin{figure}[!htbp]
    \caption{Re-Analysis of \cite{newman2016breaking}}
    \label{fig:newman}
    \includegraphics[width=\textwidth]{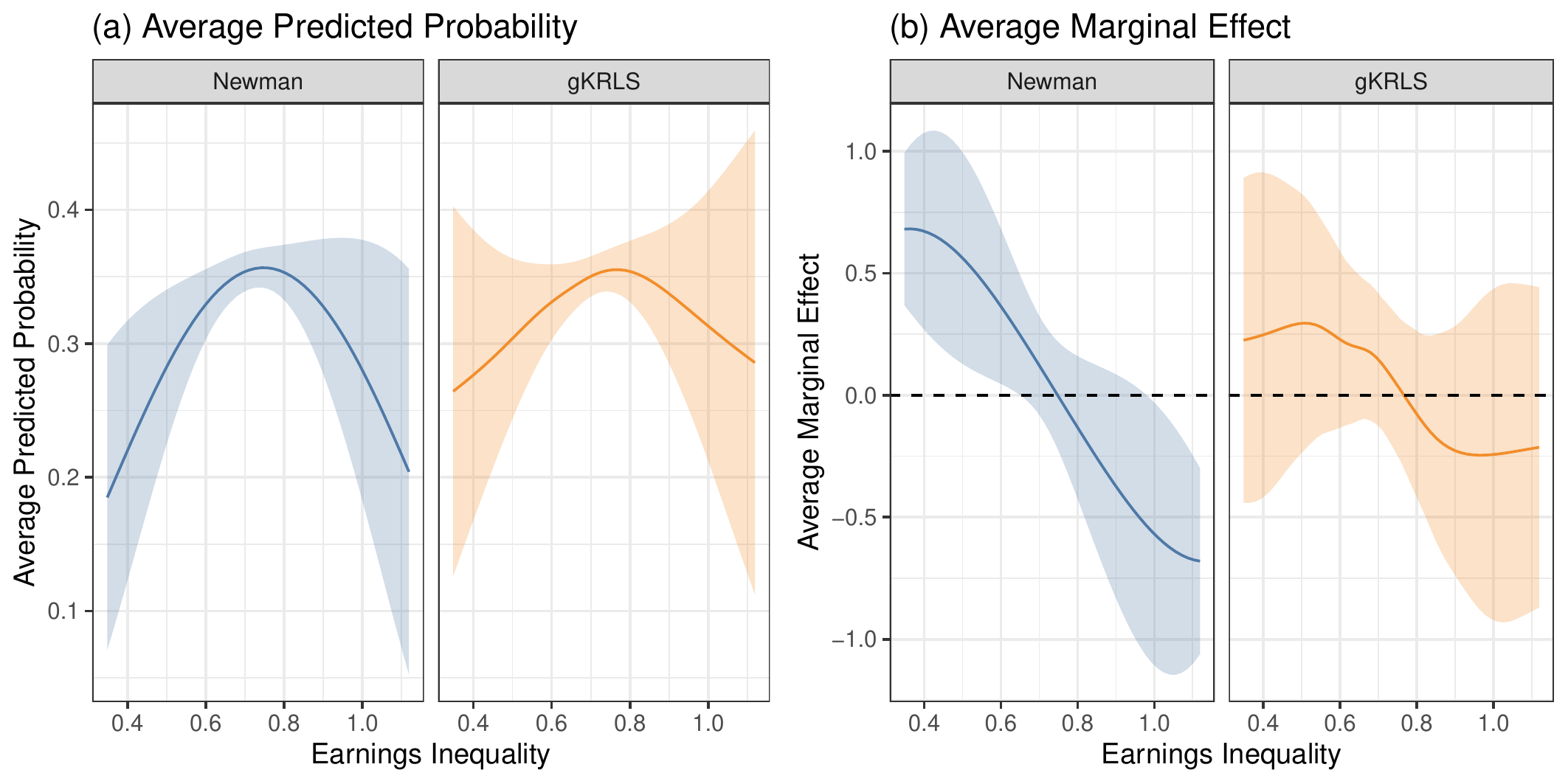}
    \caption*{\footnotesize \emph{Note:} The average predicted probability and average marginal effect with 95\% confidence intervals are shown.}
\end{figure}

Figure~\ref{fig:newman} shows (a) the average predicted probability of rejecting meritocracy and (b) the average marginal effect across a grid of earning inequality values from  the lowest to the highest value in the data---following \cite{newman2016breaking}. Appendix~\ref{app:newman} provides the question wording and definition of these quantities. Figure~\ref{fig:newman} reports the original specification in \cite{newman2016breaking} as well as \texttt{gKRLS}.

The results partially support \cite{newman2016breaking}. The point estimates from \texttt{gKRLS} show a non-linear inverted ``u-shaped'' relationship that is similar to the original results (``Newman''), although the curve is noticeably flatter for extreme values of earnings inequality. This occurs because \texttt{gKRLS} estimates relatively constant average marginal effects at extreme values of earnings inequality versus the mechanically increasing or decreasing values assumed by a quadratic specification.

When considering estimated uncertainty, however, we note that the 95\% confidence intervals for the marginal effect from \texttt{gKRLS} cross zero at all points---unlike the original model. Appendix~\ref{app:newman} provides additional tests (e.g., average second derivative, difference in the average marginal effects at the extreme values) that show the same result (confidence intervals that contain zero for \texttt{gKRLS}). Thus, despite similar point estimates, relaxing the strong functional form assumptions in \cite{newman2016breaking} returns limited evidence for a statistically detectable non-linear relationship. Appendix~\ref{app:newman} corroborates this with other examples from the original paper: Using five other questions (binary and ordered logistic regressions), \texttt{gKRLS} generally finds an inverted ``u-shaped'' in the point estimates but little evidence of a statistically detectable non-linear relationship.

\section{Generalized KRLS with Machine Learning}
\label{sec:main_gulzar}

Our second empirical replication considers a geographic regression discontinuity analysis in \cite{gulzar2020does}. They focus on the effects of improving political representation using quotas on the economic welfare of various groups in society. They examine how electoral quotas for members of Scheduled Tribes affect the economic welfare of members of that group, members of a different historically disadvantaged group \emph{not} affected by the quota (members of Scheduled Castes), members in neither group (``Non-Minorities''), as well as the total population.

We focus on their analysis of three economic outcome variables from the National Rural Employment Guarantee Scheme that offers one hundred days of employment for rural households \citep[p. 1231]{gulzar2020does}. The outcomes we consider are ``(log) jobcards'' (the total number of  documents issued to prospective workers under the program), ``(log) households'' (the number of households who participated in the program), and ``(log) workdays'' (the total number of days worked by individuals in the program). The treatment is whether a village is part of a scheduled area that imposes an electoral quota. Across the three outcomes, the key findings from \cite{gulzar2020does} are that (i) there is no effect on the total economic welfare, (ii) the targeted minorities (Scheduled Tribes) see increases in economic welfare; (iii) the non-targeted minority groups (Scheduled Castes) do not see any significant changes;  and (iv) non-minority groups see decreases in economic outcomes.

\begin{figure}[!htbp]
    \caption{Effects of Electoral Quotas}
    \label{fig:maineffect}
    \centering
    \includegraphics[width=\textwidth]{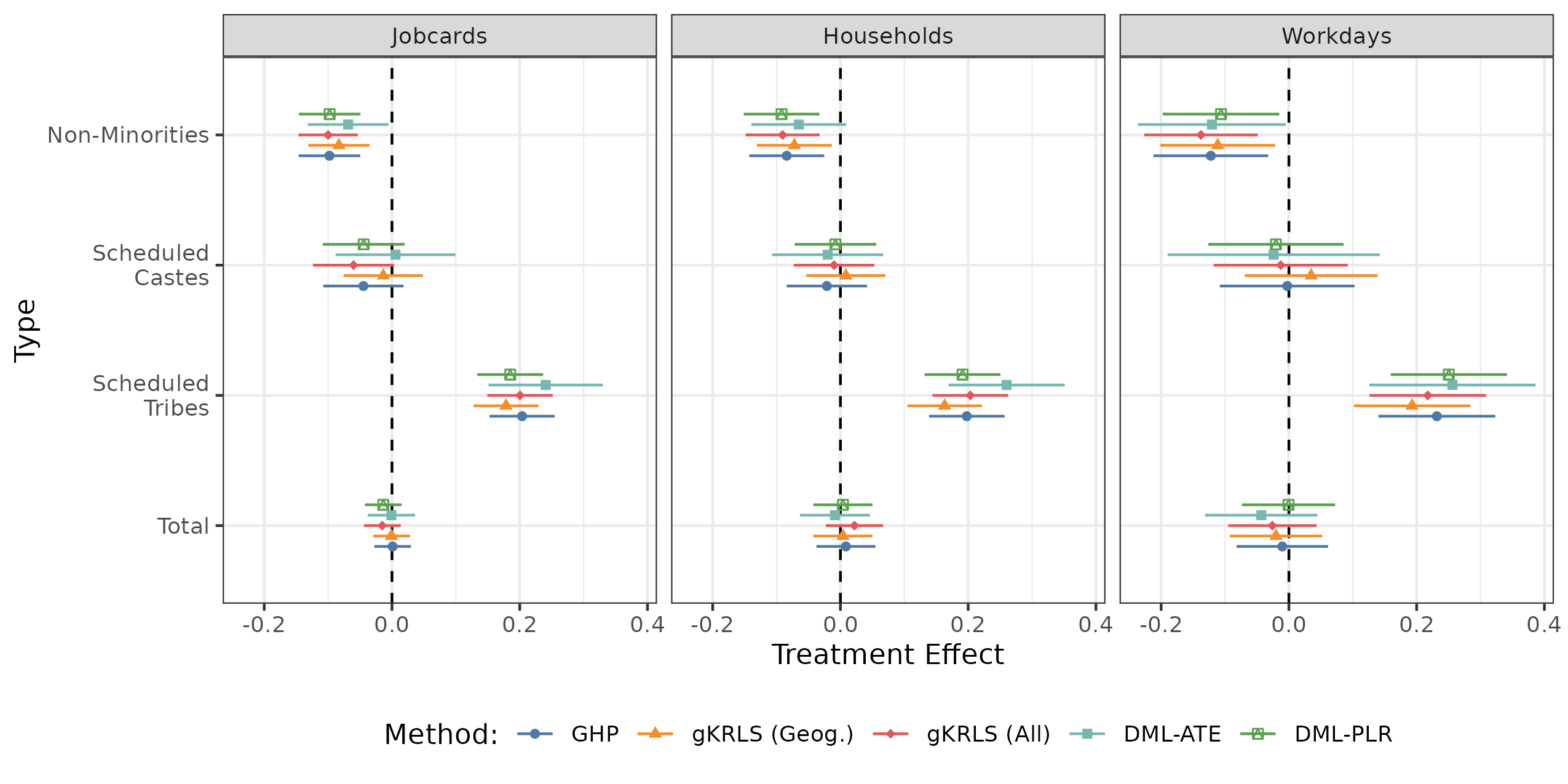}
    \caption*{\footnotesize \emph{Note}: This figure reports estimated treatment effects for all groups and outcomes. 95\% confidence intervals are shown.}
\end{figure}

\texttt{gKRLS} can improve the original analysis in two ways. First, \cite{gulzar2020does} include the interaction of fourth-order polynomials on latitude and longitude following previous work on geographic regression discontinuity designs (replicated as ``GHP'' in Figure~\ref{fig:maineffect}). \texttt{gKRLS} enables a more flexible solution, even on this larger dataset (32,461 observations), by using a kernel on the geographic coordinates\footnote{In this specification only, we rely on raw Euclidean distance, without standardization, due to the direct meaning of geographic distance.} while including the treatment and other covariates linearly as unpenalized terms ($\bm{\beta}$). We denote this model as ``gKRLS (Geog.)’’. Second, \citet[p. 1238]{gulzar2020does} report some imbalance on certain pre-treatment covariates; they include controls additively and linearly to improve the robustness of their results. Including these variables (and treatment) in a KRLS term provides additional robustness. We use ``gKRLS (All)’’ for this model that includes the KRLS term ($J=1$) as well as all variables linearly in the fixed effects ($\bm{\beta}$) to ensure their inclusion. In both models, we use cluster-robust standard errors following the original specification.

The use of penalized terms, however, raises a concern about regularization bias in the estimated treatment effect; we address this using double/debiased machine learning (DML) that removes such bias (\citealt{chernozhukov2018dml}).  We use the specification from ``gKRLS (All)'' (after removing the treatment indicator) for our machine learning model. Estimation with five folds requires fitting \texttt{gKRLS} ten or fifteen times depending on whether one uses the partially linear model (``DML-PLR'') or the dedicated algorithm for estimating the ATE (``DML-ATE''), respectively.\footnote{Following \cite{chernozhukov2018dml}, we trim the estimated propensity scores at 0.01 and 0.99.} Both procedures estimate conditional expectation functions with Gaussian outcomes, while the latter (DML-ATE) also estimates a propensity score for being treated using a binomial outcome with a logistic link. Either procedure takes only a few minutes to estimate. To address clustering within the data, we use stratified sampling to create the folds for DML and produce the standard error on the treatment effect using an analogue to the usual cluster-robust estimator (\citealt{chiang2022multiway}).

Figure~\ref{fig:maineffect} presents the results. The results are generally robust regardless of the specification chosen. The one exception is DML-ATE that has consistently larger standard errors (around 40\% greater than other specifications) and somewhat larger point estimates for effects on Scheduled Tribes across two outcome variables.

Appendix~\ref{app:gulzar} provides additional analyses. Appendix~\ref{app:gulzar_main} repeats the analysis fifty times to examine variability across different sketching matrices. It finds relatively low variability of the point estimates relative to the magnitude of the estimated standard errors. Appendix~\ref{app:heteff_gulzar} uses \texttt{gKRLS} with a machine learning algorithm to estimate heterogeneous treatment effects (``R-learner''; \citealt{nie2021quasi}). Even though this method requires fitting \texttt{gKRLS} over a dozen times (with both Gaussian and binomial outcomes), estimation takes only a few minutes. We find that one state (Himachal Pradesh) has noticeably larger treatment effects than other states.

\section{Conclusion}

Our paper generalized KRLS in two meaningful directions by drawing together different existing literatures. First, we recast the original model into the modular framework of hierarchical and generalized additive models where adding a kernel on some variables can be thought of as simply adding one additional hierarchical term (i.e., increasing $J$ by one). This allows researchers using \texttt{gKRLS} to modularly build their model by including variables in different ways based on their substantive knowledge. For models with multiple hierarchical terms and/or non-Gaussian outcomes, a hierarchical perspective on KRLS allows for easy tuning of the regularization parameters, efficient estimation, and well-calibrated standard errors. Empirically, we show that in a stylized example with additive fixed effects, thinking carefully about how to include different terms in the model (e.g., unregularized fixed effects versus including them in the kernel) can be critically important to performance. The second generalization employed sub-sampling sketching to allow \texttt{gKRLS} to be easily scalable to most datasets encountered in social science. By breaking the requirement that the cost of the model depends on the cube of the number of observations, sub-sampling sketching allows the model to be estimated very quickly on tens or hundreds of thousands of observations. Even for methods that require repeated estimation of \texttt{gKRLS} (e.g.,  double/debiased machine learning), models can be estimated with limited computational cost. Our paper and accompanying software therefore allows KRLS to become a more widely used part of the applied researcher's toolkit.

\bibliography{reference}
\bibliographystyle{mod_apsr}

\clearpage

\appendix

\newcommand{\argmin}{\operatornamewithlimits{argmin}}

\renewcommand{\thefigure}{A.\arabic{figure}}
\renewcommand{\thetable}{A.\arabic{table}}
\renewcommand{\theequation}{A.\arabic{equation}}
\setcounter{figure}{0}
\setcounter{table}{0}
\setcounter{equation}{0}

\begin{center}
	\textbf{\Large Supplementary Material for ``Generalized Kernel Regularized Least Squares''}
\end{center}

\setcounter{page}{1}

\section{Additional Theoretical Results}

This section contains additional theoretical results. First, we provide a more detailed exposition of the Gaussian model. It unifies our presentation in the main text with more classical presentations of hierarchical/multilevel models (e.g., \citealt{hazlett2022mlm}). It also allows us to derive our justification for robust/clustered standard errors as well as providing new standard errors for (traditional) KRLS. Second, we discuss the alternative estimation methods available in \texttt{mgcv} (\texttt{gam} vs \texttt{bam}). Finally, we explain how our software estimates average marginal effects.

\subsection{Alternative Standard Errors}
\label{app:robust_SE}

We consider the following Gaussian outcome model for our discussion on standard errors following \cite{hazlett2022mlm}. Extensions to non-Gaussian outcomes can be made using standard arguments (see \citealt{wood2006ci}). We do not assume any special structure on $\bm{\Omega}$ but are more general than the main text insofar as $\bm{\Sigma}$ need not be diagonal. We assume that $\bm{\Omega}$ and $\bm{\Sigma}$ are known and full rank. A rank-deficient $\bm{\Omega}$ can be addressed via an eigen-decomposition and adjusting the fixed effect design matrix $\bm{X}$. $\bm{\Sigma} = \sigma^2 \bm{I}$ and $\bm{\Omega} = 1/\sigma^2 \bm{S}_{\bm{\lambda}}$ recovers the model in the main text (Equation~\ref{eq:logposterior_multi}).

\begin{align}
\bm{y} &\sim N\left(\bm{X}\bm{\beta} + \bm{Z}\bm{\alpha}, \bm{\Sigma}\right) \\
p(\bm{\beta}) &\propto 1; \quad \bm{\alpha} \sim N(\bm{0}, \bm{\Omega}^{-1})
\end{align}	

The penalized maximum likelihood estimator is shown below, mirroring Equation~\ref{eq:pen_estimates}.

\begin{align}
\left[\begin{array}{l} \hat{\bm{\beta}} \\ \hat{\bm{\alpha}} \end{array}\right] = \left[\begin{array}{ll} \bm{X}^T\bm{\Sigma}^{-1}\bm{X} & \bm{X}^T\bm{\Sigma}^{-1}\bm{Z} \\ \bm{Z}^T\bm{\Sigma}^{-1}\bm{X} & \bm{Z}^T\bm{\Sigma}^{-1}\bm{Z} + \bm{\Omega}\end{array}\right]^{-1} \left[\begin{array}{l} \bm{X}^T \\ \bm{Z}^T\end{array} \right]\bm{\Sigma}^{-1}\bm{y}
\end{align}

Despite our different presentation, it can be easily verified (see also \citealt[p.80-81]{wood2017mgcv}) that $\hat{\bm{\beta}}$ and $\hat{\bm{\alpha}}$ are identical to the standard multilevel estimators (e.g., \citealt[p. 51]{hazlett2022mlm}). In terms of appropriate estimators for the variance of $\hat{\bm{\beta}}$ and $\hat{\bm{\alpha}}$, \cite{wood2006ci} suggests two options; a ``Bayesian'' estimator that is the inverse of the Hessian of the log-posterior and a frequentist estimator. We consider each in detail. First, the Bayesian estimator $\mathcal{V}_B$ is shown below. This is justified based on the posterior distribution of $\{\bm{\beta}, \bm{\alpha}\}$ given $\bm{y}$ if $\bm{\Sigma}$ and $\bm{\Omega}$ are known (\citealt{wood2006ci}).

$$\mathcal{V}_B = \left[\begin{array}{ll} \bm{X}^T\bm{\Sigma}^{-1}\bm{X} & \bm{X}^T\bm{\Sigma}^{-1}\bm{Z} \\ \bm{Z}^T\bm{\Sigma}^{-1}\bm{X} & \bm{Z}^T\bm{\Sigma}^{-1}\bm{Z} + \bm{\Omega}\end{array}\right]^{-1}; \quad \mathcal{V}_B^\beta = \left(\bm{X}^T \left[\bm{\Sigma} + \bm{Z} \bm{\Omega}^{-1} \bm{Z}^T\right]^{-1}\bm{X}\right)^{-1}$$

Note that the upper left-block, corresponding to the estimated variance of $\hat{\bm{\beta}}$, denoted as $\mathcal{V}^\beta_B$ is identical to the ``standard'' multilevel model estimator for $\hat{\bm{\beta}}$ (see \citealt[p. 80]{wood2017mgcv} and \citealt[p. 52]{hazlett2022mlm}). Second, Wood (2006) suggests a ``frequentist'' estimator where, for some choice of variance of $\bm{y}$ denoted as $\mathrm{Var}(\bm{y})$, the estimator $\mathcal{V}_F$ is shown below.

$$ \mathcal{V}_F = \mathcal{V}_B  \left[\begin{array}{l} \bm{X}^T \\ \bm{Z}^T\end{array}\right] \bm{\Sigma}^{-1} \mathrm{Var}(\bm{y}) \bm{\Sigma}^{-1} [\bm{X}, \bm{Z}]\mathcal{V}_B  $$

There are different possible choices for $\mathrm{Var}(\bm{y})$. Following \cite{wood2006ci}, we can assume the model is correct, condition on $\bm{\beta}$ and $\bm{\alpha}$ and thus use $\mathrm{Var}(\bm{y}) = \bm{\Sigma}$. We denote this estimator as $\mathcal{V}^{Wood}_F$. When comparing $\mathcal{V}_B$ and $\mathcal{V}^{Wood}_F$, Wood (2006) suggests that $\mathcal{V}_B$ is preferable insofar as it reflects a coherent Bayesian model and the $\mathcal{V}_F$ estimator is likely to have considerably undercoverage due to the fact that neither $E[\hat{\bm{\beta}}] \neq \bm{\beta}$ nor $E[\hat{\bm{\alpha}}] \neq \bm{\alpha}$. A variety of work has corroborated this suggestion empirically and theoretically (see, amongst others, \citealt[Ch. 6.10]{wood2017mgcv} or \citealt{marra2012coverage}). $\mathcal{V}_B$ has been shown to have usually rather good frequentist coverage despite coming from a regularized model. In general, this is our preferred estimator as it is naturally derived from our Bayesian and hierarchical interpretation of KRLS. However, $\mathcal{V}_F$ is useful insofar as it lends itself to ``robust'' estimators by choosing a specific choice of $\mathrm{Var}(\bm{y})$ that does not hue to the assumptions of the model. For example, if one suspected heteroskedastic errors, $\mathrm{Var}(\bm{y})$ could be swapped with the usual ``meat'' of the squared residuals, possibly scaled by a finite sample correction (\citealt{cameron2015practitioner}). 

This idea is not novel to our paper, although we think it is perhaps underappreciated in the literature: \cite{hazlett2022mlm} consider it in the case of clustered standard errors and a hierarchical model with a single random effect. Our formula, with the same corresponding ``meat'', recovers an identical variance matrix on $\hat{\bm{\beta}}$ to their proposal in Equation 8 (p. 51) with some re-arrangement. However, a benefit of this formulation is that the joint uncertainty on $\hat{\bm{\beta}}$ and $\hat{\bm{\alpha}}$ is quantified so any quantity of interest that includes \emph{both} terms (i.e., most marginal effects and predicted values) can be fully incorporated.

Our justification does not assume any particular structure on $\bm{\Omega}$ or $\bm{\Sigma}$ so applies to generic generalized additive models as other types of robust standard errors (e.g., multiway clustering, Conley standard errors for spatial dependence, etc.; \citealt{cameron2015practitioner}). Following the arguments in \cite{wood2006ci} that generalize $\mathcal{V}_F$ and $\mathcal{V}_B$ to non-Gaussian outcomes, one could also apply our logic to non-Gaussian outcomes. Exploring these alternative standard errors in more detail is an interesting area for future research.

\subsubsection{Implications for Traditional KRLS}

We briefly consider the implications for traditional KRLS. Beyond justifying (cluster) robust standard errors, there is a more subtle point about the appropriate ``regular'' standard errors for KRLS. \citet[p. 153]{hainmueller2014kernel} propose frequentist standard errors, i.e. $\mathcal{V}_F$ assuming $\mathrm{Var}(\bm{y}) = \sigma^2 \bm{I}$. Noting that in their model, $\bm{\beta}$ does not exist (and $J = 1$), $\bm{Z} = \bm{K}$, $\bm{\Omega} = \frac{\lambda}{\sigma^2}\bm{K}$, and $\bm{\Sigma} = \sigma^2 \bm{I}$, the possible variance estimators are shown below where $\mathcal{V}_{HH}$ is the suggestion in the original paper and coincides with $\mathcal{V}^{Wood}_F$.

\begin{equation}
\begin{split}
\mathcal{V}_B &= \sigma^2 \left[\bm{K} \bm{K} + \lambda \bm{K}\right]^{-1} \\
\mathcal{V}_{HH} = \mathcal{V}^{Wood}_F &=  \sigma^2 \left(\bm{K} + \lambda \bm{I}\right)^{-1} \left(\bm{K} + \lambda \bm{I}\right)^{-1}
\end{split}
\end{equation}

While exploring the coverage properties of these estimators in detail is outside of the scope of this paper, we note that the suggested variance-covariance matrix from \cite{hainmueller2014kernel} is thus \emph{not} the standard recommendation for generalized additive models ($\mathcal{V}_B$) and thus may have worse coverage properties than the Bayesian alternative. We examine this in a simple stylized example. We use a bivariate smoothing example from \texttt{mgcv}, shown below. We draw 250 observations indexed with $i \in \{1, \cdots, N\}$.

\footnotesize
\begin{subequations}
\begin{align}
y_i | x_i, z_i &\sim N\left(f(x_i, z_i), 1\right); \quad x_i \sim \mathrm{Unif}(0,1); \quad z_i \sim \mathrm{Unif}(0,1) \\
f(x,z) &=  \pi^{\sigma^2_x}  \sigma^2_z \cdot \left(1.2 \cdot \exp\left[-\frac{(x - 0.2)^2}{\sigma^2_x} - \frac{(z - 0.3)^2}{\sigma^2_z}\right] + 0.8 \cdot \exp\left[-\frac{(x - 0.7)^2}{\sigma^2_x} - \frac{(z - 0.8)^2}{\sigma^2_z}\right]\right)
\end{align}
\end{subequations}
\normalsize

In this case, we compare the coverage over the response surface itself, i.e. over the Cartesian product of a grid of 40 evenly spaced values of $x$ (from 0 to 1) and 40 evenly spaced values of $z$ (from 0 to 1). We repeat the simulation 200 times, i.e. generate 200 sets of $\bm{y}$ and report the average coverage on the predicted values across all simulations. 

We consider \texttt{KRLS} with its default standard errors ($\mathcal{V}_{HH}$), \texttt{KRLS} with $\mathcal{V}_B$, \texttt{gKRLS} with default standard errors ($\mathcal{V}_B$), \texttt{gKRLS} with default standard errors and the two variables $x_i$ and $z_i$ also included linearly \emph{outside} the kernel as fixed effects. This latter model is more comparable to the default splines in \texttt{mgcv} where the null space of the penalty contains an additive linear model. In this specification, as $\lambda \to \infty$, the model approaches a linear additive one in $x$ and $z$ versus an intercept only model for standard (g)KRLS. We finally compare against the tensor smoothing spline suggested by \texttt{gam}. Table~\ref{app:tab_se} shows the results. It reports the size of the average standard error on a predicted value, the coverage averaged across all points and simulations, and the mean absolute error (MAE) in prediction.

\begin{table}[!htbp]
	\begin{center}
	\caption{Performance of Methods for Bivariate Smooth}
	\label{app:tab_se}
	\begin{tabular}{lccc}
		\hline\hline
		Method & Coverage & MAE & Average SE \\
		\hline
		\input{figures/Table_A1}
	\end{tabular}
	\caption*{\footnotesize \emph{Note}: The coverage proportion of a 95\% confidence interval across the entire grid of test values is shown. The mean absolute error (MAE) in prediction and the average standard error for each prediction is shown. These quantities are all averaged across 200 simulations.}	
	\end{center}
\end{table}

It illustrates the expected problem with \texttt{KRLS}'s default standard errors, $\mathcal{V}_{HH}$. While the point estimates are of high quality (better than \texttt{gam}), the coverage is very poor. By contrast, simply replacing the variance estimator with $\mathcal{V}_B$ immediately improves coverage dramatically---albeit remaining below nominal. In terms of \texttt{gKRLS}, we see that it performs equivalently to KRLS in performance---beating \texttt{mgcv}---and has similar coverage to the KRLS with $\mathcal{V}_B$. The model that includes $x_i$ and $z_i$ linearly outside of the kernel (\texttt{gKRLS} + Linear) has slightly worse performance but close to nominal coverage and improves in terms of MAE upon the tensor product smooth (\texttt{gam}).

\begin{figure}[!htbp]
    \caption{Coverage by True Value}
    \label{app:fig_se}
\includegraphics[width=\textwidth]{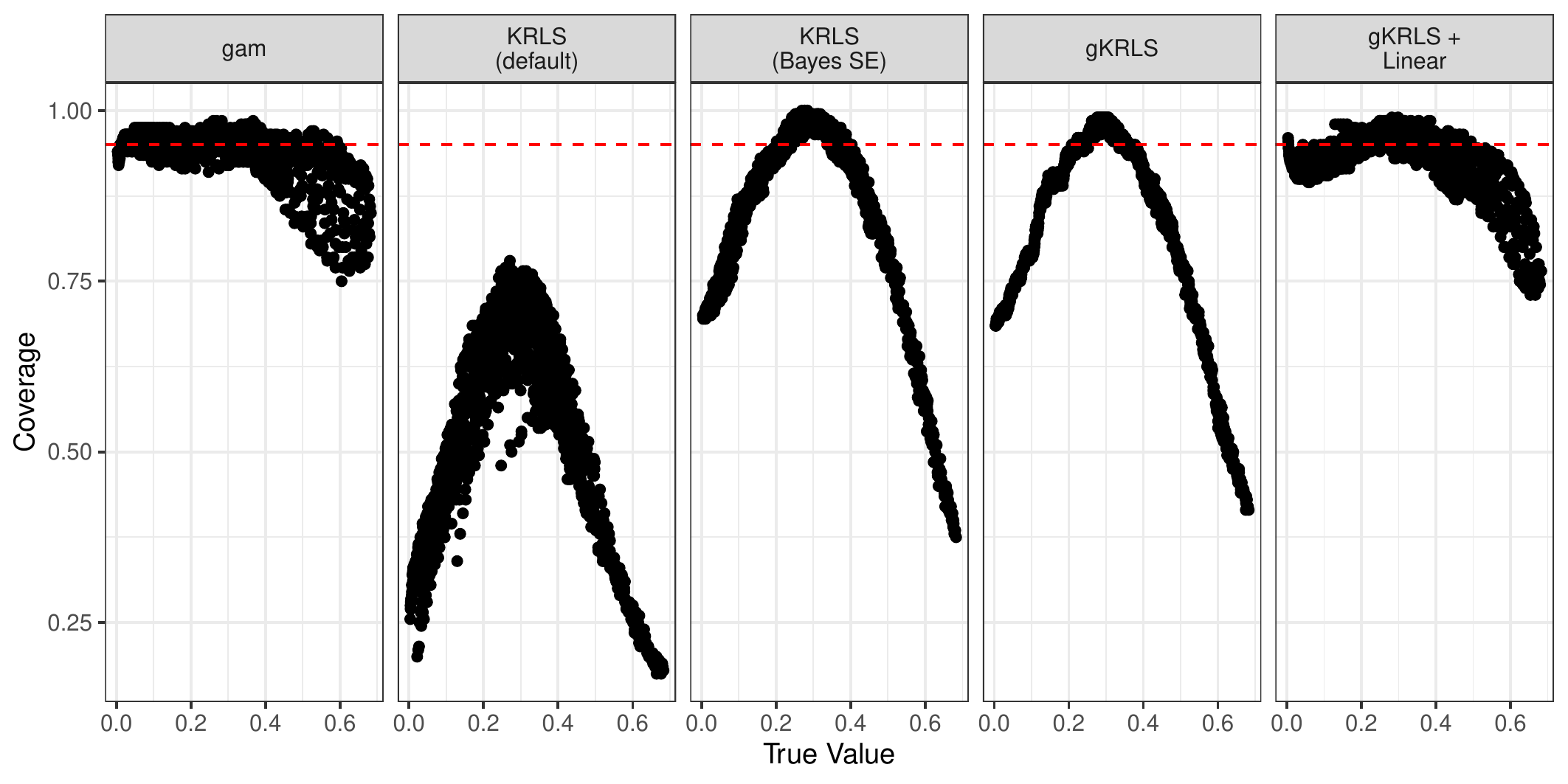} 
    \caption*{\footnotesize \emph{Note}: The coverage proportion for each observation given the 95\% confidence interval on the predicted value, averaged across 200 simulations, is shown. The true fitted value is on the horizontal axis.}
\end{figure}

Figure~\ref{app:fig_se} plots the average coverage for each observation by the true value; it shows that for more extreme fitted values, the coverage of KRLS and \texttt{gKRLS} is increasingly poor. Using the Bayesian standard errors improves KRLS considerably although it is still usually below nominal.

Finally, we use a simple example that illustrates the importance of clustered standard errors. Sticking to the above simulation, we assume that there is a group $g$ that is drawn randomly from one of forty clusters ($g \in \{1, \cdots, 40\}$). For two observations in the same cluster, our generative model assumes they are correlated with $\rho = 0.6$ and are otherwise uncorrelated. We re-estimate the same models from before, but also consider cluster-robust errors for \texttt{gam} and \texttt{gKRLS}. We see from Table \ref{app:tab_cluster_se} that cluster-robust standard errors improve the coverage for the \texttt{gKRLS} and \texttt{mgcv} methods. We have written additional software to allow \texttt{mgcv} methods (\texttt{gam} and \texttt{bam}) to be used with the \texttt{sandwich} package. Without our software, there is a bug (at the time of writing [April 2023] affecting at least versions 3.0.2 and older) that results incorrect robust/clustered standard errors for Gaussian outcomes and certain models with non-canonical links.

\begin{table}[!htbp]
\begin{center}
\caption{Cluster-Robust Errors for Smoothing Methods}
\label{app:tab_cluster_se}
    \begin{tabular}{lccc|ccc}
        \hline\hline
        & \multicolumn{3}{c}{Regular SE} & \multicolumn{3}{c}{Clustered SE} \\
        Method & Coverage & MAE & Average SE & Coverage & MAE & Average SE \\
        \hline
        \input{figures/Table_A2} 
    \end{tabular}
\caption*{\footnotesize \emph{Note}: The same quantities from Table~\ref{app:tab_cluster_se} are shown. The first three columns consider regular standard errors; the final three consider the cluster-robust version.}	
\end{center}
\end{table}

\subsection{Alternative Estimation Methods}
\label{app:gam_bam}

The main text notes that \texttt{mgcv} provides two methods for estimation: \texttt{gam} and \texttt{bam}. \texttt{gam} is designed to be a highly stable numerical procedure. However, it is sometimes slow especially when the number of observations is very large (e.g., the simulations in Appendix~\ref{app:scale_sims}) or when the number of parameters is very large (e.g., the replication of \citealt{newman2016breaking}). The latter may often occur when a random intercept with many levels is included. In these cases, some alternative estimation technique is needed. \cite{wood2015generalized} develop \texttt{bam} to address these limitations. The two main innovations are as follows (see \citealt[p. 290]{wood2017mgcv} for a concise summary): First, \texttt{bam} slightly alters the estimation procedure for $\bm{\lambda}$. In the initial exposition in the main paper, for a proposed $\bm{\lambda}$, $\hat{\bm{\beta}}_{\bm{\lambda}}$ and $\hat{\bm{\alpha}}_{\bm{\lambda}}$ are estimated to convergence---e.g. using penalized iteratively re-weighted least squares (PIRLS) with multiple iterations for a non-Gaussian outcome. This can be expensive so \texttt{bam} optimizes $\bm{\lambda}$ after each step in PIRLS estimation. This is known as ``performance orientated iteration'' and can be less numerically stable than the main algorithm \citep{wood2015generalized}. 

Second, and perhaps more interestingly, \texttt{bam} never forms the entire design matrix. Rather, it splits the data into chunks of size $p$ (10,000 by default) and then builds the matrix iteratively as needed. This allows it to exploit multiple cores---although we do not use this in our paper. This can lower the memory footprint of the algorithm considerably. One point of caution, however, is that the basis for the penalized terms (i.e., the hierarchical terms) are formed only on a random sample of chunk size $p$ taken at the start of the algorithm. For our purposes, this means that \texttt{bam} freezes the size of the sketching matrix at $\delta (p)^{1/3}$. If $p = 10,000$ (default), this is around $21 \delta$ if $N > 10,000$. This chunk size can be modified using arguments to \texttt{bam} but may slow down the algorithm somewhat. Thus, while \texttt{bam} can be helpful for large-scale problems, one should be careful to check any potential impacts of chunk size $p$ (perhaps by increasing $\delta$).

We examine \texttt{bam} systematically throughout the appendices. The main places where the chunk size issue could cause different results would be for (i) the simulations in Section~\ref{sec:validate_gKRLS} where $N > 10,000$ and (ii) the analysis of \cite{gulzar2020does} where $N = 30,000$ for the full-sample analysis (although note that $N < 10,000$ for the algorithms relying on sample-splitting). We find little evidence of difference between \texttt{bam} and \texttt{gam}.

\subsection{Calculating Average Marginal Effects}
\label{app:mfx_diff}

Our accompanying software provides the ability to calculate average expected outcomes (e.g., average predicted probabilities) as well as ``marginal effects'' (e.g., first differences). For continuous predictors, we also include the ability to calculate the average marginal effect, i.e., the average of the partial derivative of the prediction with respect to a single covariate \citep{hainmueller2014kernel}; Appendix~\ref{app:newman} provides a specific example. 

Following existing software, we do this using numerical differentiation; \cite{leeper2016mfx} provides a detailed discussion. This is especially important for complex models where a single covariate could appear multiple times and using an analytical approach is difficult to implement in a flexible fashion. We use following formula following \cite{leeper2016mfx}, shown for a two argument function for simplicity:

\begin{equation*}
    \frac{\partial f(x,y)}{\partial x} = \lim_{h \to 0} \frac{f(x + h,y) - f(x - h, y)}{2 h}
\end{equation*}

In practice, some small $h$ is used to approximate the derivative. The default setting for $h$ is $h = max(|x|,1)\sqrt{\epsilon}$ where $max(|x|,1)$ ranges over data distribution that one is marginalizing over (following \citealt{leeper2016mfx}) and $\epsilon$ is machine precision.  $h$ can be modified by the user if desired. Our package includes a function \texttt{legacy\_marginal\_effect} that can calculate the analytical average marginal effect in the simple case of a single kernel and some limited choices for outcome (e.g., Gaussian). Standard errors and confidence intervals on these quantities (or their averages) are calculated using the delta method.

\section{Additional Simulations: Scalability of \texttt{gKRLS}}
\label{app:scale_sims}

This appendix provides additional simulations to complement Section~\ref{sec:validate_gKRLS}. Appendix~\ref{app:define_perf} defines how we measure model performance and shows the error on estimating the average marginal effect. Appendix~\ref{app:alt_sketch} shows the performance of alternative sketching methods as the dataset grows to one million observations. Appendix~\ref{app:disagg_time} breaks down estimation time by different parts of \texttt{gKRLS}.

\subsection{Assessment of Model Performance}
\label{app:define_perf}

We consider two ways of assessing the performance of \texttt{gKRLS} following \cite{hainmueller2014kernel}. First, the main text considers the out-of-sample predictive accuracy. We do this by generating a dataset of identical size to the estimation data using the same data generating process. We then calculate the prediction for each observation in the test data and summarize the performance using the root mean squared error (RMSE).

Alternatively, we compare the estimated average marginal effect to the true value. The true value is computed by calculating the marginal effect for each observation in the training data (i.e. the analytical partial derivative of the data generating process with respect to the covariate) and then taking their average. Computing this quantity with respect to an out-of-sample dataset, i.e. the average effect on an identically sized population not used to estimate the original model, returns nearly identical results.

\begin{figure}[!htbp]
\caption{Error on Estimating Average Marginal Effect}
\label{fig:app_rmse_ame_sim1}
\includegraphics[width=\textwidth]{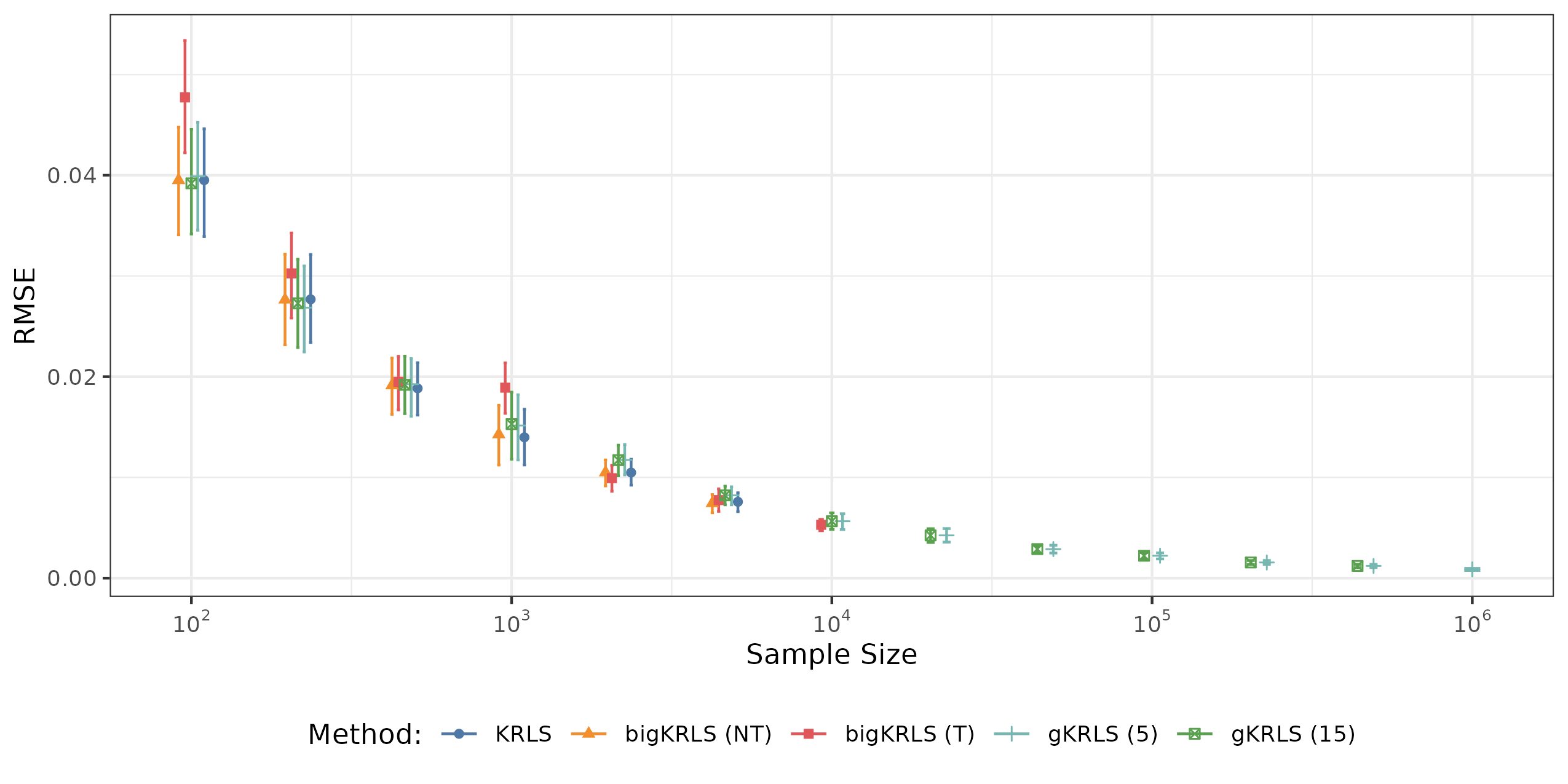}
\caption*{\footnotesize \emph{Note}: The figure reports the RMSE of the estimated average marginal effect (AME) averaged across both covariates and fifty simulations. 95\% confidence intervals are reported using a percentile bootstrap using 1,000 bootstrap samples.}
\end{figure}

Figure~\ref{fig:app_rmse_ame_sim1} shows the results for the five models in the main text on the root mean squared error (RMSE; averaged across fifty simulations and both variables) of the estimated average marginal effect. It shows a worse performance of bigKRLS with truncation (``bigKRLS (T)'') for some sample sizes---although less dramatically than in the main text.

\subsection{Alternative Sketch Methods}
\label{app:alt_sketch}

\cite{yang2017randomized} raise a number of issues with random sketching---especially in the case of complex data where sub-sampling is likely to yield a bad representation of the original data; \cite{lee2020econometric} provides a recent overview of different techniques that may address this problem. To explore this for our initial simulations, we also consider \cite{yang2017randomized}'s Gaussian sketching; Appendix~\ref{app:sketch_sims} considers it for the second set of simulations. 

Formally, for some sketching dimension $M$, Gaussian sketching generates a matrix that is $M \times N$ where each element is drawn from a $N(0,1/\sqrt{M})$ following \cite{yang2017randomized}. In this case, the sketched kernel is some randomized combination of \emph{all} observations---rather than simply being constructed on a subset of observations. Despite the beneficial nature of this method, it incurs a considerably higher computational burden in building the sketched kernel as the kernel matrix $\bm{K}$ ($N \times N$) must be multiplied by a dense $N \times M$ matrix. If $N$ is very large, this can be quite expensive to compute even once. One possible solution is the parallelization of $\bm{K}\bm{S}^T$.

\begin{figure}[!htbp]
\begin{center}
\caption{Estimation Time of Different Methods}
\label{fig:app_timing}
\includegraphics[width=0.9\textwidth]{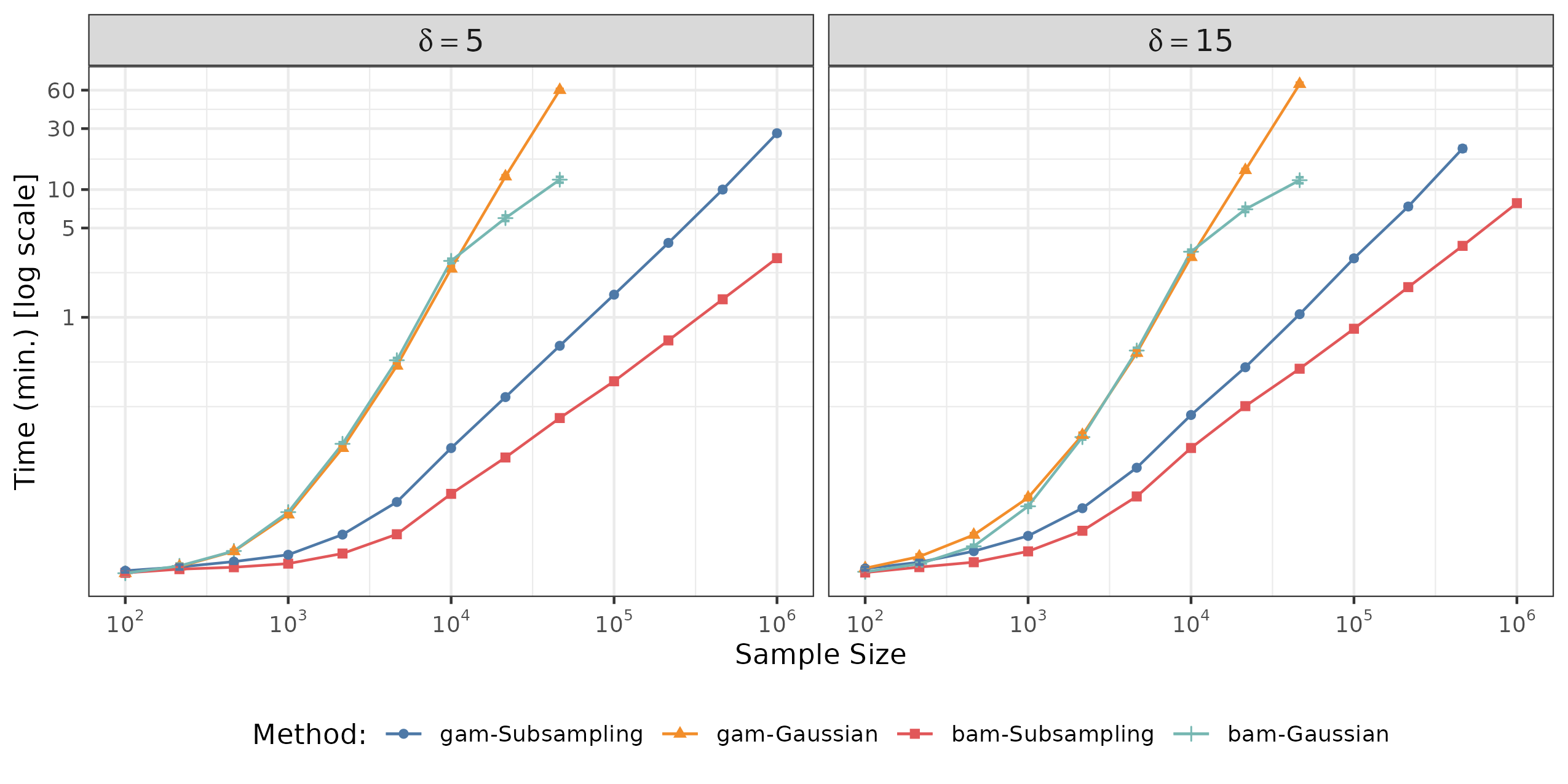}    
\caption*{\footnotesize \emph{Note:} This figure shows the average computational time in minutes averaged across fifty iterations. The left panel shows a multiplier of $\delta=5$ and the right shows a multiplier of $\delta=15$. The main text defines the abbreviations. 95\% confidence intervals are shown.}
\end{center}
\end{figure}

Figure~\ref{fig:app_timing} compares computational cost of Gaussian sketching. To examine alternative estimation methods designed for huge datasets (e.g., \texttt{gam} vs. \texttt{bam}; see Appendix~\ref{app:gam_bam}), we also compare results for \texttt{gam} and \texttt{bam} and types of sketching (sub-sampling sketching or Gaussian), denoted by ``method-sketching type''. We see that Gaussian sketching is considerably slower and rather expensive after around 45,000 observations. 

As noted in the main text, when using \texttt{gam} (sub-sampling sketching; multiplier $\delta=5$), estimation takes around 30-40 minutes for 1,000,000 observations. This figure illustrates that \texttt{bam} can help considerably. Estimation time is around three minutes, although as discussed in Appendix~\ref{app:gam_bam} this is partially due to a freezing of the size of the sketching dimension. 

We can also use the log-log plot to provide a rough estimate of the computational complexity of the various algorithms and sketching procedures; since the plot looks highly linear (above a certain sample size), we can use the slope $p$ of that log-log line to estimate the complexity as roughly $N^p$ for sufficiently large $N$. As expected, KRLS and bigKRLS is around $N^3$, with an estimated slope of around 3 when the sample size is over 100 for KRLS and 1,000 for bigKRLS. For the sub-sampling sketched methods, if we focus on a sample size above 40,000---as the relationship appears clearly linear---the estimated slope for the methods is around 1.3, suggesting a cost that increases much more slowly than traditional methods. When using \texttt{bam} (discussed in Appendix~\ref{app:gam_bam}), the slope is around 0.95, presumably because the size of the sketching dimension does not grow after $N > 10,000$. With Gaussian sketching, the estimated slope (when $N > 2500$) for \texttt{gam} is around 2.10; lower than the unsketched methods but considerably larger than the sub-sampling methods.

\begin{figure}[!htbp]
\begin{center}
\caption{Performance of Alternative Sketching and Estimation Methods}
\label{fig:app_timing_perf}
\includegraphics[width=\textwidth]{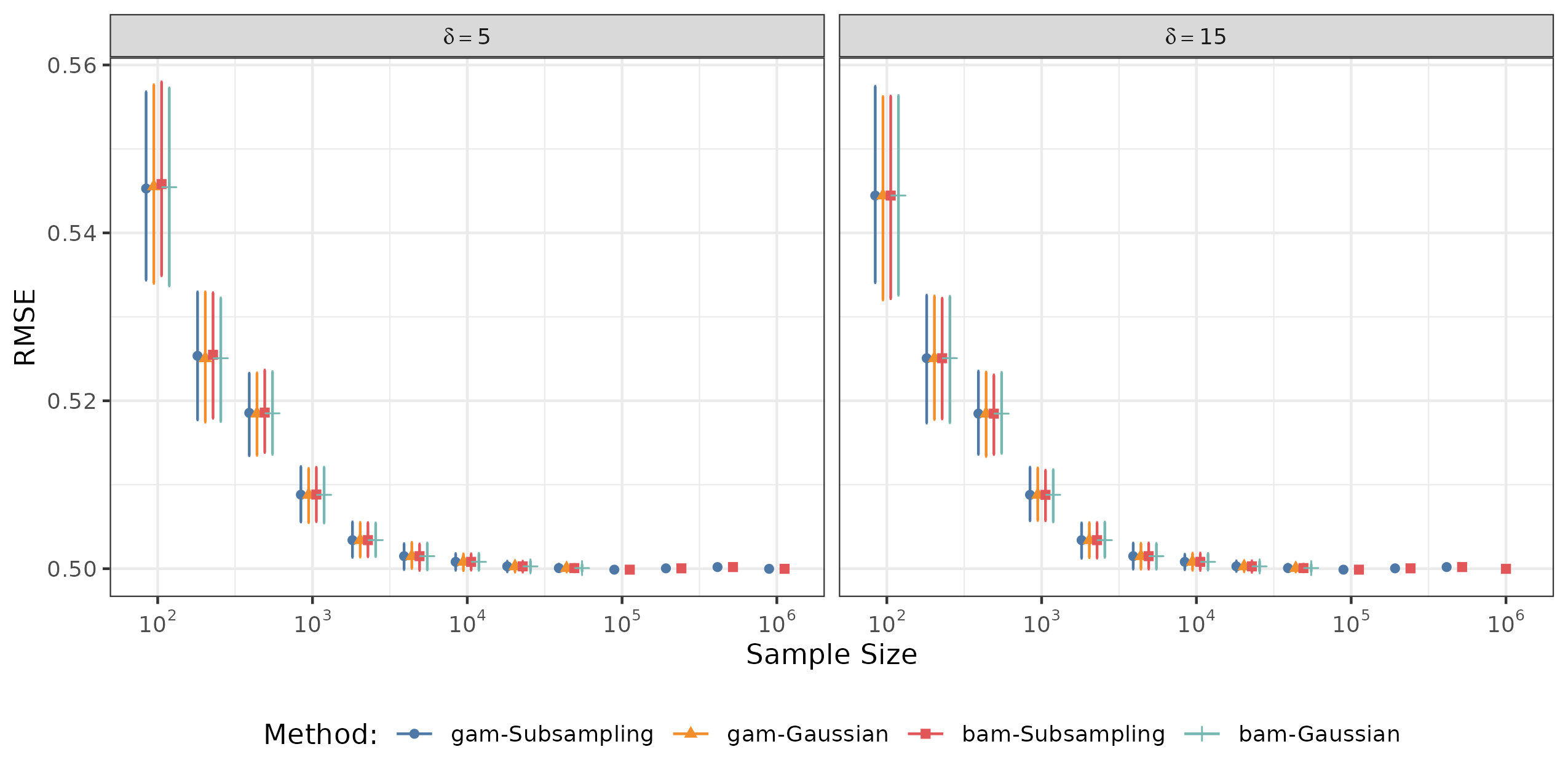}    
\caption*{\footnotesize \emph{Note:} This figure shows the root mean squared error (RMSE) of predicting the outcome on a dataset of the same size to the estimation data, averaged across the fifty simulations. The main text defines the abbreviation for each method. 95\% confidence intervals are shown; they are calculated using a percentile bootstrap using 1,000 bootstrap samples.}
\end{center}
\end{figure}

Figure~\ref{fig:app_timing_perf} shows that all sketching methods are equally accurate.

\subsection{Estimation Time Disaggregation}
\label{app:disagg_time}

When estimating \texttt{gKRLS} and using it for inference, there are three steps that the applied user may perform; it is useful to know which takes more time as sample size increases. First, the model must create the (sketched) kernel and estimate the parameters (``Estimation''); second, one might wish to perform prediction on a new dataset; we use one of the same size as the estimation data (``Prediction''). Finally, one might wish to calculate marginal effects (``Marg. Effects''); this requires repeated predictions on counterfactual versions of the estimation data. The total time is reported in the main text.

\begin{figure}[!htbp]
\caption{Estimation Time of \texttt{gKRLS} by Step}
\label{fig:app_disagg_time}
\includegraphics[width=\textwidth]{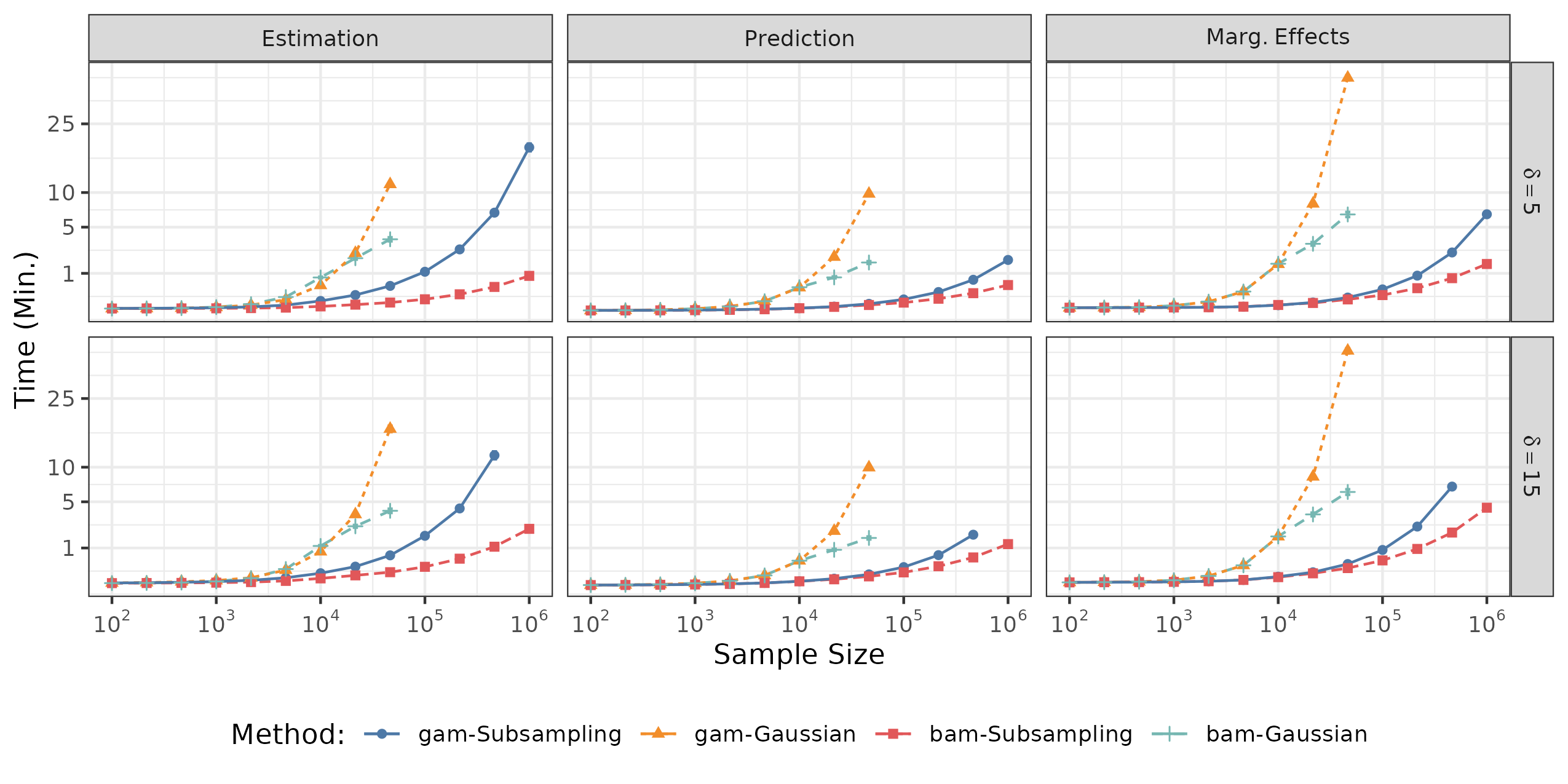}
\caption*{\footnotesize \emph{Note:} This figure shows the average computational time in minutes averaged across fifty iterations by each stage of estimation (``Estimation'', ``Prediction'', ``Marg. Effects''). The main text defines the abbreviation for each method. 95\% confidence intervals are shown.}
\end{figure}

Figure~\ref{fig:app_disagg_time} reports the average time for each stage across sample size and for the four sketching methods considered. It uses a log-scale for readability. We see that, for sub-sampling sketching, estimation time becomes highly expensive as $N$ (and the parameter dimension) grows very large and dominates the overall cost. For Gaussian sketching, the cost of calculating marginal effects becomes increasingly expensive as the cost of actually evaluating the kernel begins to dominate the computational cost. This also likely explains the explosion of computational time for estimation at an earlier stage than the sub-sampling method. As expected, \texttt{bam} provides considerable gains in speed---mostly in the estimation stage.

\section{Additional Simulations: \texttt{gKRLS} with Fixed Effects}
\label{app:extra_sims}

This appendix contains additional information on the simulations in Section~\ref{sec:sim2main} where we include fixed effects in the data generating process. Appendix~\ref{app:alt_metrics} reports different performance metrics such as bias and out-of-sample predictive RMSE (see Appendix~\ref{app:define_perf} for definitions). Appendix~\ref{app:alt_control} reports the performance on the other covariate ($x_{i,2}$). Appendix~\ref{app:vary_sample_size} varies the number of observations. Appendix~\ref{app:alt_flavors} considers different variants of \texttt{gKRLS} to understand what drives the improved performance versus bigKRLS. Appendix~\ref{app:sketch_sims} provides information on the impact of sketching in this more complicated scenario.

\subsection{Alternative Performance Metrics}
\label{app:alt_metrics}

\begin{figure}[!htbp]
    \caption{Bias of AME}
    \label{fig:app_AME_bias}
    \includegraphics[width=\textwidth]{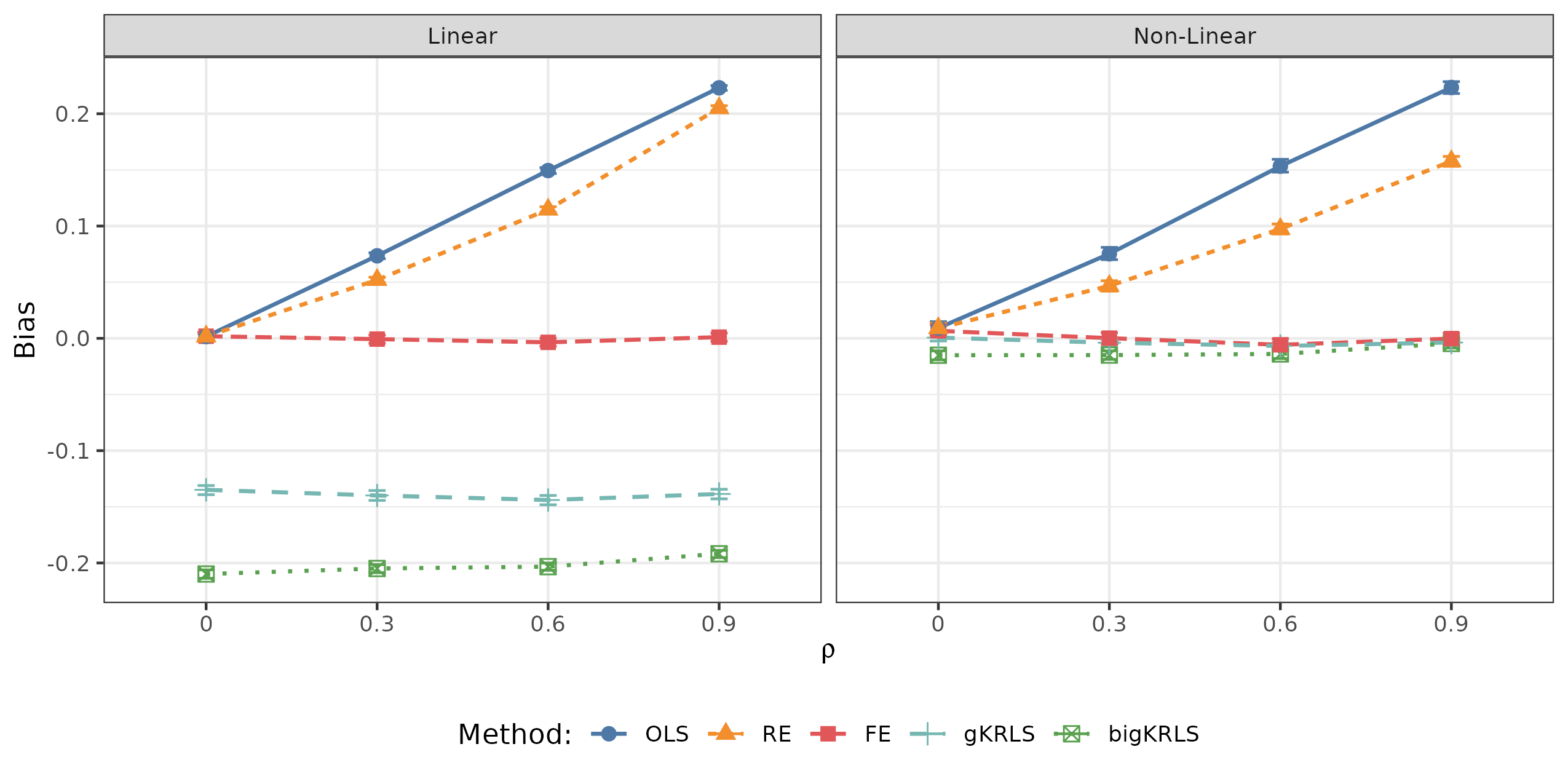}
    \caption*{\footnotesize \emph{Note:} The figure reports the bias of the estimated average marginal effect (AME) on the first covariate ($x_{i,1}$), averaged across all simulations, as $\rho$ varies. The left panel shows the linear data generating process and the right shows the non-linear data generating process. 95\% confidence intervals are shown; they are calculated using a percentile bootstrap using 1,000 bootstrap samples.}
\end{figure}

In addition to the RMSE, Figure~\ref{fig:app_AME_bias} shows the estimated bias of the methods in Figure~\ref{fig:sim2} for the AME of the main covariate of interest ($x_{i,1}$). It shows, as expected, that the fixed effects estimator is unbiased and that the bias of the OLS and RE methods increase considerably as $\rho$ increases. In the linear model, the kernel methods are biased downwards although the bias for \texttt{gKRLS} is considerably smaller. In the non-linear case, both have a small bias, although bigKRLS's is slightly larger at small $\rho$.

We next compare the out-of-sample predictive accuracy of the methods---estimated using the procedure in Appendix~\ref{app:define_perf}. Figure~\ref{fig:app_pred_oos} reports the RMSE averaged across all 1,000 simulations. For the linear data generating process, the kernel methods perform worse than fixed effects or random effects, although the differences are more modest. In the non-linear case, we see the kernel methods perform the best although \texttt{gKRLS} out-performs bigKRLS by a considerable margin.

\begin{figure}[!htbp]
\begin{center}
\caption{Out-of-Sample Predictive Accuracy}
\label{fig:app_pred_oos}
\includegraphics[width=\textwidth]{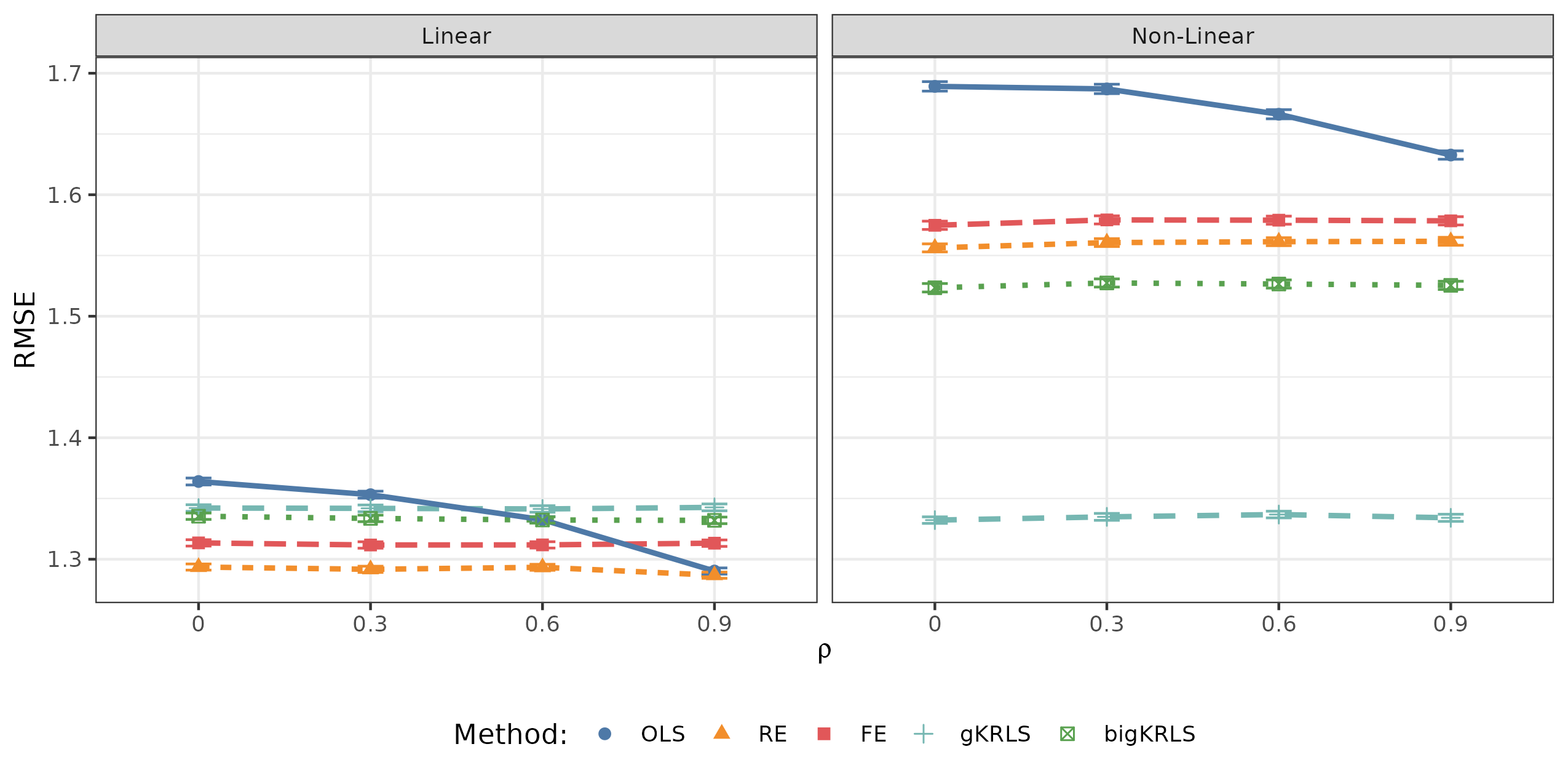}
\caption*{\footnotesize \emph{Note:} This figure shows the root mean squared error (RMSE) of predicting the outcome on a dataset of the same size to the estimation data, averaged across 1000 simulations, as $\rho$ varies. The left panel shows the linear data generating process and the right shows the non-linear data generating process. 95\% confidence intervals are shown.}
\end{center}
\end{figure}

\subsection{Performance on Other Covariate ($x_{i,2}$)}
\label{app:alt_control}

\begin{figure}[!htbp]
\begin{center}
\caption{RMSE on the Average Marginal Effect of $x_{i,2}$}
\label{fig:app_rmse_control}
\includegraphics[width=\textwidth]{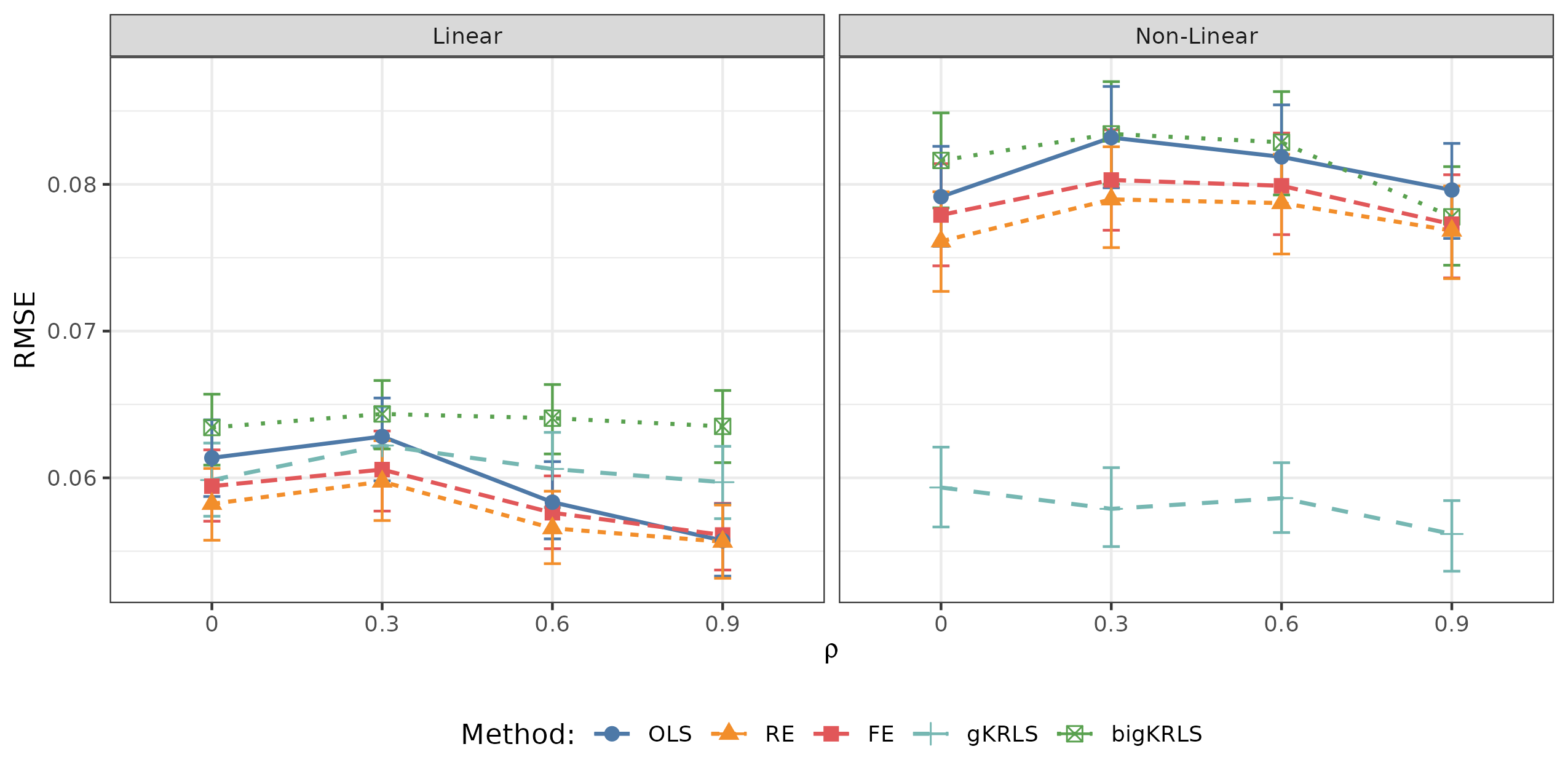}
\caption*{\footnotesize \emph{Note:} The figure reports the bias of the estimated average marginal effect (AME) on the second covariate ($x_{i,2}$), averaged across 1000 simulations, as $\rho$ varies. The left panel shows the linear data generating process and the right shows the non-linear data generating process. 95\% confidence intervals are shown; they are calculated using a percentile bootstrap using 1,000 bootstrap samples.}
\end{center}
\end{figure}

We replicate Figure~\ref{fig:sim2} from the main text on the covariate $x_{i,2}$, i.e. one that is not correlated with the fixed effect. Figure~\ref{fig:app_rmse_control} shows that, in the linear case, all methods perform rather similarly at estimating this average marginal effect---including methods such as random effects or OLS. In the spline case, \texttt{gKRLS} clearly out-performs all other methods.

\subsection{Varying Number of Observations Per Group}
\label{app:vary_sample_size}

\begin{figure}[!htbp]
\begin{center}
\caption{Estimating AME for Varying Sample Size (Linear)}
\label{fig:app_vary_AME_linear}
\includegraphics[width=0.95\textwidth]{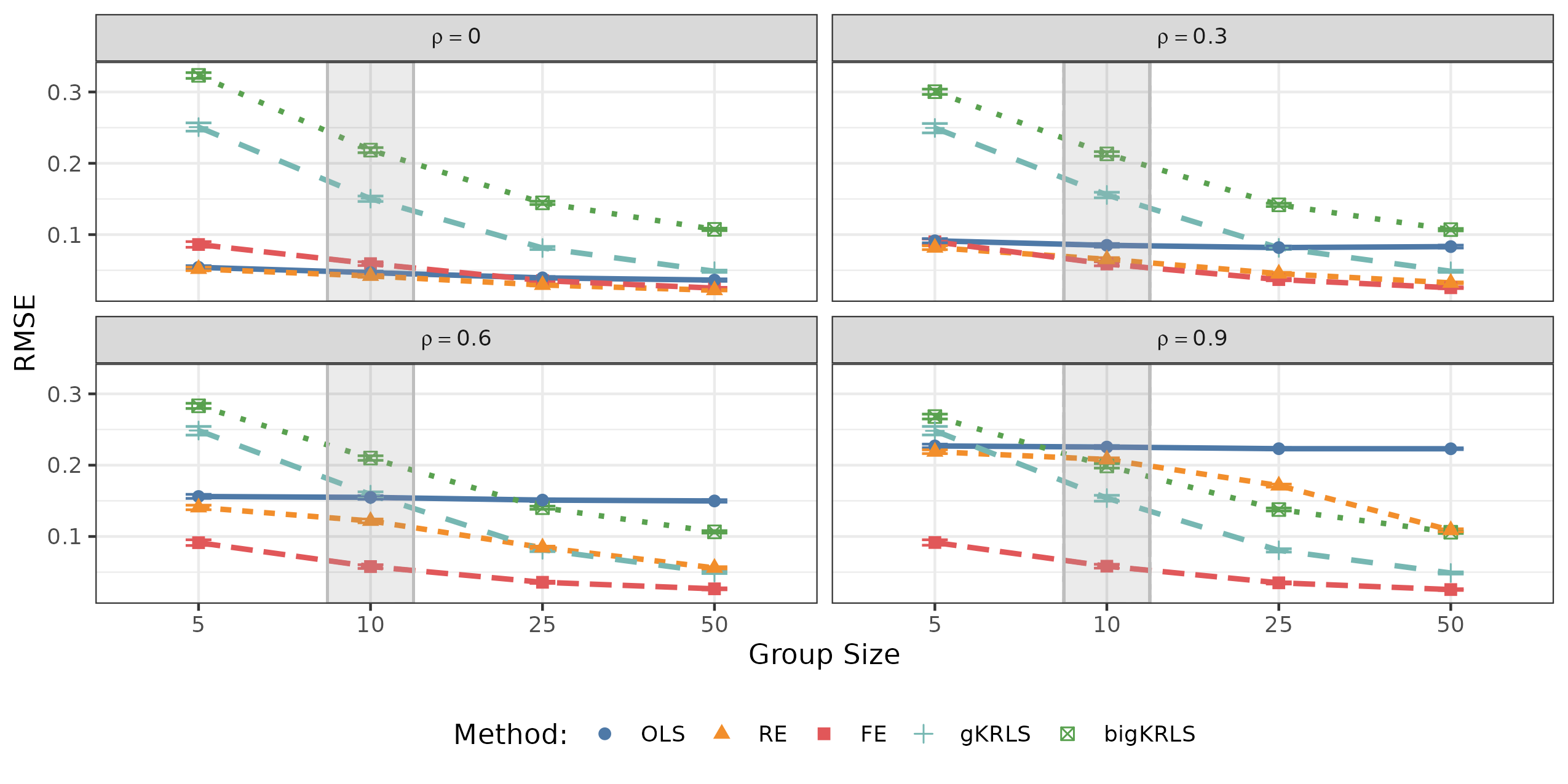} \caption*{\footnotesize \emph{Note:} This figure reports the RMSE of the estimated AME on the first covariate ($x_{i,1}$) at varying sample sizes for the linear data generating process. The shaded box indicates the values reported in the main analyses, i.e. $T = 10$. 95\% confidence intervals are shown; they are calculated using a percentile bootstrap using 1,000 bootstrap samples.}  
\end{center}
\end{figure}

\begin{figure}[!htbp]
\begin{center}
\caption{Estimating AME for Varying Sample Size (Non-Linear)}
\label{fig:app_vary_AME_spline}
\includegraphics[width=0.95\textwidth]{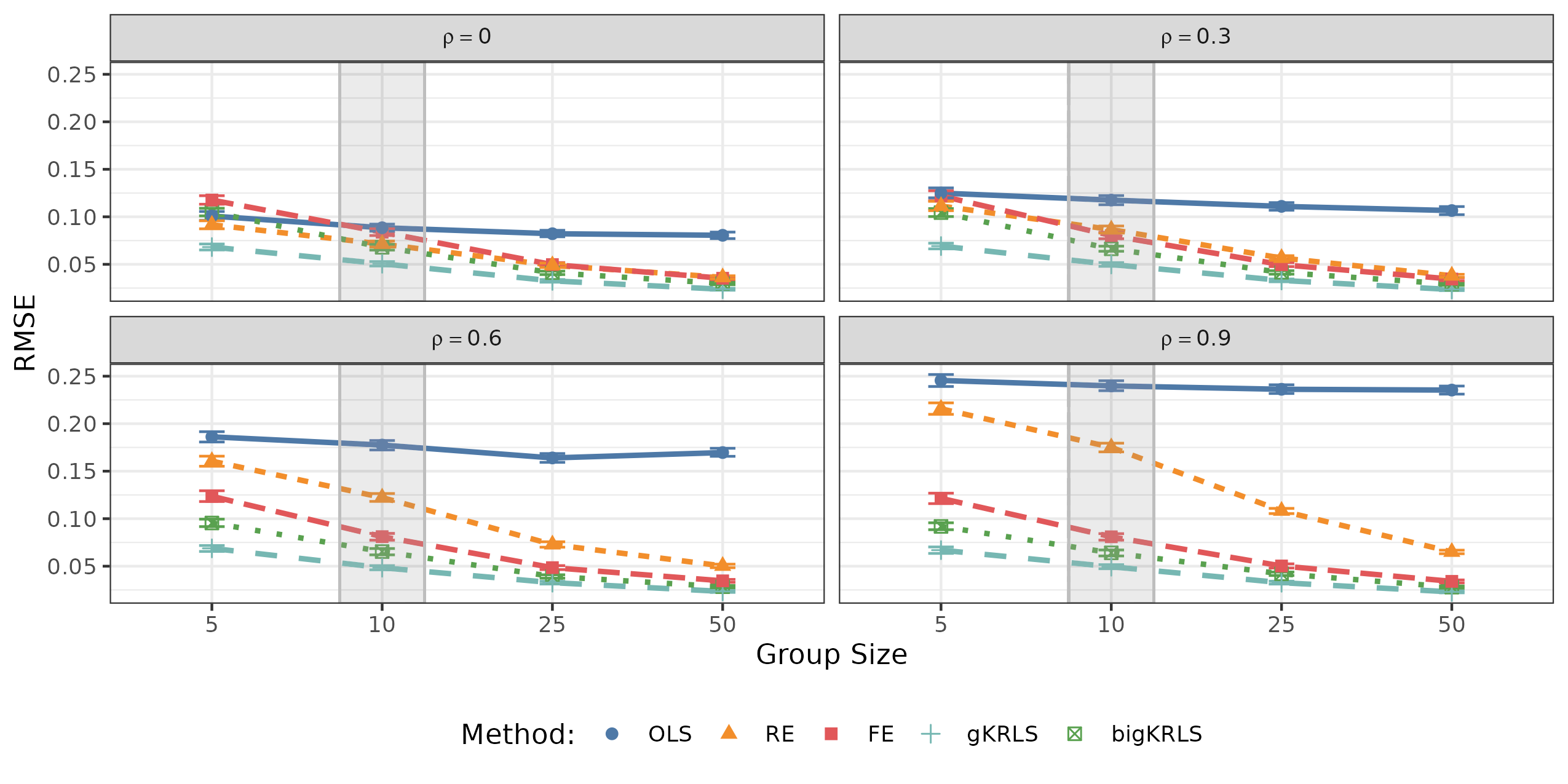} 
\caption*{\footnotesize \emph{Note:} This figure reports the RMSE of the estimated AME on the first covariate ($x_{i,1}$) at varying sample sizes for the non-linear data generating process. The shaded box indicates the values reported in the main analyses, i.e. $T = 10$.  95\% confidence intervals are shown; they are calculated using a percentile bootstrap using 1,000 bootstrap samples.}
\end{center}
\end{figure}

In the main text, we consider 50 groups ($J = 50$) and set the group size (i.e. number of observations per group) at 10. We vary that here and consider $T \in \{5, 10, 25, 50\}$. Figures~\ref{fig:app_vary_AME_linear} and~\ref{fig:app_vary_AME_spline} show the results as group size varies where the gray box indicates the results in the main text. Each panel displays a different value of $\rho$.

Looking first at the linear case (Figure~\ref{fig:app_vary_AME_linear}), we see that as the cluster size increases, random effects and both kernel methods improve considerably. For larger cluster sample sizes, \texttt{gKRLS} is close to the RMSE for the fixed effect method although bigKRLS remains noticeably worse. This provides further evidence for the limitations of adding the fixed effects directly into the kernel. In the non-linear case, the story is similar---random effects improves with increasing sample size towards fixed effects---although \texttt{gKRLS} and bigKRLS are much closer in performance (as in the main text).

\subsection{Understanding Why \texttt{gKRLS} Does Better}
\label{app:alt_flavors}

This section explores in more depth \emph{why} \texttt{gKRLS} provides an improvement on bigKRLS. We first consider the most similar model to bigKRLS fit using \texttt{mgcv} (i.e. no sketching, standardized covariates, and fixed effects inside the kernel). We then vary each of these dimensions separately to the default settings in \texttt{gKRLS} and see which affects performance. This allows us to disaggregate what seems to improve performance in a more controlled setting. We consider the following models.

\begin{enumerate}
    \item ``Baseline'': \texttt{gKRLS} estimated with no sketching, fixed effects inside the kernel, standardized covariates (mean zero; variance one), GCV for penalty parameter selection.
    \item ``Mahal.'': ``Baseline'' but use Mahalanobis distance between covariates in the kernel.
    \item ``FE'': ``Baseline'' but include the fixed effects \emph{outside} the kernel.
    \item ``Sketch(5)'': ``Baseline'' but use sketching with the default multiplier of five.
    \item ``Sketch(5)+REML'': Identical to ``Sketch(5)`` but use REML instead of GCV to select $\lambda$.
    \item ``Naive'':  This method includes the fixed effects in the kernel, but uses sketching, Mahalanobis distance, and GCV for penalty selection. It is ``naive'' as it simply swaps bigKRLS for \texttt{gKRLS} under the default settings of the package and \texttt{mgcv}.
    \item ``gKRLS (GCV)'':  \texttt{gKRLS} estimated with sub-sampling sketching (multiplier of 5), fixed effects \textbf{outside} the kernel, Mahalanobis standardization, GCV for penalty parameter selection.
    \item ``gKRLS + Lin.'': This method includes the fixed effects and both covariates both inside and outside of the kernel. It ensures that as the amount of regularization increases, this shrinks towards a linear additive model.
    \item ``gKRLS'': The method shown in the main text. That is, ``gKRLS (GCV)'' but using REML for penalty parameter selection.
\end{enumerate}

\begin{figure}[!htbp]
\caption{Additional Versions of \texttt{gKRLS}}
\label{fig:app_AME_extra}
\includegraphics[width=\textwidth]{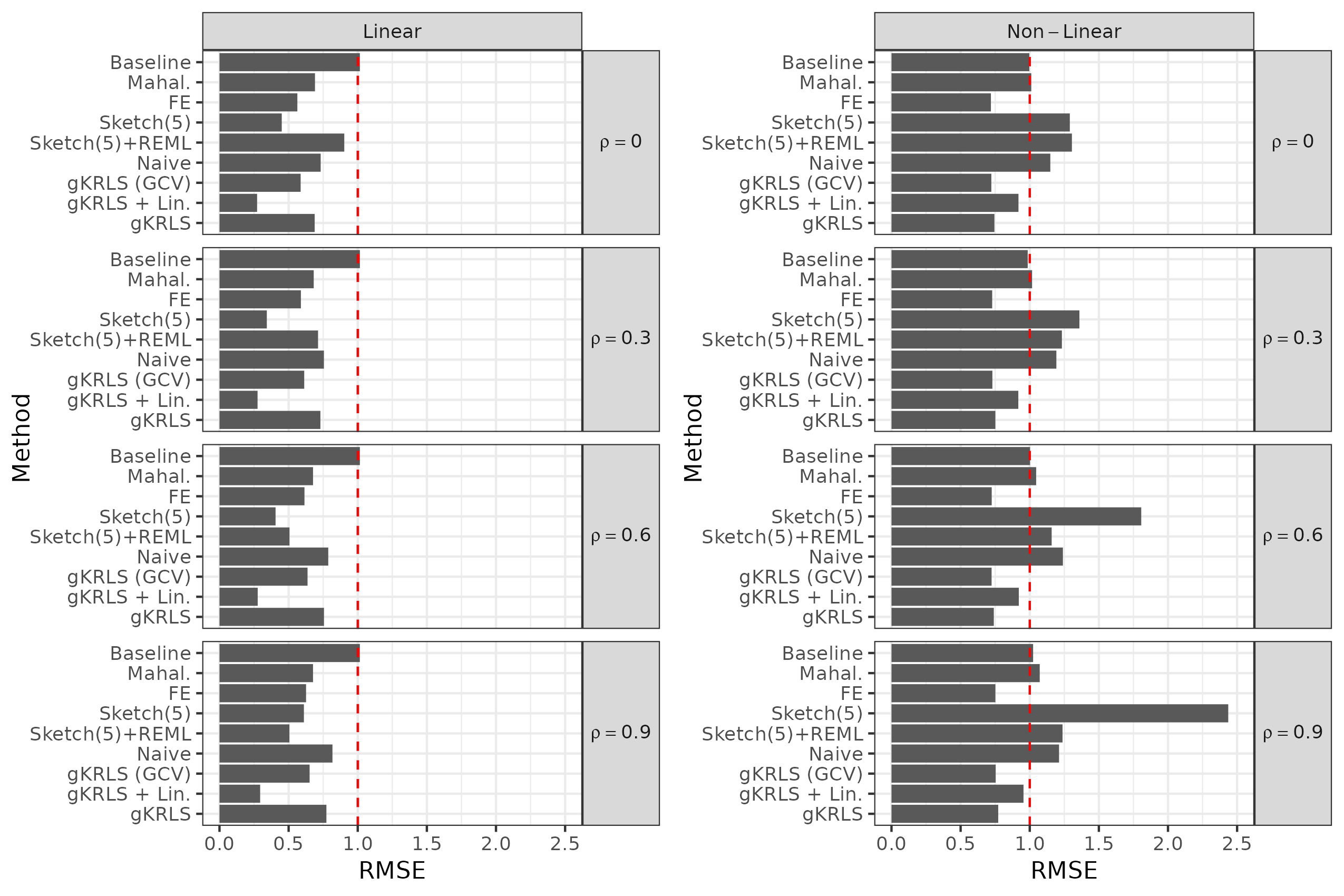}
\caption*{\footnotesize \emph{Note:} This figure reports the RMSE of various methods for estimating the AME on $x_{i,1}$ relative to the RMSE of bigKRLS. The abbreviations are defined in the text.}
\end{figure}

Figure~\ref{fig:app_AME_extra} reports the results where the estimated RMSE on the AME is divided by the RMSE from bigKRLS without truncation. Values above ``1'' indicate worse performance; values below 1 indicate better performance.

The first observation is that the ``Baseline'' \texttt{gKRLS} model that is as similar as possible to bigKRLS returns nearly equivalent performance. This corroborates the results in the first set of simulations (Section~\ref{sec:validate_gKRLS}).

Next, consider the models that change only one feature of the ``Baseline'' model: ``Mahal.'', ``FE'', ``Sketch(5)''. With a linear data generating process, all three result in improved performance. In the non-linear data generating process, ``Sketch(5)'' results in worse performance. After further investigating, this is seemingly a function of using GCV (Generalized Cross-Validation) to select $\lambda$; the bar below (``Sketch(5)+REML'') uses REML instead of GCV and finds a slight degradation in performance (see Appendix~\ref{app:sketch_sims}) but nothing as catastrophic as with GCV. This may be due to some weaknesses of GCV, including a tendency to overfit for modestly sized datasets (see \citealt{wood2011fast}). 

Of all of the simple modifications, note that ``FE'' is the only one that results in considerably improved performance. Thus, this provides evidence that putting the fixed effects outside of the kernel is what drives stronger performance of \texttt{gKRLS}.

Next, we examine the ``Naive'' specification; this simply swaps bigKRLS for \texttt{gKRLS}---using random sketching and Mahalanobis distance while keeping the fixed effects \emph{inside} the kernel. In the non-linear setting, ``Naive'' incurs some penalty against a method with no sketching (``Baseline''). By contrast, ``gKRLS (GCV)'' that is the sketched, Mahalanobis, version of ``FE'' incurs a very slight penalty upon the unsketched version (``FE'').  Methods that include fixed effects outside the kernel (``gKRLS (GCV)'', ``gKRLS'', ``gKRLS + Lin.'') improve considerably upon the ``Baseline'' and ``Naive'' model---especially with a non-linear data generating process. In terms of penalty parameter selection after using sketching and Mahalanobis distance (REML vs GCV; ``gKRLS'' and ``gKRLS (GCV)'', respectively), we see a small negative impact for REML in the linear model, although there is little difference in the non-linear case.

Finally, we consider the ``gKRLS + Lin.'' specification that includes the group indicators \emph{and} two covariates as unpenalized fixed effects $\bm{\beta}$ as well as a kernel on the two continuous covariates. In this model, there is one hierarchical term ($J=1$) and as $\lambda \to \infty$, this model converges to a linear additive model of the group indicators and two covariates. By contrast, simply placing a kernel on the two variables (with no linear component) would converge a model that predicts the mean for all observations as $\lambda \to \infty$. We see that in the linear case, where this model is true, this specification has very strong performance. However, in the non-linear model, it incurs some penalty versus the ``default'' \texttt{gKRLS}. Thus, this suggests that deciding whether include linear terms in addition to kernel is either tested empirically or motivated based on what theory suggests is a reasonable limiting case for the model as $\lambda \to \infty$.

\subsection{Impact of Sketching}
\label{app:sketch_sims}

Give the somewhat varying results of sketching procedures depending on whether the fixed effects are included in the kernel or not, we conducted one additional set of simulations to more systematically understand the impact of sub-sampling sketching in this more complicated example. We focus on sub-sampling sketching and conducted the following set of simulations using the same data generating process in Section~\ref{sec:sim2main}.

\begin{itemize}
    \item Generate a set of simulated data; for $\delta \in \{1, 3, 5, 7, 9, 11, 13, 15\}$, estimate \texttt{gKRLS} 100 times (i.e., varying the sketching matrix).
    \item We consider two quantities of interest. First, what is the RMSE of the average marginal effect on $x_{i,1}$ (the key quantity of interest in the main text) across the 100 repetitions for each $\delta$? We define this as follows, where $\mathrm{AME}^*$ is the true AME as calculated in our main simulations (see Appendix~\ref{app:define_perf}) and $\mathrm{AME}^{(m)}_\delta$ is the AME estimated for the $m$-th repetition with sketching multiplier $\delta$.

    $$\mathrm{RMSE}_\delta = \sqrt{\frac{1}{100}\sum_{m=1}^{100} \left[\mathrm{AME}^{(m)}_\delta - \mathrm{AME}^*\right]^2}$$

    Next, we also calculate the AME for the \emph{unsketched} method, define this as $\mathrm{AME}_{\mathrm{unsketch}}$; we can calculate the corresponding RMSE without sketching as $\mathrm{RMSE}_{\mathrm{unsketch}}$---corresponding to the absolute error the unsketched model is deterministic for a single dataset:

    $$\mathrm{RMSE}_{\mathrm{unsketch}} = | \mathrm{AME}_{\mathrm{unsketch}} - \mathrm{AME}^*| = \sqrt{\left[\mathrm{AME}_{\mathrm{unsketch}}- \mathrm{AME}^*\right]^2}$$

    \item Given those quantities for a single simulated dataset, we repeat this process 150 times as the quality of the unsketched estimates can vary across datasets. Define $s \in \{1, \cdots, 150\}$ as indexing this ``outer'' simulation. We average the estimated RMSEs across simulations and create a measure of ``relative impact'' of sketching, i.e. the relative increase in (averaged) RMSE for using sketching versus the unsketched estimates.
    $$\overline{\mathrm{RMSE}}_\delta = \frac{1}{150}\sum_{s=1}^{150} \mathrm{RMSE}^{(s)}_\delta; \quad \overline{\mathrm{RMSE}}_{\mathrm{unsketch}} = \frac{1}{150}\sum_{s=1}^{150} \mathrm{RMSE}^{(s)}_{\mathrm{unsketch}}$$
    $$\mathrm{RelativeImpact}_\delta = \frac{\overline{\mathrm{RMSE}}_\delta - \overline{\mathrm{RMSE}}_{\mathrm{unsketch}}}{\overline{\mathrm{RMSE}}_{\mathrm{unsketch}}}$$

\end{itemize}

\begin{figure}[!htbp]
\caption{RMSE For Varying Sketching Multiplier}
\label{fig:app_sketch_multi_RMSE}
\includegraphics[width=\textwidth]{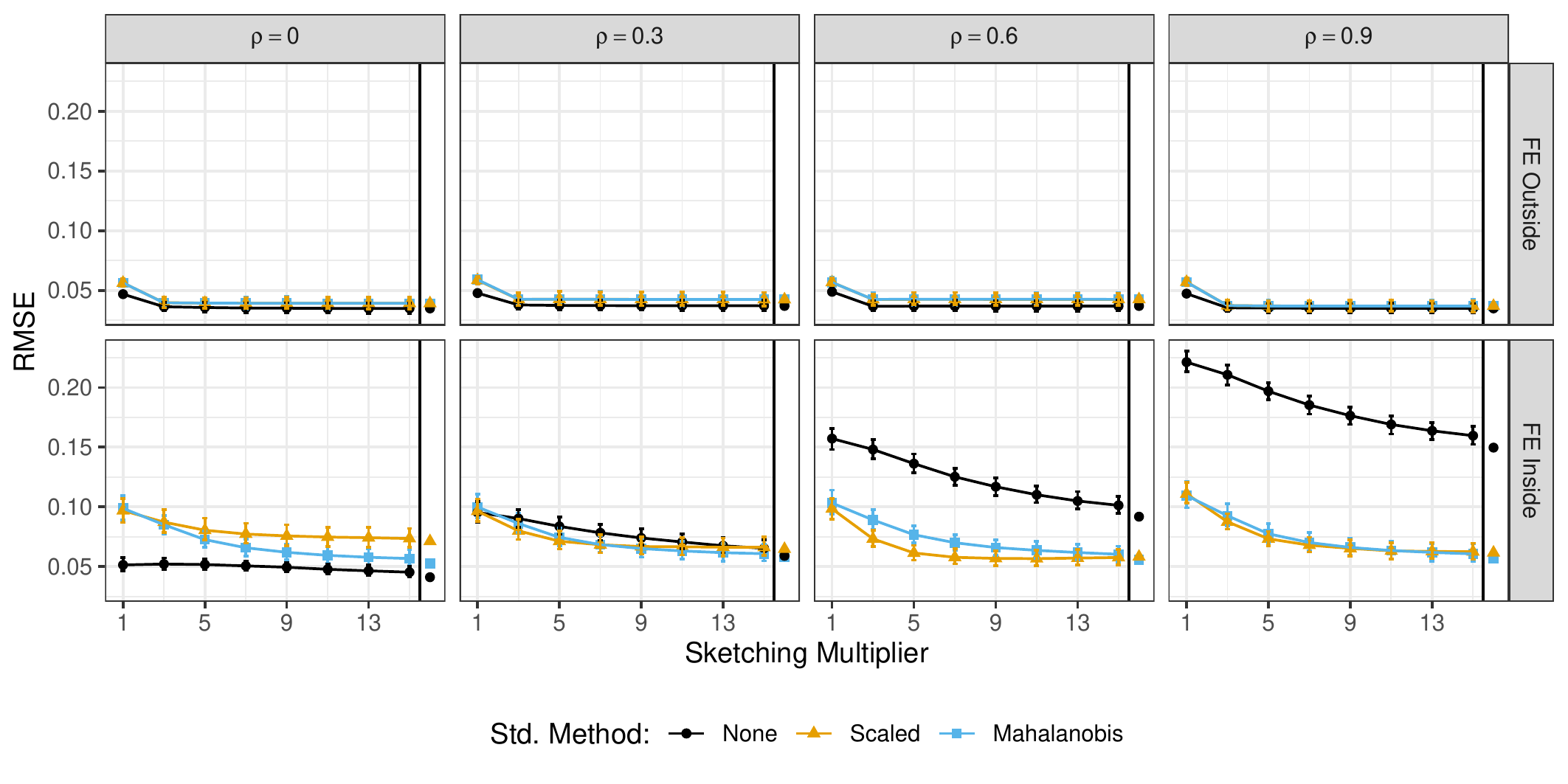}
\caption*{\footnotesize \emph{Note:} The RMSE for each sketching multiplier, $\rho$ and model specification is shown. 95\% confidence intervals, using a percentile bootstrap over the 150 datasets using 1,000 bootstrap samples, are shown. The RMSE of the unsketched method is shown to the right of the vertical bar.}
\end{figure}

We consider all four $\rho$ from the original simulations, sub-sampling sketching, and three different standardizations methods for the covariates (none [no standardization], scaled [i.e., all covariates have zero mean and unit variance], and Mahalanobis). Figure~\ref{fig:app_sketch_multi_RMSE} begins by showing the RMSE, averaged across simulations, for each combination. ``FE Outside'' denotes a model where the fixed effects are included outside the kernel and ``FE Inside'' denotes the one where they are included inside. We include the average RMSE of unsketched estimates as a dot on the right side of the figure after the black vertical bar.

As expected from the earlier simulations, including the fixed effects outside the kernel is highly amendable to sketching; at almost any multiplier, the RMSE is very close to the unsketched estimates, and this does not seem especially dependent on $\rho$ or the standardization method.

By contrast,  including the fixed effects 
inside the kernel shows more nuanced results. Before examining sketching, we note that including the fixed effects inside the kernel results in worse performance, on average, at every $\delta$, $\rho$ and standardization method than including them outside the kernel. 
However, given that choice of specification, there is more divergence between the standardization methods (especially with poor performance for no standardization as $\rho$ grows). For this complex kernel, the convergence in the estimated RMSE to the unsketched method is much slower (i.e., a large multiplier $\delta$ is needed to closely approximate the RMSE of the unsketched method).

\begin{figure}
\caption{Relative Impact of Sketching}
\label{fig:app_relimpact_sketching}
\includegraphics[width=\textwidth]{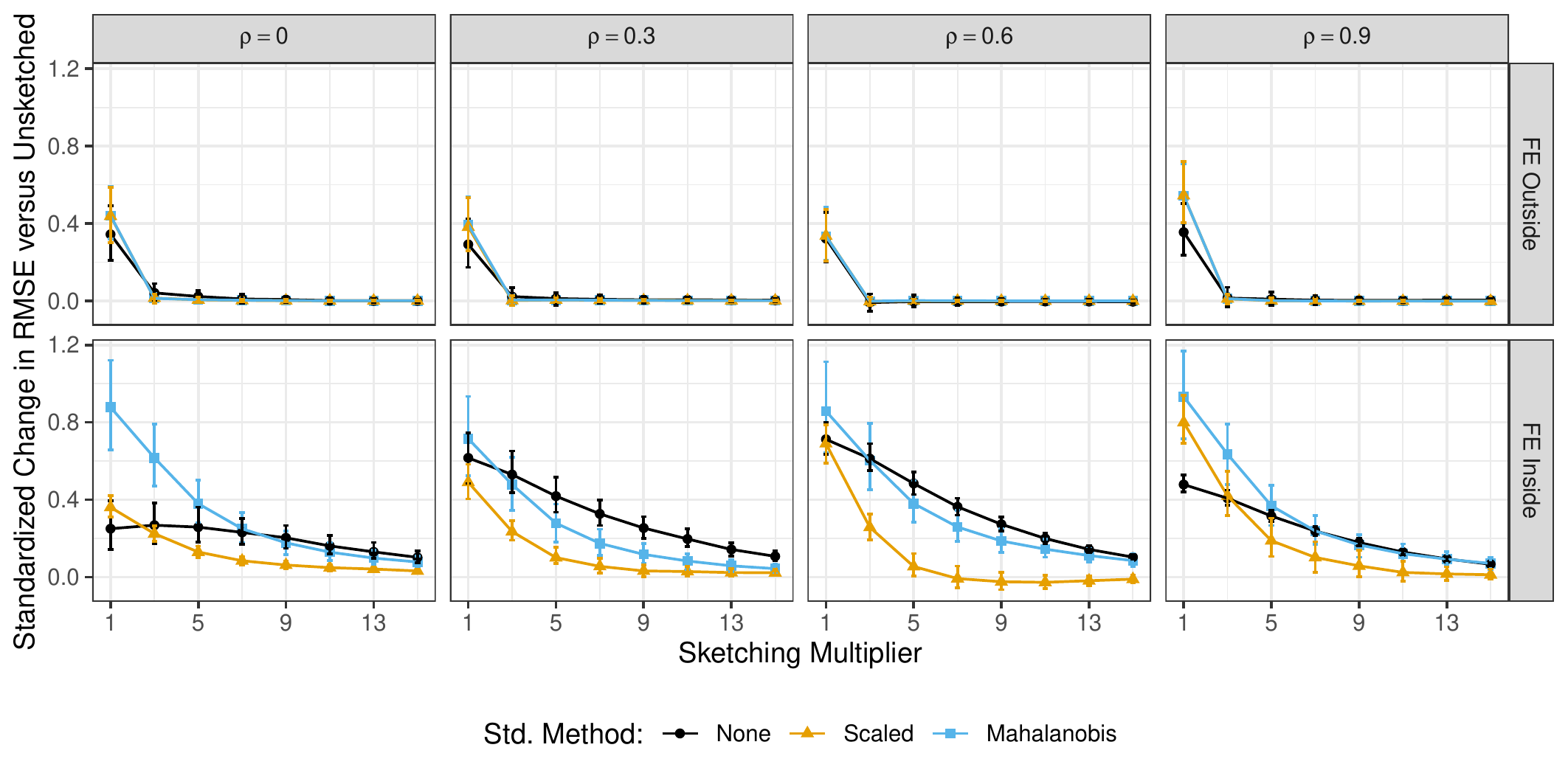}
\caption*{\footnotesize \emph{Note:} This figure shows the relative impact of sketching ($\mathrm{RelativeImpact}_\delta$) for each sketching multiplier, $\rho$ and model specification (described in the main text). 95\% confidence intervals are reported using a percentile bootstrap over the 150 datasets using 1,000 bootstrap samples.}
\end{figure}

To show this more clearly, we report the standardized change in RMSE for sketching: $\mathrm{RelativeImpact}_\delta$, defined above. This can be interpreted the proportional increase in RMSE that comes from using sketching with multiplier $\delta$ versus the unsketched method. Figure~\ref{fig:app_relimpact_sketching} corroborates the above results. After a sketching multiplier of around $\delta=3$, the RMSE for sketching when the fixed effects are outside the kernel are nearly identical to the unsketched RMSE.  When fixed effects are included inside the kernel, there is a much more gradual decline in the relative impact of using sketching on the RMSE versus the unsketched estimates. 

Interestingly, scaled covariates (mean-zero; variance one) seems to improve more quickly than Mahalanobis relative to the unsketched baseline, although note that the \emph{absolute} performance of scaled vs. Mahalanobis standardization depends on $\rho$ (see Figure~\ref{fig:app_sketch_multi_RMSE}.

Overall, while more work is needed to understand the impacts of sketching with these complex kernels, we note that a likely cause is the fact that the sketched kernel likely contains zero or a few observations for each group and thus the estimation may be unreliable. This intuition is discussed further in \cite{yang2017randomized} (Example 4) about issues with sub-sampling sketching when there is considerable variation in the data. Future research might explore better methods for sketching to improve performance even with kernels that include fixed effects (e.g., those discussed in \citealt{lee2020econometric}). For example, stratified sampling based on the fixed effect may be helpful in improving performance.

\section{Additional Results: Newman (2016)}
\label{app:newman}

This appendix provides additional results for the analysis in Section~\ref{sec:newman}. First, we provide information on the questions analyzed in \cite{newman2016breaking}. For the main analysis, the respondent is coded as ``1'', rejecting meritocracy, if they agreed with the statement presented to them or with both statements if both were presented---shown below. Otherwise, they are coded ``0'' (\citealt[p. 1013]{newman2016breaking}).

\begin{itemize}
    \item ``Success in life is pretty much determined by forces outside our control'' 
    \item ''Hard work and determination are no guarantee of success for most people''
\end{itemize}

We consider the following two quantities. First, the average predicted probability $\bar{p}(e)$ as a function of earning inequality $e$ where $\bm{x}_i$ indicates all other covariates in the model. Second, we consider the popular average marginal effect $\mathrm{AME}(e)$ where we evaluate the derivative using the finite difference method (Appendix~\ref{app:mfx_diff}). Both quantities are calculated using the ``observed value'' strategy (e.g., \citealt{hanmer2013behind}) where one covariate (inequality) is set to some counterfactual value, all other covariates are held at their observed values, and the quantity of interest is the average across the observed sample.

\begin{align}
\bar{p}(e) &= \frac{1}{N} \sum_{i=1}^N \mathrm{Pr}(Y_i = 1 | \bm{x}_i, e_i = e) \\
\mathrm{AME}(e) &= \frac{1}{N} \sum_{i=1}^N \left.\frac{\partial \mathrm{Pr}(Y_i = 1 | \bm{x}_i, e_i)}{\partial e_i}\right|_{e_i = e}
\end{align}

For a secondary analyses as to the lack of a statistically detectable non-linear effect of economic inequality, we considered the following tests. Define $e^*_{min}$ as the smallest value of earnings inequality considered (0.349), define $e^*_{inf}$ as the model-specific inflection point---or the closest point in the grid of values considered, and define $e^*_{max}$ as the largest value considered (1.119). 

The first test examines the difference in average marginal effect at the end points, i.e. $\mathrm{AME}(e^*_{max}) - \mathrm{AME}(e^*_{min})$. For the original model in \cite{newman2016breaking}, the point estimate is -1.36 and the 95\% confidence interval (using the delta method) is $[-2.05, -0.67]$. For the \texttt{gKRLS} model, the point estimate is -0.44 and the 95\% confidence interval $[-1.28, 0.40]$ does contain zero. The second test compares the difference in changes in predicted probability above and below the inflection point, i.e. $\left[\bar{p}(e^*_{max}) - \bar{p}(e^*_{inf})\right] - \left[\bar{p}(e^*_{inf}) - \bar{p}(e^*_{min})\right] $. The 95\% confidence interval on this quantity for the \cite{newman2016breaking} model is $[-0.58, -0.07]$. The 95\% confidence interval for \texttt{gKRLS} is $[-0.36, 0.04]$ and does contain zero.

The third test considers the average \emph{second} derivative, i.e., average of the derivatives of the individual marginal effects:

$$ \frac{1}{N} \sum_{i=1}^N \left.\frac{\partial^2 \mathrm{Pr}(Y_i = 1 | \bm{x}_i, e_i)}{\partial e_i^2}\right|_{e_i = e} $$

\begin{figure}[!ht]
\caption{Average Second Derivative of Earnings Inequality}\label{fig:app_second_newman}
\includegraphics[width=\textwidth]{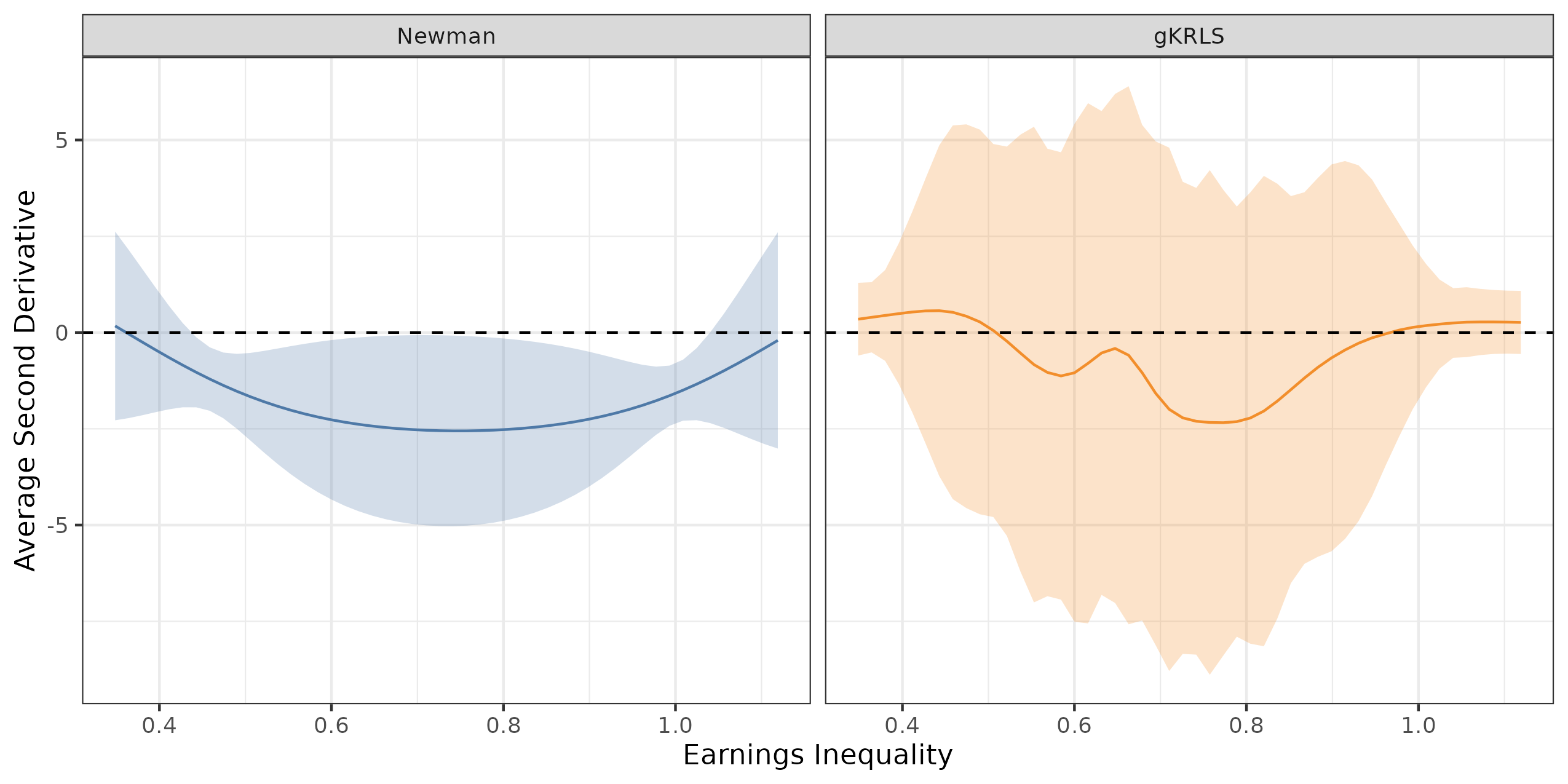}
\end{figure}

This tests the idea that for a concave function (as posited by \citealt[p. 1011]{newman2016breaking}), the second derivative should be negative. Figure~\ref{fig:app_second_newman} shows that, for the model used in \cite{newman2016breaking}, the average second derivative is usually negative and its confidence intervals do not contain zero except in the most extreme regions. For \texttt{gKRLS}, however, the confidence intervals contain zero at all points considered.

Thus, none of the tests shows evidence of a statistically distinguishable effect of a ``u-shaped'' relationship using \texttt{gKRLS}.

\subsection{Additional Questions}

The final test uses a secondary analysis in \cite{newman2016breaking} where, on a different survey, the following five questions were analyzed. The below text is quoted directly from the supporting information in \cite{newman2016breaking}. Each question is colored in \textcolor{red}{red} with our annotation. For the two ordinal questions, \cite{newman2016breaking} focuses on the probability of ``major reason''.

\begin{quote}
\textbf{Barriers to Women's Professional Advancement}

``As you may know, although women have moved into the work force in great numbers, very few top level business positions in this country are filled by women. There may be many reasons that there are so few women in high corporate positions. Here is a list of some of them. For each one, would you tell me whether you think it is a major reason, a minor reason, or not a reason why.'' (1) \textcolor{red}{Discrimination}: ``Women are discriminated against in all areas of life, and business is no exception'', and (2) \textcolor{red}{Old-Boy Networks}: ``Women who try to rise to the top of major corporations get held back by the `old-boy network''' [Q15B \& Q15E].  Constructed variables each have three ordered categories, ranging from (1)-``Not a reason'' (2)-``Minor reason'' (3)-``Major reason.'' 

\textbf{Traits of Men and Women}

``Now I would like to ask about some specific characteristics of men and women. For each  one I read, please tell me whether you think it is generally more true of men or more true of women'': (1) \textcolor{red}{Intelligent} [Q11A], and (2) \textcolor{red}{Arrogant} [Q11G]. For ``Intelligent,'' constructed variable is dichotomous, and coded "1" for respondents who believed the trait is ``More true of women'' and ``0'' otherwise. For ``Arrogant,'' constructed variable is dichotomous, and coded ``1'' for respondents who believed the trait is ``More true of men,'' and ``0'' otherwise.

\textbf{Gender Equality in Nation}

"Which of these two statements comes closer to your own views—even if neither is exactly right": "This country has made most of the changes needed to give women equal rights with men" OR "The country needs to \textcolor{red}{continue making changes}  to give women equal rights with men." [Q13]. Constructed variable coded "1" if respondent selected latter statement and "0" otherwise.
\end{quote}

\begin{figure}[!htbp]
\caption{Additional Questions in Newman (2016)}
\label{fig:app_extra_newman}
\includegraphics[width=\textwidth]{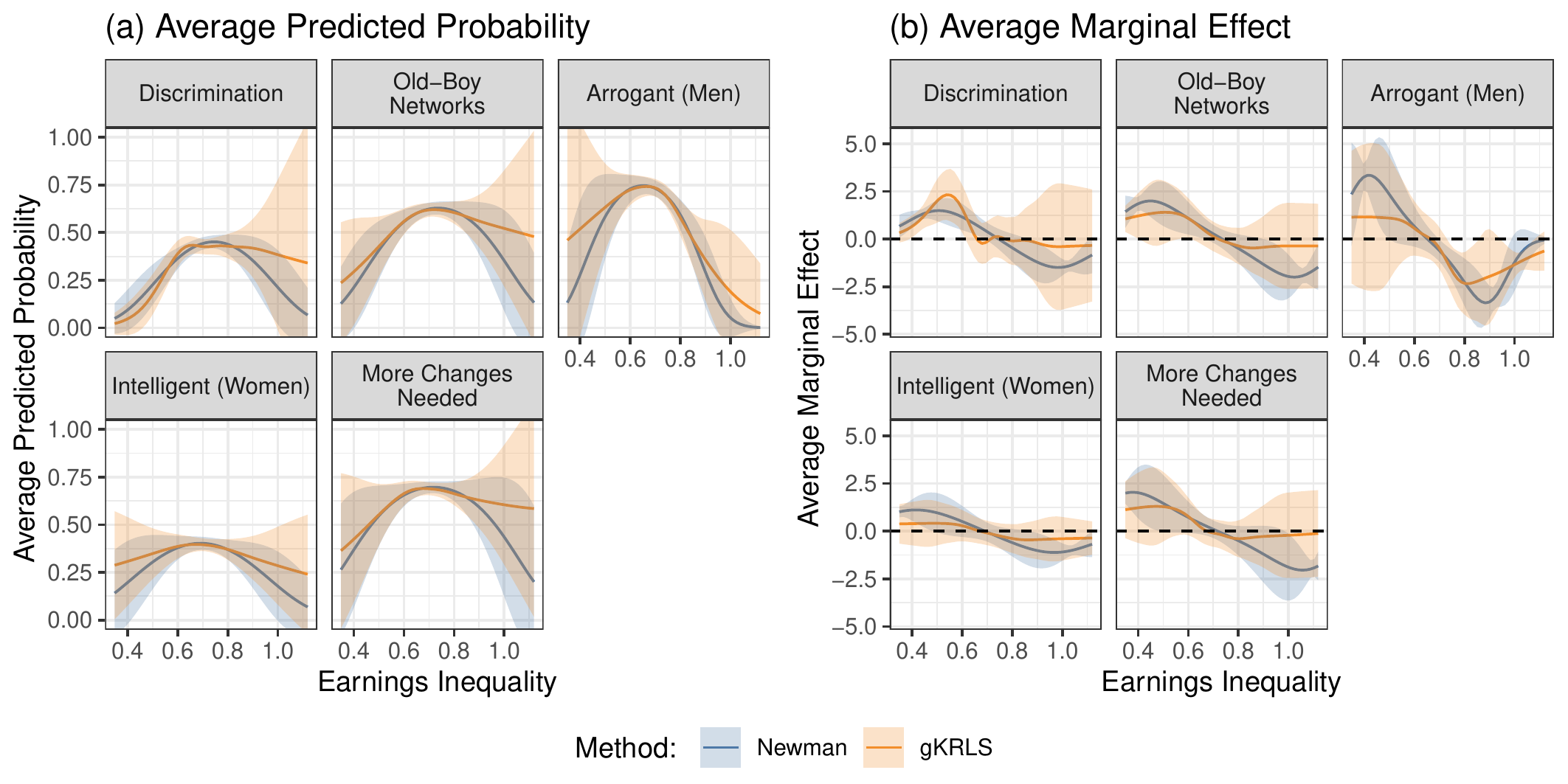}
\end{figure}

\begin{figure}[!htbp]
\caption{Average Second Derivative Additional Questions in Newman (2016)}
\label{fig:app_second_pew}
\includegraphics[width=\textwidth]{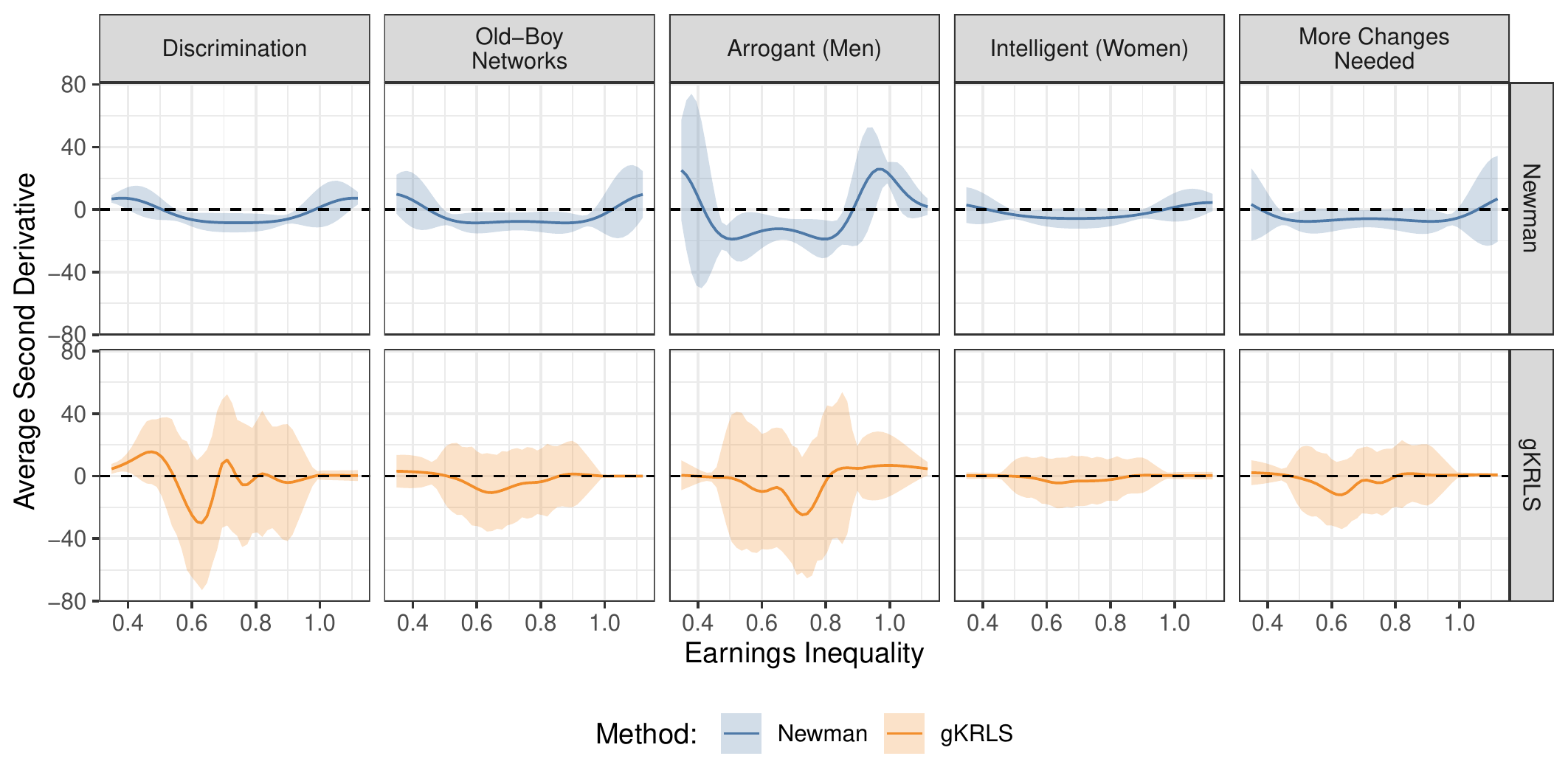}
\end{figure}

Figure~\ref{fig:app_extra_newman} shows the predicted probabilities and average marginal effects as a function of earnings inequality. As before, while the questions generally show an inverted ``u-shape'' estimated using \texttt{gKRLS}, there are concerns about a lack of a statistically detectable relationship. For most questions, the confidence intervals for the average marginal effects usually contain zero for most of the values for \texttt{gKRLS} but not for the specification in \cite{newman2016breaking}. Applying the additional tests in the previous section of the Appendix shows limited evidence for a statistically detectable effect for \texttt{gKRLS}---only one question passes either of the first two tests (men arrogant; difference in differences of $\bar{p}(e)$). The confidence interval on the average second derivative does not contain zero (and is negative) for many values of earnings inequality with the \cite{newman2016breaking} model, but usually contains zero for the \texttt{gKRLS} model; Figure~\ref{fig:app_second_pew} shows the results.

\section{Additional Results: Gulzar et al. (2020)}
\label{app:gulzar}

This appendix provides additional results for the analysis in Section~\ref{sec:main_gulzar}.

\subsection{Additional Results: Main Effects (Section~\ref{sec:main_gulzar})}
\label{app:gulzar_main}

We examine the sensitivity of the estimates to (i) the size of the sketching multiplier (5 or 15) and (ii) the use of \texttt{bam} or \texttt{gam} (see Appendix~\ref{app:gam_bam} for a discussion). For each of the twelve outcomes, we ran the model fifty times. Note that for the double/debiased machine learning method (DML-PLR; DML-ATE), there is an additional source of randomness---the five folds that the data is split into---that is also accounted for in the uncertainty.

\begin{figure}[!htbp]
\caption{Repeated Estimation of Double/Debiased Machine Learning with Sketching}
\label{fig:app_gulzar_rep_DML}
\begin{subfigure}[b]{\textwidth}
\caption{Partially Linear Regression (PLR)}
\includegraphics[width=\textwidth]{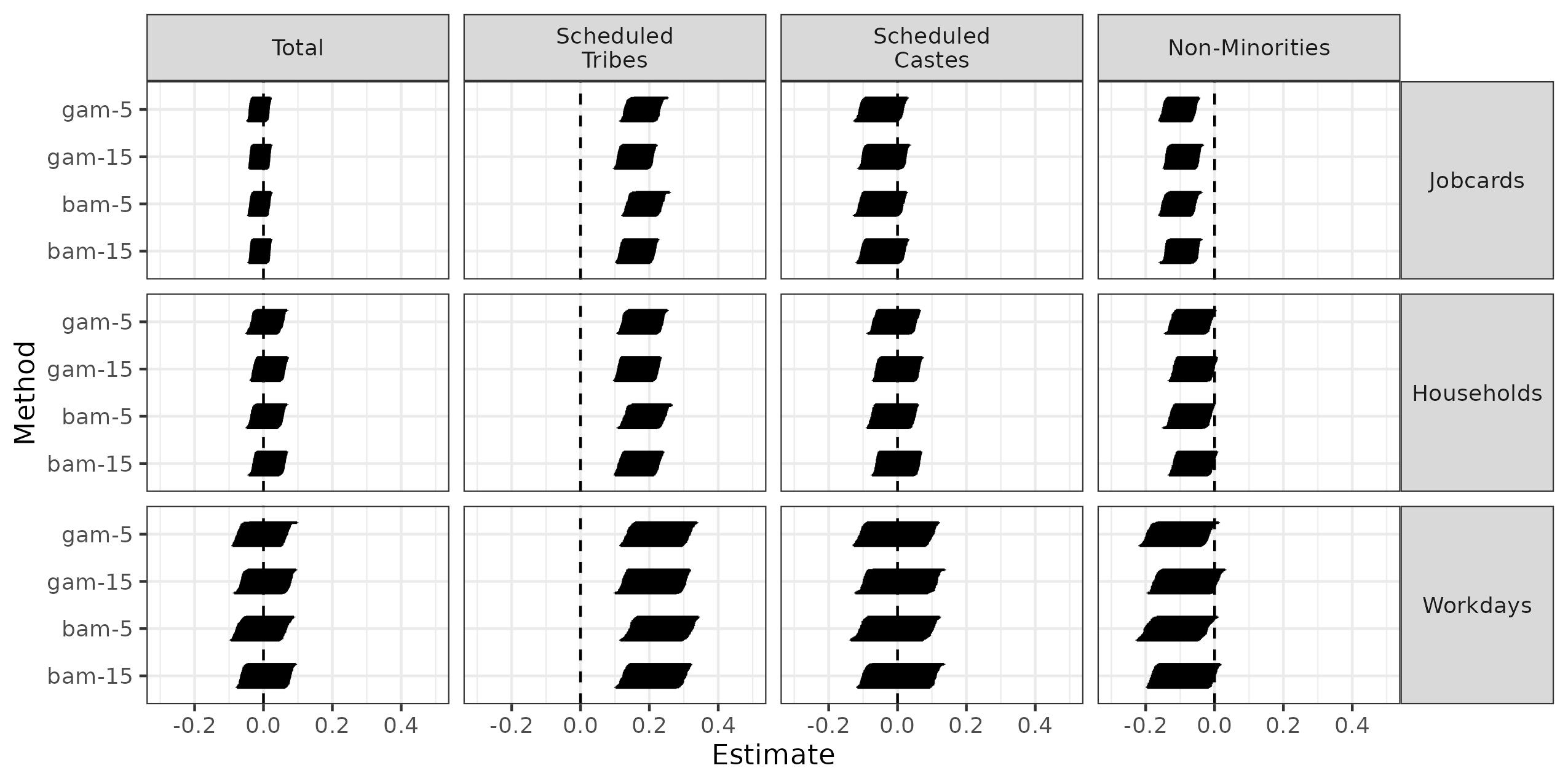}
\end{subfigure}
\begin{subfigure}[b]{\textwidth}
\caption{Average Treatment Effect (ATE)}
\includegraphics[width=\textwidth]{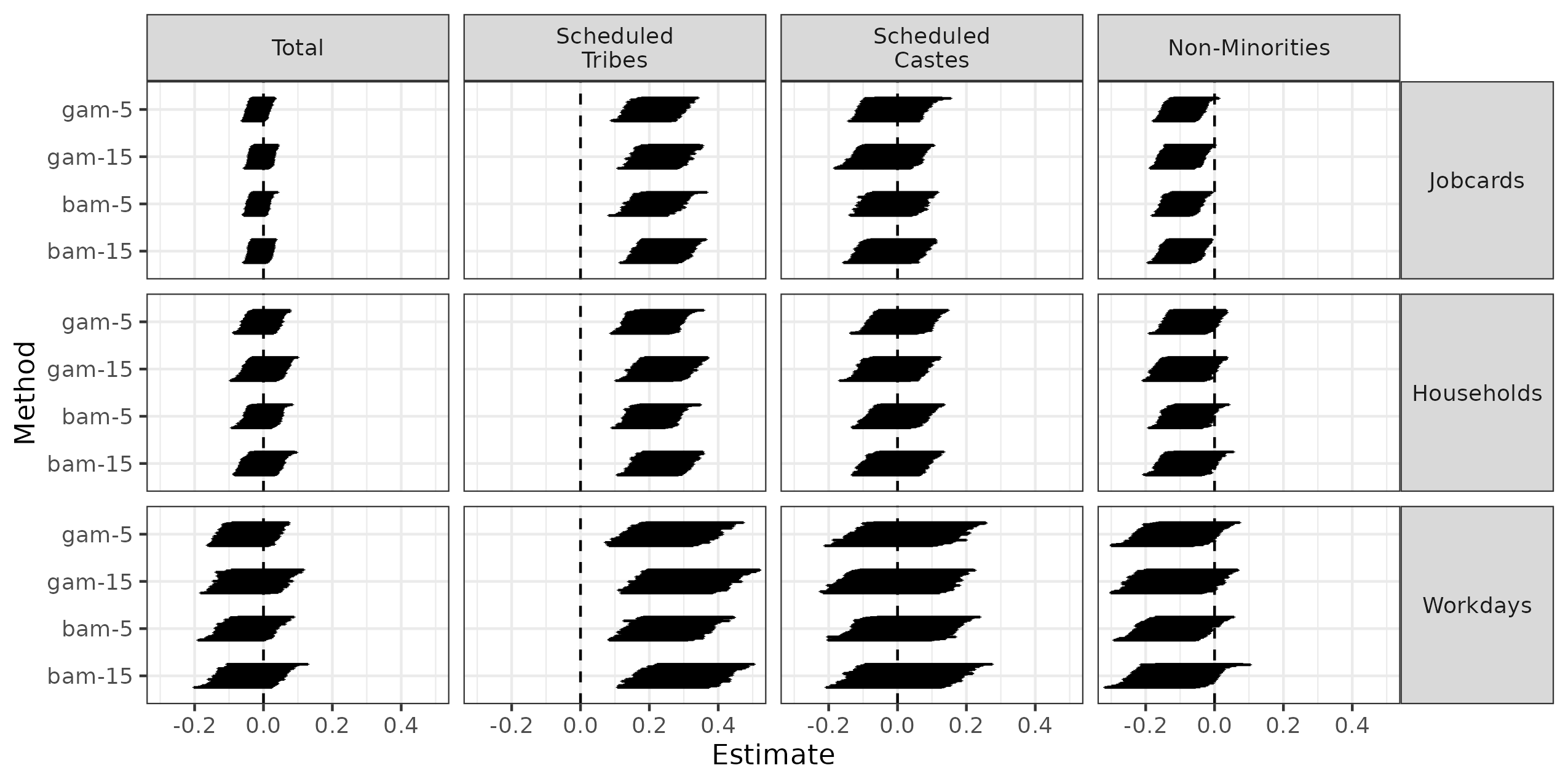}
\end{subfigure}
\caption*{\footnotesize \emph{Note:} This figure reports the estimated effects and 95\% confidence intervals from fifty repetitions of each model for each outcome and group. The horizontal axis is truncated as occasionally the confidence interval from a method is very large. The top panel shows thee results of DML partially linear regression (DML-PLR) and the bottom shows the results for the DML algorithm for estimating the ATE (DML-ATE).}
\end{figure}

Figure~\ref{fig:app_gulzar_rep_DML} presents the results for the two double/debiased machine learning methods (DML); the partially linear regression model (DML-PLR) and the average treatment effect model (DML-ATE). 95\% confidence intervals are shown and the estimated results are sorted by their point estimates for clarity. We see that, for both methods, there is a large amount of stability in the confidence intervals across repeated runs of the double/debiased machine learning procedure and the accompanying random sketching. It is rarely the case that re-running the model would change whether the confidence interval contains zero. The choice of \texttt{bam} vs \texttt{gam} and the multiplier also seem somewhat less important here. Across all methods, the ratio of the standard deviation of the point estimates to the average standard error is around 0.40 (for DML-ATE) and 0.20-0.25 (for DML-PLR).

\begin{figure}[!htbp]
\caption{Repeated Estimating \texttt{gKRLS} with Sketching}
\label{fig:app_gulzar_rep_gKRLS}
\begin{subfigure}[b]{\textwidth}
\caption{All Variables in Kernel}
\includegraphics[width=\textwidth]{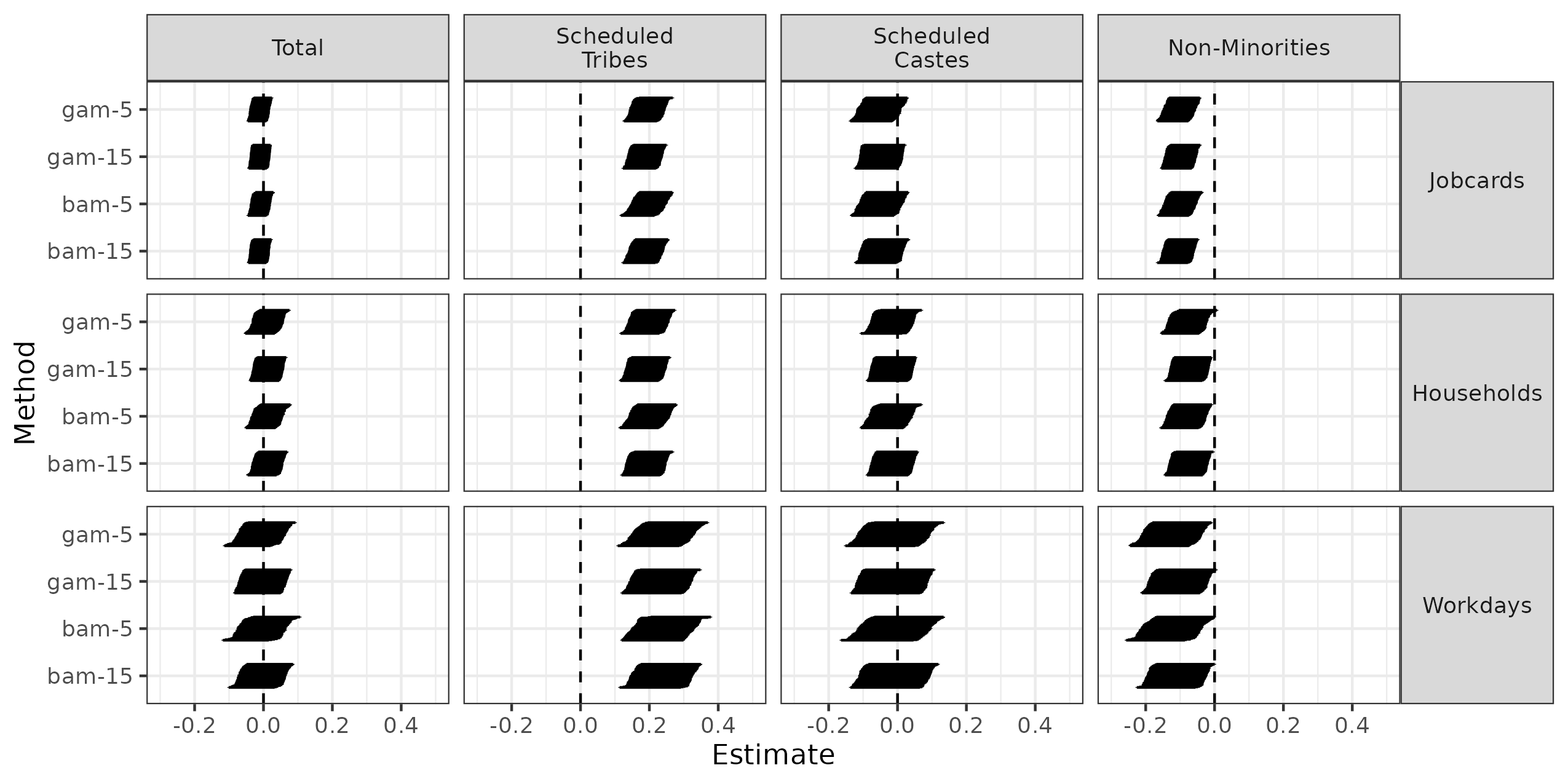}
\end{subfigure}
\begin{subfigure}[b]{\textwidth}
\caption{Geographic Kernel}
\includegraphics[width=\textwidth]{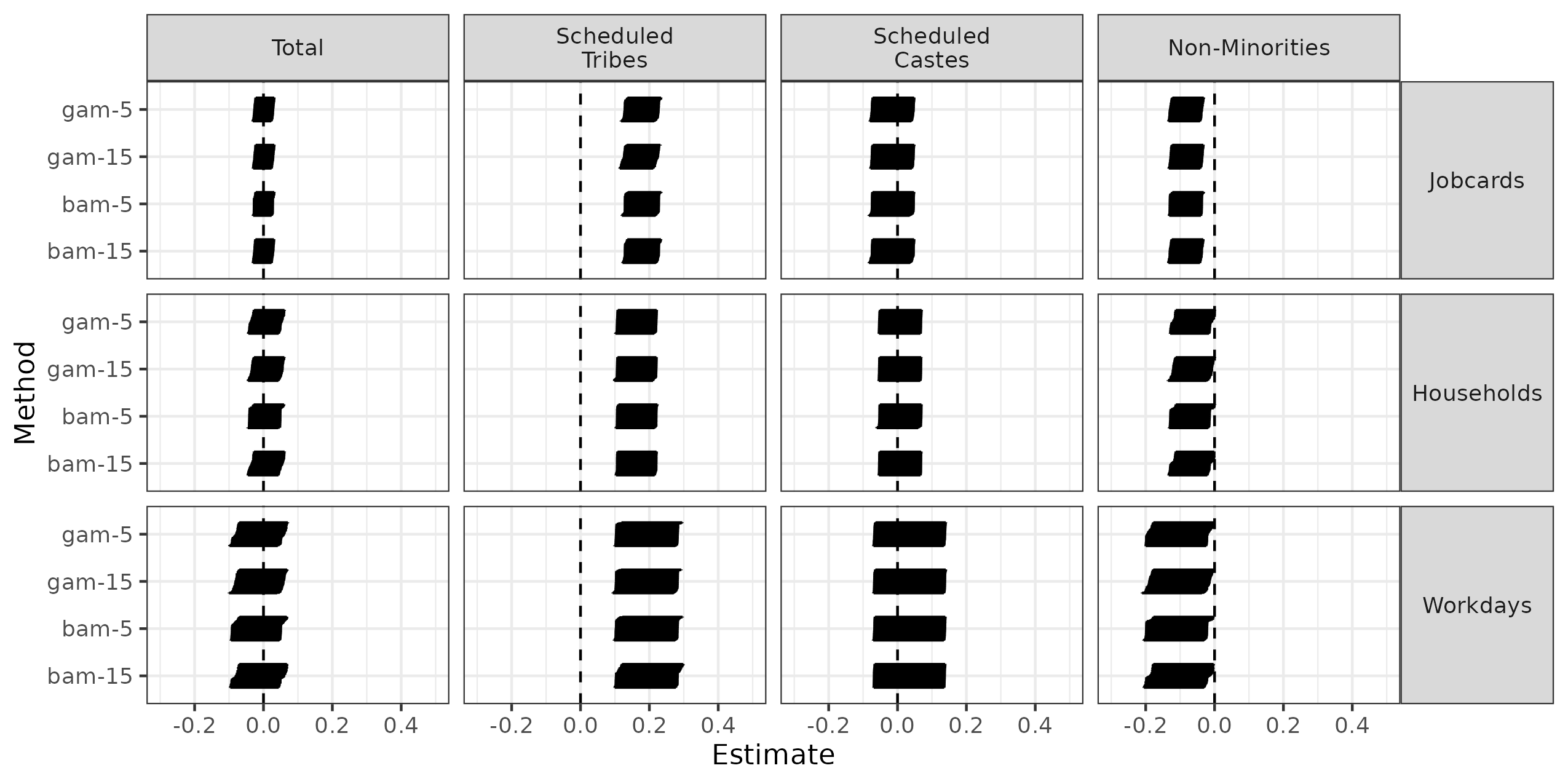}
\end{subfigure}
\caption*{\footnotesize \emph{Note:} This figure reports the estimated effects and 95\% confidence intervals from fifty repetitions of each model for each outcome and group. The horizontal axis is truncated as occasionally the confidence interval from a method is very large. The top panel shows the results of \texttt{gKRLS} that includes all covariates (``gKRLS (All)'') and the bottom includes results from a method that only includes the geographic coordinates (``gKRLS (Geog.)'' .}
\end{figure}

Figure~\ref{fig:app_gulzar_rep_gKRLS} considers repeated estimation of the two \texttt{gKRLS} models that include either all variables in the kernel (``gKRLS (All)'') or only the geographic coordinates (``gKRLS (Geog.)''), both described in the main text. It shows a broadly similar story to the DML methods. There is somewhat more variability when all variables are included in the kernel. Especially in this case, using the larger multiplier ($\delta=15$) decreases the ratio of the standard deviation of the estimates to the average standard error from around 0.40 (comparable to DML-ATE) to around 0.20 (around 0.30 for \texttt{bam}).

In terms of computational cost, Table~\ref{tab:app_time_rep_gulzar} reports the average estimation time for each method, averaged across all repeated estimations and outcome variables. It shows, as expected, that increasing the sketching multiplier can increase the cost considerably---especially for double/debiased machine learning methods or a kernel with many covariates (``gKRLS (All)'').  \texttt{bam} provides considerable increases in speed especially in the case of the DML methods. 

\begin{table}[!htbp]
\caption{Run Time of DML and \texttt{gKRLS} Methods on Gulzar et al. (2020)}
\label{tab:app_time_rep_gulzar}
\centering
\begin{tabular}{lll|ll}
\multirow{2}{*}{Method} & \multicolumn{2}{c|}{DML} & \multicolumn{2}{c}{\texttt{gKRLS}} \\
& PLR & ATE & Geog & All\\
\hline
\input{figures/Table_A3}
\end{tabular}
\caption*{\footnotesize \emph{Note:} This table reports the average estimation time in minutes on a computer with 8GB of RAM and 1 core, averaged across the simulations and outcome variables. ``DML'' reports the time of two methods (DML-PLR; DML-ATE) discussed in the main text. ``\texttt{gKRLS}'' reports the two kernel methods (``gKRLS (Geog.); ``gKRLS (All)'') discussed in the main text. ``Method'' indicates the method used with \texttt{bam} or \texttt{gam} and the sketching multiplier (``-5''; ``-15'').}
\end{table}

\subsection{Additional Results: Heterogeneous Effects}
\label{app:heteff_gulzar}

We conduct an additional application of \texttt{gKRLS} on the \cite{gulzar2020does} to estimate heterogeneous effects. There is a large and active literature on how to use arbitrary machine learning algorithms to estimate heterogeneous effects (e.g., \citealt{kunzel2019metalearners,nie2021quasi}). We focus on a recent method (``R-learner'') by \cite{nie2021quasi} as representing a current state-of-the-art method.

The R-learner is a type of meta-learner (see \citealt{kunzel2019metalearners} for a general review) that allows the researcher to use an arbitrary machine learning algorithm to estimate heterogeneous treatment effects. The core innovation of the R-learner is to estimate a heterogeneous treatment effect function $\tau^*(.)$ (as a function of a vector of pre-treatment covariates $X_i$---noting their notation \emph{differs} from that in the rest of our paper) by minimizing the following empirical analogue $\tilde{\tau}(.)$ on a dataset with $N$ observations where $m^*(x)$ and $e^*(x)$ represent the conditional mean outcome---$E[Y_i | X_i = x]$---and treatment propensity---$P(W_i = 1 | X_i = x)$, respectively. $\Lambda_n\{\tau(.)\}$ regularizes the estimated $\tau(.)$ function. 

\begin{equation}
\tilde{\tau}(.) = \argmin_{\tau} \frac{1}{N} \sum_{i=1}^N \left( \left[Y_i - m^*(X_i)\right] - \left[W_i - e^*(X_i)\right]\tau(X_i)\right)^2 + \Lambda_n\{\tau(.)\}
\end{equation}

One of the meanings of ``R'' in R-learner gestures at the fact that this function depends on \emph{residualizing} the observed outcome $Y_i$ and the treatment $W_i$ from their expected values ($m^*(X_i)$ and $e^*(X_i)$, respectively). This ensures a more robust objective function when trying to estimate the heterogeneous treatment effects (\citealt{nie2021quasi}). The key difficulty of estimating $\tilde{\tau}(.)$ is that the two key functions $m^*(.)$ and $e^*(.)$ are unknown so $\tilde{\tau}(.)$ cannot be estimated directly. A second key innovation of \citet[p. 301]{nie2021quasi} is to first estimate $m^*$ and $e^*$ and use these ``pilot estimates'' to create a feasible version for estimating $\tilde{\tau}(.)$.

They propose obtaining estimates $\hat{m}$ and $\hat{e}$ using a procedure known as ``cross-fitting.'' This process is similar in implementation to ensemble methods (stacking, SuperLearning) and starts by separating the data into $K$-folds. Using $K$-1 folds of the data, one estimates the conditional mean outcome (i.e. predict $Y_i$ with covariates $X_i$) and the propensity score (i.e. predicting treatment $W_i$ with covariates), and then generate predictions on the held out fold. By cycling through all of the folds, one gets an out-of-sample prediction for each observation. These are combined as shown below to estimate the heterogeneous treatment effect function $\hat{\tau}(.)$.  Formally, the R-learner algorithm is sketched below \citep[p. 301]{nie2021quasi}.

\begin{enumerate}
    \item Split the data into $K$ folds. Fit $\hat{m}$ and $\hat{e}$ using cross-fitting with some machine learning algorithm, i.e. hold out one fold and estimate the model using the other $K-1$.
    \item Estimate $\tau(.)$ using the out-of-sample predictions for each observation. Define $\hat{m}^{-k(i)}(X_i)$ as the estimate of $m^*(x)$ that does not include the fold $k$ of which $i$ is a member; similarly define $\hat{e}^{-k(i)}(X_i)$. The objective is shown below where some machine learning algorithm is used to estimate $\hat{\tau}(.)$.\footnote{In the spirit of their accompanying code, we perform one final step of $K$-fold cross-validation (with the same folds) for the final estimation of $\hat{\tau}(.)$. If one were to estimate effects on truly out-of-sample data, the $\hat{\tau}(.)$ estimated on the entire training data could be used.}

\begin{equation}
\hat{\tau}(.) = \argmin_{\tau} \frac{1}{N} \sum_{i=1}^N \left( \left[Y_i - m^{-k(i)}(X_i)\right] - \left[W_i - e^{-k(i)}(X_i)\right]\tau(X_i)\right)^2 + \Lambda_n\{\tau(.)\}
\end{equation}
\end{enumerate}

While any machine learning algorithm can be used for the R-learner, there is an especially interesting reason to use \texttt{gKRLS}. \citet{nie2021quasi} prove that if one uses traditional KRLS in estimating $\hat{\tau}(.)$ (and some weaker assumptions on the quality of the estimates of $\hat{m}$ and $\hat{e}$), then the resulting estimator $\hat{\tau}(.)$ has a ``quasi-oracle'' property. Roughly speaking, this means that the accuracy on estimating $\hat{\tau}(.)$ is asymptotically equivalent to the accuracy one would obtain if the researcher \emph{knew} $m^*(X_i)$ and $e^*(X_i)$ exactly. Thus, there is no loss of information from having to use the estimated analogues. This provides a formal justification for using the ``pilot estimates'' $\hat{m}^{-k(i)}(X_i)$ and $\hat{e}^{-k(i)}(X_i)$ in the final estimation of $\hat{\tau}(.)$.

In their numerical experiments, \citet{nie2021quasi} use \citet{sonnet2018krls}'s implementation of traditional KRLS for Gaussian and binary outcomes for estimating $\hat{m}(.)$, $\hat{e}(.)$ and $\hat{\tau}(.)$. However, this has the disadvantages discussed in Sections~\ref{sec:define_gKRLS} and~\ref{sec:scalable_gKRLS} (e.g., requiring cross-validation for the binary case and being quite slow). Indeed, \cite{nie2021quasi} consider a coarse grid of only thirteen values when calibrating $\lambda$ and only examine problems of 500 or 1000 observations. Thus, \texttt{gKRLS} allows for the theoretical promise of the R-learner when combined with KRLS to be scaled to much larger datasets.

We apply this to the \citet{gulzar2020does} application, using \texttt{gKRLS} for all machine learning procedures where we include the covariates linearly as well as kernel that includes all of the covariates. The formula in pseudo-code (see Appendix~\ref{app:software}) is shown below where ``\texttt{x1 + x2 + ...}'' indicates the covariates.\footnote{After our first explorations of the \cite{gulzar2020does} data, we noted that two controls (share of scheduled tribes in 1991 and 2001) are extremely highly correlated with the treatment indicator in Himachal Pradesh (0.87 and 0.94, respectively; standard error of 0.039 and 0.027, respectively) but in no other state (correlations ranging from -0.05 and 0.25; standard errors ranging from 0.010 to 0.031). No other control variable's correlation with treatment within a state with treatment is higher than 0.47 (standard error of 0.070). Following supplemental analyses in the original paper, we find that this imbalance on "share of scheduled tribes in 2001" in Himachal Pradesh between treated and untreated units does not decrease as the geographic bandwidth decreases, although it does decline in all other states. The imbalance on "share of scheduled tribes in 1991" does decrease although more slowly. Exploring this and the implications for \cite{gulzar2020does} in more detail is outside of the scope of this project. However, we exclude these two covariates for the heterogeneous effect analysis as their inclusion led to unstable estimates for Himachal Pradesh given minor changes in specification. Their exclusion has a limited effect on the results in the main text (e.g., Figure~\ref{fig:maineffect}).}

\begin{center}
\begin{verbatim}
            y ~ x1 + x2 + ... + s(x1, x2, ..., bs = "gKRLS")
\end{verbatim}
\end{center}

\begin{figure}[!htbp]
    \caption{Heterogeneous Effects in \cite{gulzar2020does}}
    \label{fig:heter_eff}
    \centering
    \includegraphics[width=\textwidth]{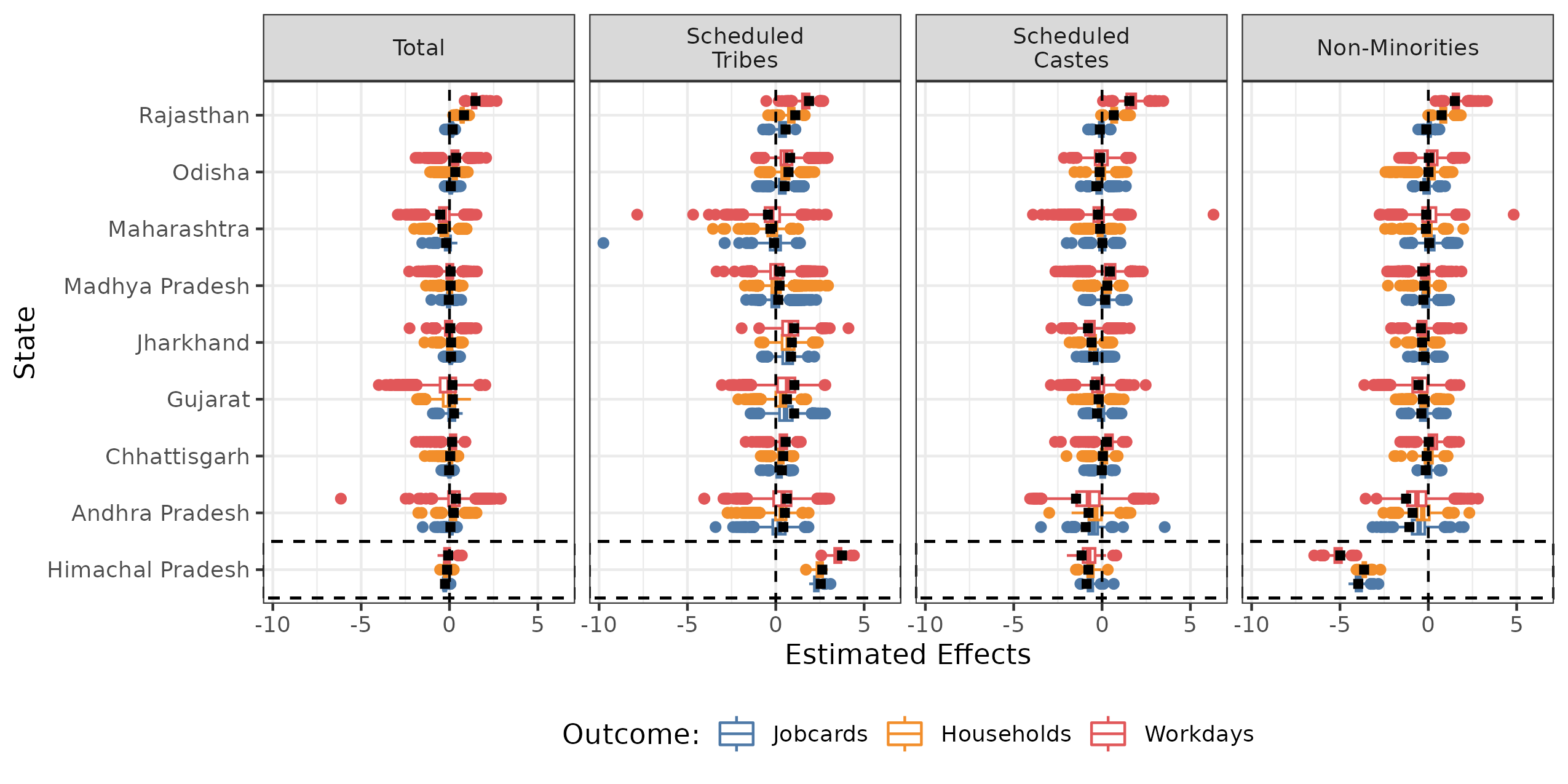}
    \caption*{\footnotesize \emph{Note}: This figure shows the distribution of heterogeneous treatment effects by state, outcome, and group using the R-learner approach described in the main text. The $\blacksquare$ indicates the treatment effect estimated within each state using a difference-in-means estimator.}
\end{figure}

The procedure takes under three minutes on a machine with 8 GB of RAM and a single core.  Our initial examinations of the estimated heterogeneous effects suggested a key role of geography. Figure~\ref{fig:heter_eff} shows that, in general, one state---Himachal Pradesh---exhibits quite different patterns of estimated treatment effects. It shows the distribution of effects for each observation in each state across the four groups and three outcomes. In total, their analysis includes nine states. We also include a symbol ($\blacksquare$) to indicate the estimated treatment effect, without controls, for each variable inside of each state. It is reassuring that the treatment effects obtained from the R-learner are similar to those calculated by simply calculating the difference-in-means within each state.

In substantive terms, Himachal Pradesh has much larger effects for the targeted minority group (Scheduled Tribes). We also find slightly negative effects for the non-targeted minority group (Scheduled Castes) and large negative effects for the other individuals (non-minorities). While fully exploring the reason for the differential effects in Himachal Pradesh is outside of the scope of this paper, we note that it is distinct amongst the states considered in that it (i) has the smallest population, (ii) has the largest share of the \emph{non-targeted} minority group (Scheduled Castes; 25\% of the population in 2001), and (ii) has the smallest share of the targeted group (4\% of the population in 2001). This suggests a possibility of an interesting qualification of the effect of an electoral quota scheme to benefit a small minority group when there is another larger, although still historically disadvantaged group, who is not affected by the scheme. We found similar distributions of estimates by state if we used different multipliers, e.g., 5 vs. 15, and \texttt{gam} instead of \texttt{bam}.


\section{Software}
\label{app:software}

This appendix briefly discusses software packages for estimating KRLS and provides a brief demonstration on how to use the core function in the \texttt{gKRLS} package.   

\subsection{Existing Software Packages}

In the main paper, we consider \texttt{KRLS} and bigKRLS as the two well-known packages for estimating KRLS to political scientists. They are only able to assume a Gaussian outcome and must include all predictors in the kernel. An additional package \texttt{KSPM} (\citealt{schramm2020kspm}) provides additional flexibility by allowing multiple kernels and additive terms. It is still considerably less flexible than \texttt{mgcv} (e.g., a lack of other methods of penalization, the use of non-Gaussian outcomes, sketching, selection of penalty parameter using something other than LOO-CV, etc.).  It does have some novel features (e.g., the use of alternative kernels besides the Gaussian one; interactions between kernels) that are straightforward to include in future development of \texttt{gKRLS}. Unfortunately, preliminary experiments also suggested that it was considerably slower than either \texttt{KRLS} or \texttt{mgcv} and thus it was not explored further in our empirical analyses.

\subsection{Software Demonstration}

We demonstrate how to estimate \texttt{gKRLS}, average marginal effects, and double/debiased machine learning.

\begin{lstlisting}[language=R]
# to install gKRLS
# install.packages('gKRLS')
# load SuperLearner before gKRLS/mgcv to address clashing object names
library(SuperLearner) 
library(gKRLS) # this also load "mgcv" and "sandwich"
library(DoubleML)
library(caret) # for sample splitting 

# simulate data
n <- 1000
x1 <- rbinom(n, 1, 0.45)
x2 <- rnorm(n, 0, 1)
x3 <- rnorm(n, 0, 1)
x4 <- rnorm(n, 0, 1)
state <- as.factor(sample(letters[1:10], n, replace = TRUE))
error <- rnorm(n, 0, 0.5)

# target function
y <- 0.6*x1 +0.15*x2 -0.2*x1*x2 + 0.3*x3^3 - 0.2*x4^2 + error

data <- data.frame(y, x1, x2, x3, x4, state)

\end{lstlisting}

\hfill \break
We estimate three versions of \texttt{gKRLS}; one that includes all variables in the kernel, one that includes some in a linear fashion (i.e., fixed effects for state), and one that includes two kernels.

\begin{lstlisting}
# All variables in kernel: "gam" is exported from "mgcv"
gkrls_est <- gam(y ~ s(x1, x2, x3, x4, bs = "gKRLS"),
    data = data)
summary(gkrls_est)

# State fixed effect 
gkrls_fx <- gam(y ~ state + s(x1, x2, x3, x4, bs = "gKRLS"), 
    data = data)

# Multiple kernels
gkrls_dk <- gam(
    y ~ state + s(x1, x2, bs = "gKRLS") + 
        s(x3, x4, bs = "gKRLS"), 
    data = data)
\end{lstlisting}

\hfill \break
We can estimate predictions and marginal effects using the accompanying functions. If \texttt{individual=TRUE}, the individual marginal effect (i.e. for each observation) in addition to the average is returned. If marginal effects on only certain variables are desired, this can be specified using the \texttt{variables} argument.

\begin{lstlisting}
gkrls_pred <- predict(gkrls_fx, newdata = data) 
gkrls_ame <- calculate_effects(gkrls_fx, 
    variables = c("x1", "x2"),
    continuous_type = "derivative", individual = T)
\end{lstlisting}

\hfill \break

A variety of estimation options can be set using the \texttt{gKRLS} argument to \texttt{xt}. This is further described on the package documentation and some options are shown below, e.g. changing multiplier and sketching method.

\begin{lstlisting}
gkrls_alt <- gam(y ~ s(x1, x2, x3, x4, bs = "gKRLS", 
        xt = gKRLS(sketch_method = "gaussian",
            sketch_multiplier = 15)),
    data = data)
\end{lstlisting}

\hfill \break
We can also integrate gKRLS with double/debiased maching learning method. 

\begin{lstlisting}
# double/debiased machine learning
ml_g <- LearnerRegrBam$new()
ml_g$param_set$values$formula <- ~ s(x2, x3, x4, bs = "gKRLS")
ml_m <- LearnerRegrBam$new()
ml_m$param_set$values$formula <- ~ s(x2, x3, x4, bs = "gKRLS")
ml_g$param_set$values$method <- "REML"
ml_m$param_set$values$method <- "REML"

data_DML <- double_ml_data_from_data_frame(
  df = data,
  y_col = "y",
  d_cols = "x1",
  x_cols = setdiff(names(data), c("y", "x1", "state"))
) 
# Fit Partial Linear Regression
dml_plr <- DoubleMLPLR$new(data_DML, ml_g, ml_m)
dml_plr$fit()
\end{lstlisting}

\hfill \break
Finally, we show how to use \texttt{gKRLS} to estimate heterogeneous effects using the \texttt{SuperLearner} package and the R-learner 
(\citealt{nie2021quasi}) discussed in Section~\ref{app:heteff_gulzar}.

\begin{lstlisting}
# Set x1 as "treatment" for this analysis
data$treatment <- data$x1
data$x1 <- NULL
# Define function for R learner
heteff_SL <- function(...){suppressMessages(
    SL.mgcv(..., bam = T, method = "REML",
    formula = ~ s(x2, x3, x4, bs = "gKRLS")))
}
# Create the folds for post-stratification using
# caret. Ensure they are the same across both
# initial estimations
id <- createFolds(1:nrow(data), 5)

# Estimate the conditional mean function
fit_SL_m <- SuperLearner(Y = data$y, 
 X = data, family = 'gaussian',
 SL.library = 'heteff_SL', 
 cvControl = list(V = 5, validRows = id))

# Estimate the propensity score
fit_SL_e <- SuperLearner(Y = data$treatment,
 X = data, family = 'binomial',
 SL.library = 'heteff_SL', 
 cvControl = list(V = 5, validRows = id),
 verbose = T)

# Extract estimated propensity scores
estimated_ps <- fit_SL_e$Z[,1]
# Truncate to avoid extreme scores
estimated_ps[estimated_ps < 0.01] <- 0.01
estimated_ps[estimated_ps > 0.99] <- 0.99
# Get the R-learner outcome
data$resid_PS <- data$y - estimated_ps
data$resid_outcome <- data$y - fit_SL_m$Z[,1]
data$rlearner_outcome <- data$resid_outcome/data$resid_PS

# Esimate heterogeneous effect
fit_SL_R <- SuperLearner(Y = data$rlearner_outcome, 
 X = data, family = 'gaussian', 
 obsWeights = data$resid_PS^2,
 SL.library = 'heteff_SL', 
 cvControl = list(V = 5, validRows = id),
 verbose = T)
# Get the cross-validated estimates of heterogeneous 
# treatment effect using held-out data
data$stacked.heteffect <- fit_SL_R$Z[,1]
# Get the estimates fit on the entire dataset
data$fullsample.heteffect <- fit_SL_R$SL.predict[,1]
# To predict for new out-of-sample data
predict(fit_SL_R, newdata = data[1:5,])
\end{lstlisting}

\end{document}

%% file: figures/Table_A1.tex
\texttt{KRLS} (default) & 0.526 & 0.170 & 0.086 \\
\texttt{KRLS} (Bayes SE) & 0.859 & 0.170 & 0.167 \\
\texttt{gKRLS} & 0.850 & 0.175 & 0.182 \\
\texttt{gKRLS} + Linear & 0.931 & 0.218 & 0.261 \\
\texttt{gam} & 0.939 & 0.234 & 0.277 \\ \hline\hline

%% file: figures/Table_A2.tex
\texttt{gam} & 0.830 & 0.159 & 0.146 & 0.910 & 0.159 & 0.177 \\
\texttt{gKRLS} & 0.775 & 0.148 & 0.126 & 0.879 & 0.148 & 0.153 \\
\texttt{gKRLS} + Linear & 0.844 & 0.156 & 0.151 & 0.909 & 0.156 & 0.172 \\
\texttt{KRLS} (default) & 0.522 & 0.150 & 0.072 &  &  &  \\ \hline\hline

%% file: figures/Table_A3.tex
bam-5 &  0.63 &  1.37 & 0.04 & 0.07 \\
bam-15 &  2.31 &  6.40 & 0.09 & 0.26 \\
gam-5 &  7.54 &  5.71 & 0.48 & 1.00 \\
gam-15 & 45.14 & 38.15 & 0.63 & 5.93 \\

%% file: reference.bib
@article{hainmueller2014kernel,
  title={Kernel Regularized Least Squares: Reducing Misspecification Bias with a Flexible and Interpretable Machine Learning Approach},
  author={Hainmueller, Jens and Hazlett, Chad},
  journal={Political Analysis},
  volume={22},
  number={2},
  pages={143--168},
  year={2014}
}

@article{wood2011fast,
  title={Fast Stable Restricted Maximum Likelihood and Marginal Likelihood Estimation of Semiparametric Generalized Linear Models},
  author={Wood, Simon N},
  journal={Journal of the Royal Statistical Society: Series B (Statistical Methodology)},
  volume={73},
  number={1},
  pages={3--36},
  year={2011}
}

@article{yang2017randomized,
  title={Randomized Sketches for Kernels: Fast and Optimal Nonparametric Regression},
  author={Yang, Yun and Pilanci, Mert and Wainwright, Martin J},
  journal={The Annals of Statistics},
  volume={45},
  number={3},
  pages={991--1023},
  year={2017}
}

@article{mohanty2019messy,
  title={Messy Data, Robust Inference? Navigating Obstacles to Inference with {bigKRLS}},
  author={Mohanty, Pete and Shaffer, Robert},
  journal={Political Analysis},
  volume={27},
  number={2},
  pages={127--144},
  year={2019}
}

@article{drineas2005nystrom,
  title={On the Nystr{\"o}m Method for Approximating a Gram Matrix for Improved Kernel-Based Learning},
  author={Drineas, Petros and Mahoney, Michael W},
  journal={Journal of Machine Learning Research},
  volume={6},
  number={12},
  year={2005},
  pages={2153--2175}
}

@article{kunzel2019metalearners,
  title={Metalearners for Estimating Heterogeneous Treatment Effects using Machine Learning},
  author={K{\"u}nzel, S{\"o}ren R and Sekhon, Jasjeet S and Bickel, Peter J and Yu, Bin},
  journal={Proceedings of the National Academy of Sciences},
  volume={116},
  number={10},
  pages={4156--4165},
  year={2019}
}

@article{nie2021quasi,
  title={Quasi-Oracle Estimation of Heterogeneous Treatment Effects},
  author={Nie, Xinkun and Wager, Stefan},
  journal={Biometrika},
  volume={108},
  number={2},
  pages={299--319},
  year={2021}
}

@article{gulzar2020does,
  title={Does Political Affirmative Action Work, and for Whom? Theory and Evidence on India’s Scheduled Areas},
  author={Gulzar, Saad and Haas, Nicholas and Pasquale, Benjamin},
  journal={American Political Science Review},
  volume={114},
  number={4},
  pages={1230--1246},
  year={2020}
}

@article{bell2015explaining,
  title={Explaining Fixed Effects: Random Effects Modeling of Time-Series Cross-Sectional and Panel Data},
  author={Bell, Andrew and Jones, Kelvyn},
  journal={Political Science Research and Methods},
  volume={3},
  number={1},
  pages={133--153},
  year={2015}
}

@article{wood2016smoothing,
  title={Smoothing Parameter and Model Selection for General Smooth Models},
  author={Wood, Simon N and Pya, Natalya and S{\"a}fken, Benjamin},
  journal={Journal of the American Statistical Association},
  volume={111},
  number={516},
  pages={1548--1563},
  year={2016},
  publisher={Taylor \& Francis}
}

@article{chiang2022multiway,
  title={Multiway Cluster Robust Double/Debiased Machine Learning},
  author={Chiang, Harold D and Kato, Kengo and Ma, Yukun and Sasaki, Yuya},
  journal={Journal of Business \& Economic Statistics},
  volume={40},
  number={3},
  pages={1046--1056},
  year={2022},
  publisher={Taylor \& Francis}
}

@article{chernozhukov2018dml,
  year = {2018},
  volume = {21},
  number = {1},
  pages = {1--68},
  author = {Victor Chernozhukov and Denis Chetverikov and Mert Demirer and Esther Duflo and Christian Hansen and Whitney Newey and James Robins},
  title = {Double/Debiased Machine Learning for Treatment and Structural Parameters},
  journal = {The Econometrics Journal}
}

@book{wood2017mgcv,
  publisher = {Chapman and Hall/{CRC}},
  author = {Simon N. Wood},
  year = {2017},
  title = {Generalized Additive Models}
}

@article{sonnet2018krls,
    title = {Kernel Regularized Logistic Regression: Avoiding Misspecification Bias while Maintaining Interpretability for Binary Outcome Regressions},
    author = {Sonnet, Luke and Hazlett, Chad},
    journal = {Working Paper},
    year = {2018}
}

@article{wood2015generalized,
  title={Generalized Additive Models for Large Data Sets},
  author={Wood, Simon N and Goude, Yannig and Shaw, Simon},
  journal={Journal of the Royal Statistical Society: Series C (Applied Statistics)},
  volume={64},
  number={1},
  pages={139--155},
  year={2015},
  publisher={Wiley Online Library}
}

@article{shun1995laplace,
  title={Laplace Approximation of High Dimensional Integrals},
  author={Shun, Zhenming and McCullagh, Peter},
  journal={Journal of the Royal Statistical Society: Series B (Methodological)},
  volume={57},
  number={4},
  pages={749--760},
  year={1995},
  publisher={Wiley Online Library}
}

@article{liu2007semiparametric,
  title={Semiparametric Regression of Multidimensional Genetic Pathway Data: Least-squares Kernel Machines and Linear Mixed Models},
  author={Liu, Dawei and Lin, Xihong and Ghosh, Debashis},
  journal={Biometrics},
  volume={63},
  number={4},
  pages={1079--1088},
  year={2007},
  publisher={Wiley Online Library}
}

@article{schramm2020kspm,
  title={KSPM: A Package For Kernel Semi-Parametric Models},
  author={Catherine Schramm and S{\'e}bastien Jacquemont and Karim Oualkacha and Aurélie Labbe and Celia M. T. Greenwood},
  journal={The R Journal},
  volume={12},
  number={2},
  pages={82--106},
  year={2020}
}

@article{zhang2011bayesian,
  title={Bayesian Generalized Kernel Mixed Models},
  author={Zhang, Zhihua and Dai, Guang and Jordan, Michael I},
  journal={Journal of Machine Learning Research},
  volume={12},
  pages={111--139},
  year={2011}
}

@inproceedings{rahimi2007random,
  title={Random Features for Large-Scale Kernel Machines},
  author={Rahimi, Ali and Recht, Benjamin},
  booktitle={Advances in Neural Information Processing Systems},
  volume={20},
  year={2007},
  pages = {1177--1184}
}

@article{lee2020econometric,
  title={An Econometric Perspective on Algorithmic Subsampling},
  author={Lee, Sokbae and Ng, Serena},
  journal={Annual Review of Economics},
  volume={12},
  pages={45--80},
  year={2020},
  publisher={Annual Reviews}
}

@article{leeper2016mfx,
  author = {Leeper, Thomas J.},
  year={2016},
  title={Interpreting Regression Results using Average Marginal Effects with R's margins},
  url = {https://s3.us-east-2.amazonaws.com/tjl-sharing/assets/AverageMarginalEffects.pdf}
}

@article{newman2016breaking,
  title={Breaking the Glass Ceiling: Local Gender-Based Earnings Inequality and Women's Belief in the American Dream},
  author={Newman, Benjamin J},
  journal={American Journal of Political Science},
  volume={60},
  number={4},
  pages={1006--1025},
  year={2016},
  publisher={Wiley Online Library}
}

@article{hazlett2022mlm,
  title={Understanding, Choosing, and Unifying Multilevel and Fixed Effect Approaches},
  author={Hazlett, Chad and Wainstein, Leonard},
  journal={Political Analysis},
  volume={30},
  number={1},
  pages={46--65},
  year={2022},
  publisher={Cambridge University Press}
}

@article{wood2006ci,
  title={On confidence intervals for generalized additive models based on penalized regression splines},
  author={Wood, Simon N},
  journal={Australian \& New Zealand Journal of Statistics},
  volume={48},
  number={4},
  pages={445--464},
  year={2006},
  publisher={Wiley Online Library}
}

@article{cameron2015practitioner,
  title={A Practitioner’s Guide to Cluster-Robust Inference},
  author={Cameron, A Colin and Miller, Douglas L},
  journal={Journal of Human Resources},
  volume={50},
  number={2},
  pages={317--372},
  year={2015},
  publisher={University of Wisconsin Press}
}

@article{marra2012coverage,
  title={Coverage properties of confidence intervals for generalized additive model components},
  author={Marra, Giampiero and Wood, Simon N},
  journal={Scandinavian Journal of Statistics},
  volume={39},
  number={1},
  pages={53--74},
  year={2012},
  publisher={Wiley Online Library}
}

@article{hanmer2013behind,
  title={Behind the Curve: Clarifying the Best Approach to Calculating Predicted Probabilities and Marginal Effects from Limited Dependent Variable Models},
  author={Hanmer, Michael J and Kalkan, Kerem Ozan},
  journal={American Journal of Political Science},
  volume={57},
  number={1},
  pages={263--277},
  year={2013}
}

@inproceedings{dataverse_gKRLS,
    author = {Chang, Qing and Goplerud, Max},
    organization = {Harvard Dataverse},
    title = {{Replication Data for: Generalized Kernel Regularized Least Squares}},
    year = {2023},
    version = {Version 1},
    doi = {10.7910/DVN/WNW0AD},
    url = {https://doi.org/10.7910/DVN/WNW0AD}
}
